\documentclass[runningheads]{llncs}
\makeatletter
\def\input@path{{./eccv_style/}} 
\makeatother

\usepackage{eccv}

\usepackage{eccvabbrv}

\usepackage{graphicx}
\usepackage{booktabs}

\usepackage[accsupp]{axessibility}

\usepackage[pagebackref,breaklinks,colorlinks,citecolor=eccvblue]{hyperref}

\usepackage{orcidlink}

\newcommand{\fig}[1]{Fig. \ref{#1}}

\newcommand{\tab}[1]{Tab. \ref{#1}}

\definecolor{isabelline}{RGB}{244, 240, 236}
\definecolor{kaiming-green}{RGB}{57,181,74}
\definecolor{lightblue}{RGB}{240, 246, 254}
\definecolor{lightgray}{RGB}{242,242,242}
\definecolor{rust}{RGB}{192,79,21}
\definecolor{coral}{RGB}{243,151,144}
\definecolor{iceblue}{RGB}{199,214,234}

\usepackage{animate}
\usepackage{xcolor} 
\usepackage{arydshln}
\usepackage{float}
\usepackage{xspace}
\usepackage{placeins}
\usepackage{graphicx}
\usepackage{enumitem}
\usepackage{amsmath}
\usepackage{multirow}
\usepackage{pifont}
\usepackage{marvosym}
\usepackage{colortbl}
\usepackage{wrapfig}
\usepackage{subcaption}
\usepackage{bm}
\usepackage{threeparttable}
\usepackage{ragged2e}
\usepackage{animate}

\usepackage{algorithmic}
\usepackage{amssymb}
\usepackage{array}
\newcolumntype{C}[1]{>{\centering\arraybackslash}p{#1}}

\usepackage{setspace}
\usepackage[ruled,norelsize]{algorithm2e}

\definecolor{commentcolor}{RGB}{110,154,155}

\usepackage{booktabs}
\usepackage{makecell}

\usepackage{tabularx}
\usepackage{longtable}
\usepackage[dvipsnames,table]{xcolor}

\usepackage[many]{tcolorbox}
\usepackage{arydshln}
\newtcolorbox{conclusionbox}{
    enhanced,
    colback=lightblue,
    colframe=teal!70!black,
    boxrule=0.8pt,
    arc=4pt,
    drop shadow=gray!50,
    left=6pt, right=6pt,
    top=4pt, bottom=4pt,
    parbox=false
}

\def\dymethod{PUMA\xspace}
\def\dybench{DOMINO\xspace}
\def\dybenchwo{DOMINO}

\begin{document}

\title{Towards Generalizable Robotic Manipulation \\in Dynamic Environments} 

\titlerunning{Towards Generalizable Robotic Manipulation in Dynamic Environments}

\author{Heng Fang\inst{1} \and
Shangru Li\inst{1} \and
Shuhan Wang\inst{1} \and
Xuanyang Xi\inst{2} \and \\
Dingkang Liang\inst{1}\textsuperscript{\Letter} \and
Xiang Bai\inst{1}
}

\authorrunning{H.~Fang et al.}

\institute{Huazhong University of Science and Technology, China \and
Huawei Technologies Co. Ltd, China\\
\footnotesize
\quad \Letter~Corresponding Author.
\\
\email{\{hengfang,dkliang\}@hust.edu.cn}
\\
\footnotesize
\textbf{Project Page: \url{https://H-EmbodVis.github.io/DOMINO/}}\\
\footnotesize
\textbf{Code: \url{https://github.com/H-EmbodVis/DOMINO}}
}

\maketitle
\vspace{-1.5em}

\begin{abstract}
    Vision-Language-Action (VLA) models excel in static manipulation but struggle in dynamic environments with moving targets. 
    This performance gap primarily stems from a scarcity of dynamic manipulation datasets and the reliance of mainstream VLAs on single-frame observations, restricting their spatiotemporal reasoning capabilities. To address this, we introduce \textbf{\dybench}, a large-scale dataset and benchmark for generalizable dynamic manipulation, featuring 35 tasks with hierarchical complexities, over 110K expert trajectories, and a multi-dimensional evaluation suite. 
    Through comprehensive experiments, we systematically evaluate existing VLAs on dynamic tasks, explore effective training strategies for dynamic awareness, and validate the generalizability of dynamic data. 
    Furthermore, we propose \textbf{\dymethod}, a dynamics-aware VLA architecture. By integrating scene-centric historical optical flow and specialized world queries to implicitly forecast object-centric future states, \dymethod couples history-aware perception with short-horizon prediction. Results demonstrate that \dymethod achieves state-of-the-art performance, yielding a 6.3\% absolute improvement in success rate over baselines. 
    Moreover, we show that training on dynamic data fosters robust spatiotemporal representations that transfer to static tasks.
\end{abstract}

\section{Introduction}

\begin{figure}[t]
    \centering
    \includegraphics[width=1.\textwidth]{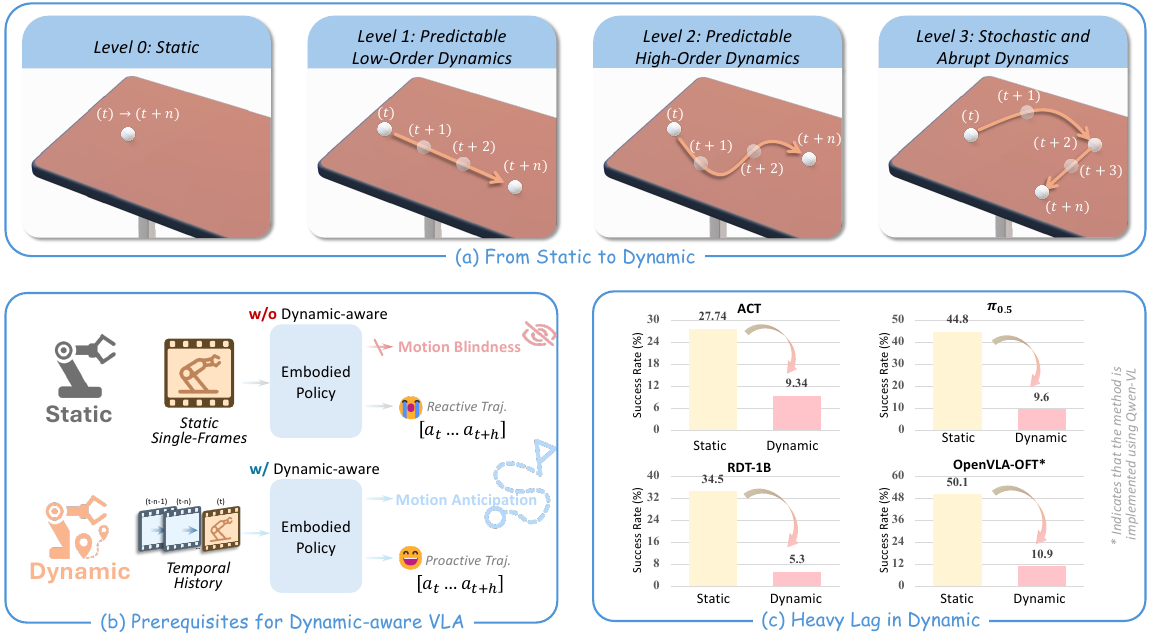}
    \vspace{-15pt}
    \caption{
    \textbf{(a)} Illustration of the defined dynamic difficulty levels, progressing from static (Level 0) to stochastic and abrupt dynamics (Level 3). 
    \textbf{(b)} Dynamic awareness requires capturing historical context and anticipating future motion.
    \textbf{(c)} Performance of SOTA models degrades when shifting from static to dynamic environments.
    }
    \vspace*{-1.8em}
    \label{fig:intro}
\end{figure}

Recent advancements in Vision-Language-Action (VLA) models have demonstrated significant success in handling manipulation tasks, establishing a foundation for generalizable embodied intelligence~\cite{bjorck2025gr00t,wang2025vlaadapter,2025openhelix,shi2025memoryvla}. 
However, most existing VLA models~\cite{kim2024openvla,kim2025openvlaoft,pi0,pi05} focus on static manipulation, which involves interacting with stationary objects in stable environments.
In contrast, dynamic manipulation requires robots to adapt to moving objects and continuous environmental changes. 
Mastering this capability is essential for deploying robots in complex real-world scenarios, such as operating on active assembly lines or working alongside humans.
Despite its importance, dynamic manipulation remains a critical yet underexplored frontier. 
Dynamic manipulation is inherently difficult as it imposes strict requirements on spatiotemporal precision and necessitates the continuous integration of real-time perception with motion prediction.
Progress in this domain is hindered by two primary factors: \textbf{1)} the scarcity of large-scale dynamic manipulation datasets, and \textbf{2)} the limited dynamic perception and motion prediction capabilities of existing mainstream VLA architectures.

High-quality dynamic data is indispensable for achieving generalizable dynamic manipulation. 
Addressing the current data scarcity is non-trivial due to the inherent complexity of dynamic tasks.
Constructing dynamic environments for data collection presents significant challenges, primarily due to the strict spatiotemporal synchronization required between moving targets and robotic actions.
Furthermore, acquiring expert demonstrations in such unpredictable settings is difficult, as human teleoperators or scripted policies often struggle to react precisely to continuous environmental changes. 
Consequently, most existing embodied datasets are confined to stationary tasks~\cite{o2024open,brohan2022rt,walke2023bridgedata}. 
Therefore, developing a scalable, low-cost pipeline to acquire diverse manipulation data in dynamic scenarios is a critical imperative.

To bridge this gap, we introduce a comprehensive benchmark tailored for generalizable dynamic manipulation, termed the \textbf{D}ynamic \textbf{O}bject \textbf{M}an\textbf{I}pulatio\textbf{N} \textbf{O}perations Benchmark (\textbf{\dybench}). 
Building upon RoboTwin 2.0~\cite{chen2025robotwin} and the SAPIEN~\cite{Xiang_2020_SAPIEN} simulation engine, we develop scalable pipelines for dynamic manipulation dataset generation and closed-loop evaluation.
The benchmark comprises 35 diverse dynamic tasks across 5 distinct robot embodiments, spanning from single-arm operations to complex dual-arm collaborations. 
As shown in \fig{fig:intro}(a), these tasks are organized into a three-tiered difficulty hierarchy, progressing from predictable low-order dynamics to high-order nonlinear trajectories and stochastic scenarios with abrupt disturbances. 
To precisely control task complexity, we parameterize the overall motion speed using a scalar dynamics coefficient $\alpha$, denoting each setting as \emph{\dybenchwo@$\alpha$}. 
Crucially, these tasks present significant challenges for current VLA models. While existing VLAs excel at static manipulation, they struggle in dynamic environments due to an inherent lack of continuous spatiotemporal reasoning. The precise timing required to interact with moving targets and handle unexpected disturbances often exceeds the capabilities of current architectures, which are typically constrained by static spatial biases.
To rigorously assess generalization, our dataset provides over 110K expert trajectories collected under both canonical and domain-randomized settings. 
Finally, our closed-loop pipeline establishes a multi-dimensional evaluation protocol that extends beyond standard binary success rates.

Evaluations of state-of-the-art models~\cite{zhao2023act,liu2024rdt,kim2024openvla,pi0} on this benchmark reveal substantial performance degradation in dynamic settings, as shown in \fig{fig:intro}(c). 
We attribute this decline to the reliance of standard VLA architectures on single-frame observations, which limits their dynamic awareness.
We suggest that this awareness requires \textit{capturing historical context and anticipating future motion}. Although recent VLAs incorporate world models~\cite{zhang2025dreamvla,ye2025dream,cen2025worldvla}, they primarily model global scene transitions or robot kinematics, neglecting the individual object dynamics crucial for spatiotemporal anticipation.
To address this, we present \dymethod as an exploratory attempt to bridge this gap. 
By integrating historical frames and motion cues to model scene-centric historical dynamics, our method introduces specialized predictive queries to implicitly infer object-centric dynamics for moving targets. 
This design endows the model with a dynamic understanding of the physical world, enabling anticipatory interactions with moving objects and yielding a 6.3\% SR improvement.

In summary, the main contributions are as follows:
\textbf{1)} 
We systematically analyze dynamic manipulation, distinguishing its unique spatiotemporal challenges from static paradigms to underscore the need for advancing dynamic embodied intelligence.
\textbf{2)} 
We introduce \dybench, a scalable pipeline for dynamic manipulation dataset generation and closed-loop evaluation. It features diverse robot embodiments, multi-tiered difficulty scaling via a dynamics coefficient, and a comprehensive multi-dimensional metric suite.
\textbf{3)} 
We propose \dymethod, which integrates historical motion cues to enhance motion anticipation. By employing specialized predictive queries to implicitly infer the future states of moving objects, our approach yields a 6.3\% performance improvement over baselines.
\vspace{-1em}
\section{\dybench Dataset}
\label{sec:dynamic_task}
\vspace{-0.5em}

\subsection{Task Definition}
\label{subsec:task_formulation}
We formulate generalizable dynamic manipulation as a Partially Observable Markov Decision Process (POMDP)~\cite{kaelbling1998planning}. At time step $t$, the full state $s_t = \{s_t^r, s_t^o\}$ comprises the robot proprioception $s_t^r$ and the physical object state $s_t^o$. Due to partial observability, the policy relies on an observation history $o_{t-h:t}$, where each observation $o_t = \{I_t, s_t^r\}$ includes high-dimensional visual inputs $I_t$ (e.g., RGB-D) and proprioception.
The continuous action $\mathbf{a}_t \in \mathcal{A}$ specifies the dual-arm control commands. In dynamic environments, the transition dynamics $\mathcal{T}(s_{t+1} | s_t, \mathbf{a}_t)$ are inherently time-varying, governed by both the independent motion of the object and the interaction dynamics induced by robot contact.

Our objective is to learn a policy $\pi_\phi(\mathbf{a}_t | o_{t-h:t})$ that minimizes the expected finite-horizon cost:
\begin{equation}
J(\phi) = \mathbb{E} \left[ \sum_{k=0}^{H-1} \gamma^k \ell(s_{t+k}, \mathbf{a}_{t+k}) \right],
\end{equation}
where $\gamma \in [0, 1)$ is the discount factor, subject to safety constraints. The cost $\ell(\cdot)$ penalizes the spatial discrepancy between the end-effectors and the object, alongside the control effort. To achieve precise spatiotemporal interception, $\pi_\phi$ must implicitly anticipate future object states from historical observations.
\vspace{-1em}

\begin{figure}[t]
    \centering
    \includegraphics[width=1.\textwidth]{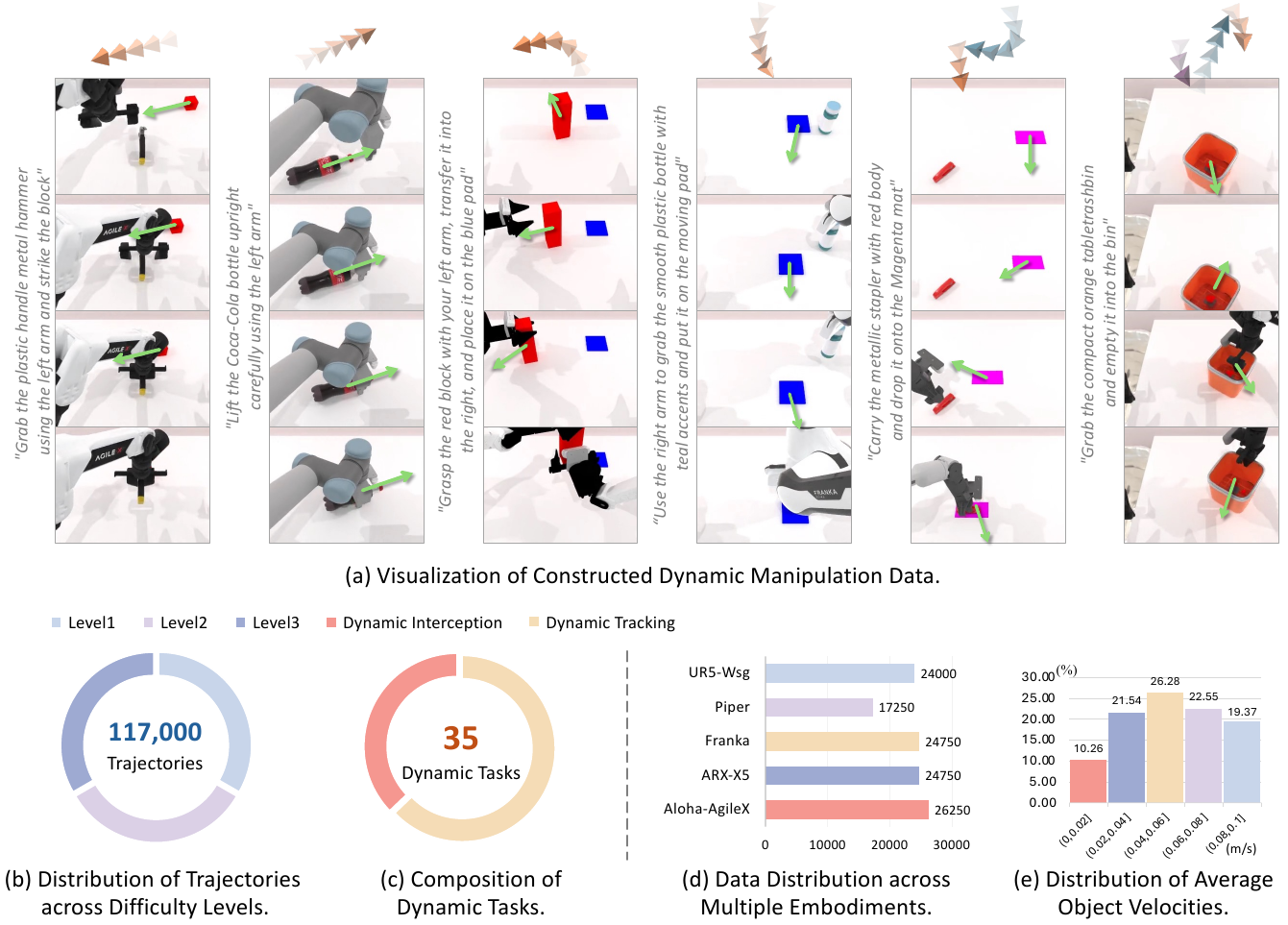}
    \vspace{-15pt}
    \caption{
    \textbf{Dataset Visualization}. We present \dybench dataset of 117,000 dynamic manipulation trajectories, covering 35 distinct tasks across five robot embodiments.
    }
    \vspace*{-1.5em}
    \label{fig:dataset}
\end{figure}

\subsection{Data Construction}
\label{subsec:dyman_construct}

To systematically evaluate generalizable dynamic manipulation, we introduce \dybench. 
Unlike existing datasets focused on stationary tasks, \dybench provides a scalable, low-cost pipeline for generating diverse dynamic data. 
As shown in ~\fig{fig:dataset}, it comprises 35 dynamic tasks across five robot embodiments. 
To facilitate rigorous generalization assessments, we provide over 110K expert trajectories collected under both canonical and domain-randomized settings.

Constructing dynamic environments poses a significant challenge, primarily due to the strict spatiotemporal synchronization required between moving targets and robotic actions. Building upon the SAPIEN~\cite{Xiang_2020_SAPIEN} physics engine and the RoboTwin 2.0~\cite{chen2025robotwin} framework, our data generation pipeline introduces a three-stage spatiotemporal synchronization method to reliably acquire expert demonstrations.
First, in the temporal dry-run phase, we randomly sample the manipulation pose of the target object and execute the task in a static environment to record the exact execution time required by the robot arms. Second, in the kinematic back-calculation phase, we reverse-engineer the object's initial spatial position based on the recorded execution time and specified motion trajectory. Third, in the synchronized dynamic execution phase, target objects are instantiated as kinematic bodies within SAPIEN and the robot replays its motion plan alongside the moving target. This ensures stable and predictable motion execution, immune to unintended physical disturbances. To guarantee expert data quality, we implement specialized adaptations for complex objects and enforce strict dynamic task-success criteria during automated generation.

\vspace{-1em}
\subsection{Data Characteristics}
\label{subsec:dyman_character}
\dybench categorizes dynamic manipulation through a spatiotemporal task taxonomy, hierarchical motion complexities, and comprehensive evaluation metrics.

\vspace{-0.5em}
\paragraph{\textbf{\emph{Spatiotemporal Task Taxonomy.}}}
To decouple the core challenges of dynamic manipulation from task-specific details, we divide the 35 benchmark tasks into two functional categories based on interaction requirements: dynamic interception and dynamic tracking.
Dynamic interception focuses on instantaneous target acquisition, where the agent transitions from free space to establish contact with a moving target. This evaluates predictive kinematics and latency compensation. For example, catching a thrown object requires precise trajectory planning to reach an optimal interception point. Isolating this discrete action assesses the agent's spatial precision under strict temporal constraints.
In contrast, dynamic tracking involves continuous synchronization, requiring the agent to maintain a consistent spatial relationship with a moving target over a time window $\Delta\tau$. This evaluates real-time closed-loop control and velocity-matching capabilities, such as placing an object into a box on a conveyor belt. Consequently, this formulation assesses agent's capacity for sustained error correction and trajectory adjustment during continuous interactions.

\vspace{-0.5em}
\paragraph{\textbf{\emph{Hierarchical Dynamic Complexity.}}}
To evaluate agents across diverse dynamic complexities, ranging from predictable tracking to reactive adaptation, we categorize the motion dynamics $\mathcal{F}$ into three levels, as illustrated in~\fig{fig:intro}(a):
\begin{itemize}
\vspace{-0.5em}
\item Level 1 (Predictable Low-Order Dynamics): Objects move with a constant velocity $\mathbf{v} \sim \mathcal{U}(v_{\min}, v_{\max})$. This zero-curvature trajectory serves as a foundational test for instantaneous state estimation and linear extrapolation.
\item Level 2 (Predictable High-Order Dynamics): Trajectories follow a polynomial curve $\mathbf{x}(t) = \sum_{k=0}^{n} \mathbf{b}_k (t/T_{traj})^k$ of degree $n \in [2, 5]$, fit to randomly sampled spatial control points. This introduces variable curvature and acceleration, requiring the agent to aggregate historical observations to implicitly model the physical dynamics.
\item Level 3 (Stochastic and Abrupt Dynamics): Trajectories comprise $s \in [2,3]$ independent segments of Level 1/2 dynamics, with segment durations drawn from a Dirichlet distribution. Velocity and acceleration are typically discontinuous at segment transitions. This unpredictability evaluates reactive robustness, compelling agent to rely on high-frequency closed-loop feedback.
\end{itemize}

\vspace{-0.8em}
\paragraph{\textbf{\emph{Comprehensive Evaluation Metrics.}}}
To assess robustness across dynamic difficulties, we introduce the benchmark variant \dybenchwo@$\alpha$. This variant parameterizes the maximum target speed in meters per second using a scalar coefficient $\alpha \ge 0$. For instance, $\alpha=0.1$ denotes a maximum speed of $0.1$ m/s, and $\alpha=0$ represents a static setting.
Under this setting, we measure the Success Rate (SR), defined as the percentage of episodes that satisfy all task conditions within a time budget $T_{\max}$. 
As SR alone is insufficient for stochastic environments, we introduce the Manipulation Score (MS), a continuous metric designed to capture execution quality. 
The MS consists of a base Route Completion ($RC$) score adjusted by penalty factors. $RC$ quantifies spatial convergence via a progress ratio $\rho=1-\|\mathbf{p}_{\text{ee}}^{(T_{end})}-\mathbf{p}_{\text{obj}}^{(T_{end})}\|_2/\|\mathbf{p}_{\text{ee}}^{(0)}-\mathbf{p}_{\text{obj}}^{(T_{end})}\|_2$, where $\mathbf{p}_{\text{ee}}$ and $\mathbf{p}_{\text{obj}}$ denote positions of the end-effector and target object at initial $(0)$ and final $(T_{end})$ timesteps. For dual-arm setups, $RC \in [0, 100]$ reflects the maximum progress of either arm, computed as $100\times\max(\rho_{\text{left}},\rho_{\text{right}})$, with successful episodes assigned 100. 
Finally, to penalize unsafe behaviors, the overall MS is calculated by multiplying $RC$ by 0.5 if the target exits the safe workspace or field of view, and by 0.8 upon collision with environmental clutter.

\vspace{-0.8em}
\section{Dynamic-Aware VLA}
\label{sec:dynamic_aware_vla}
\vspace{-0.5em}

\begin{figure}[t]
    \centering
    \includegraphics[width=0.9\textwidth]{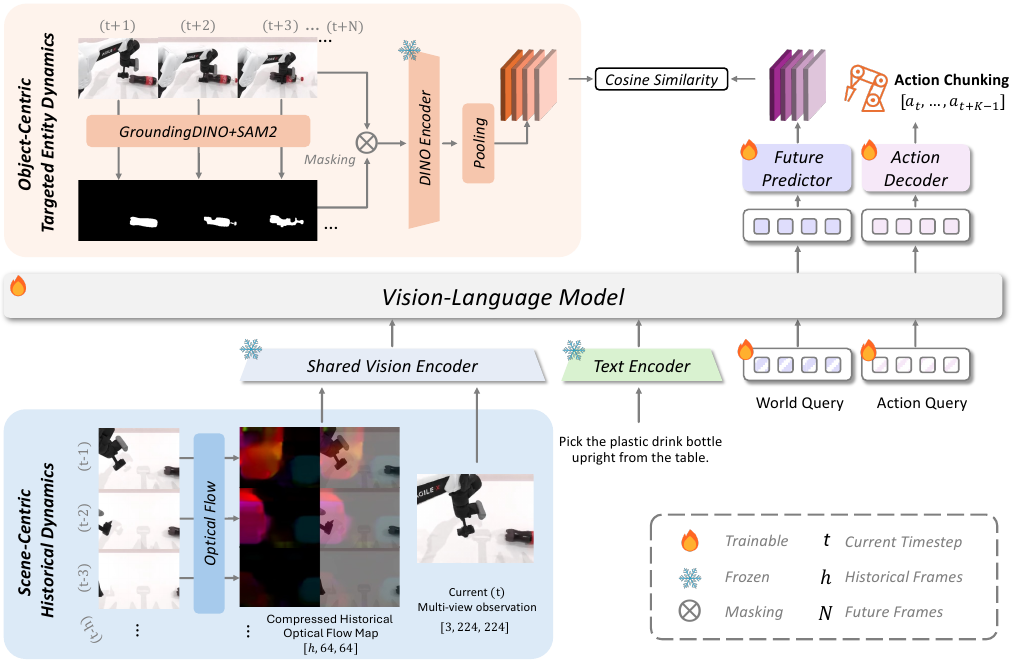}
    \vspace{-5pt}
    \caption{
    \dymethod processes historical motion flows, current observations, and instructions to encode scene-centric historical dynamics. It employs a dual-query mechanism where Action Queries decode continuous action chunks and World Queries aggregate dynamic representations. During training, world queries are supervised via a similarity loss against future features extracted by DINO to predict object-centric dynamics.
    }
    \vspace*{-1.5em}
    \label{fig:method}
\end{figure}

To achieve generalizable dual-arm manipulation in dynamic environments, we propose the \textbf{P}redictive \textbf{U}nified \textbf{M}anipulation \textbf{A}rchitecture (\textbf{\dymethod}) as shown in~\fig{fig:method}.
Dynamic awareness requires capturing the historical context of objects and anticipating their future motion.
Therefore, \dymethod couples scene-centric history-aware perception with short-horizon object-centric prediction to satisfy spatiotemporal constraints induced by object dynamics $\mathcal{F}$ within Qwen3-VL~\cite{Qwen3-VL}.
Given an observation history $o_{t-h:t}$ of length $h{+}1$ and a language instruction $l$, the model $M_\theta$ uses a shared backbone to jointly optimize action policy $\pi_\phi$ and an auxiliary future feature predictor $\psi_\omega$.
Specifically, it outputs an action chunk $\hat{\mathbf{a}}_{t:t+K-1}$ of length $K$ in a single forward pass~\cite{zhao2023act} alongside auxiliary features $\mathbf{z}_{t+1:t+N}$ of horizon $N$ encoding future object motion.
Crucially, the auxiliary future predictor is supervised only during training.
This encourages shared representation to anticipate the dynamics $\mathcal{F}$ and effectively regularizes policy without adding computational overhead during inference.

\vspace{-1em}
\subsection{Scene-Centric Spatiotemporal Dynamics Encoding}
\label{subsec:spatiotemporal_encoding}
\vspace{-0.5em}
To effectively operate in dynamic environments, our model requires a comprehensive understanding of both spatial context and temporal evolution. Therefore, the input space comprises three key components: a long-term language instruction, current multi-view observations, and a historical dynamic context.
The language instruction $l$ acts as the overarching task specification, guiding the manipulation policy. Concurrently, the current multi-view images serve as spatial visual prompts, providing dense observations of the immediate workspace.
Capturing historical scene dynamics is crucial for reacting to moving objects. 

To efficiently encode this dynamics, we introduce a compact dynamic context representation.
We sample $h$ historical third-person frames at a fixed stride and apply a spatial compression operator. 
Instead of directly stacking raw historical frames, which forces the network to implicitly deduce temporal changes, we compute optical flow maps across these compressed frames. Optical flow provides intuitive and explicit motion states, making it significantly easier for the policy to learn dynamic patterns.
These flow maps are processed alongside the current multi-view visual input through the Qwen3-VL visual encoder, forming the observation history $o_{t-h:t}$ that supplies the model with explicit dense motion cues, enabling the policy to accurately estimate object motion trends.

\vspace{-0.7em}
\subsection{Object-Centric Dynamic Representation}
\vspace{-0.3em}
To interact in dynamic environments, the policy must anticipate the future motion of target objects.
While existing frameworks~\cite{zhang2025dreamvla} forecast scene-level dynamic regions, we posit that generalizable dynamic manipulation requires an object-centric focus, isolating the target's trajectory from irrelevant scene dynamics.
To construct supervision for this capability, we sample $N$ future frames $I_{t+1:t+N}$ at a fixed interval during training and isolate the target object using a frozen grounding module~\cite{ren2024grounded}.
Specifically, the manipulated object parsed from the language instruction serves as a text prompt $p$ for GroundingDINO~\cite{liu2023grounding}, which generates a bounding box, then processed by SAM2~\cite{ravi2024sam2segmentimages} to yield a segmentation mask.
Let $\mathcal{B}(I,p)$ denote the binary mask, $\mathcal{E}(I)$ the frozen DINO patch-token encoder, and $\mathcal{P}(\cdot,\cdot)$ masked average pooling.
The object-centric future feature $\mathbf{f}_{t+i}$ is computed as:
\begin{equation}
    \mathbf{f}_{t+i}=\mathcal{P}\!\left(\mathcal{E}(I_{t+i}), \mathcal{B}(I_{t+i}, p)\right), \quad i=1,\dots,N.
\end{equation}
To model these future states, we introduce $N$ learnable world queries that aggregate the spatiotemporal context to predict the target's future representations within the latent space. 
We then optimize these predictions by enforcing a similarity loss against the extracted ground-truth DINO features. This explicit supervision forces the latent world representation to capture and anticipate the underlying object dynamics.
This supervision is applied only during training, requiring no future frames at inference.

\vspace{-0.5em}
\subsection{Training Strategy}
We train the unified model $M_\theta$ end-to-end using supervised behavioral cloning combined with an auxiliary future-feature prediction objective. Each training tuple comprises $\{o_{t-h:t}, l, \mathbf{f}_{t+1:t+N}, \mathbf{a}^*_{t:t+K-1}\}$.
The action policy is supervised via an $\ell_1$ regression loss over the predicted action chunk:
\begin{equation}
    \mathcal{L}_{action} = \frac{1}{K} \sum_{i=0}^{K-1} \| \hat{\mathbf{a}}_{t+i} - \mathbf{a}^*_{t+i} \|_1.
\end{equation}
Simultaneously, the auxiliary future predictor is optimized by minimizing the cosine distance between the predicted representations $\mathbf{z}_{t+i}$ and the ground-truth object-centric features $\mathbf{f}_{t+i}$:
\begin{equation}
    \mathcal{L}_{world} = \frac{1}{N}\sum_{i=1}^{N}\left(1-\frac{\mathbf{z}_{t+i}^{\top}\mathbf{f}_{t+i}}{\|\mathbf{z}_{t+i}\|_2\,\|\mathbf{f}_{t+i}\|_2}\right).
\end{equation}
Overall training objective is formulated as a weighted sum of two components:
\begin{equation}
    \mathcal{L}_{total} = \mathcal{L}_{action} + \lambda\,\mathcal{L}_{world},
\end{equation}
where $\lambda$ is a balancing hyperparameter controlling influence of dynamics task.

\vspace{-0.5em}

\section{Experiment}
\label{sec:experiments}
All VLA models are trained on NVIDIA A100 GPUs, while data generation and evaluation are performed on NVIDIA RTX GPUs.
Appendix provides additional implementation details and hyperparameters.
To balance broad experimental coverage with limited compute, unless otherwise stated, all experiments are conducted in clean setting (plain backgrounds, no domain randomization) with the Aloha-AgileX robot, using a training mixture of 35 dynamic tasks under Level 1 dynamics with dynamic coefficient $\alpha=0.1$.

\subsection{Experimental Setup}
\label{subsec:exp_setup}

\paragraph{\emph{\textbf{Benchmarks.}}}
We primarily evaluate on our proposed \dybenchwo$@$0.1 benchmark, reporting the Success Rate (SR) and Manipulation Score (MS). Additionally, we provide selected results in static environments using RoboTwin 2.0~\cite{chen2025robotwin}.

\vspace{-0.5em}
\paragraph{\emph{\textbf{Baselines.}}}
We evaluate \dymethod against the standard policy learning framework ACT~\cite{zhao2023act} and several state-of-the-art VLA models, including OpenVLA~\cite{kim2024openvla}, OpenVLA-OFT~\cite{kim2025openvlaoft}, RDT~\cite{liu2024rdt}, $\pi_0$~\cite{pi0}, $\pi_{0.5}$~\cite{pi05}, $\pi_0$-FAST~\cite{pertsch2025fast}, VLA-Adapter~\cite{wang2025vlaadapter}, InternVLA-M1~\cite{internvlam1}, and Qwen3-VL-based VLAs~\cite{starvla2025}. For a fair comparison, all baselines are fine-tuned on proposed \dybench dataset. 
Note that while the VLA models are fine-tuned across all tasks, ACT is fine-tuned on a per-task basis.

\vspace{-1em}
\subsection{Challenges of the Proposed \dybench Dataset}
\label{subsec:exp_challenge}
\vspace{-0.5em}

\begin{table*}[ht]
\centering
\vspace{-2em}
\scriptsize
\setlength{\tabcolsep}{0.5mm}
\caption{Quantitative evaluation of VLA models in static and our proposed \dybenchwo@$0.1$ ~under Level 1. $X\to Y$ denotes training in environment $X$ and testing in $Y$ (S: static, D: dynamic), with ZS representing zero-shot evaluation and FT indicating fine-tuning. Subscripts denote performance $\Delta$ (ZS vs. Static; FT vs. ZS).}
\label{tab:vla_in_sd}
\vspace{-10pt}
\begin{tabular}{l *{9}{C{9mm}}}
\toprule
\multirow{2}{*}{\textbf{Simulation Task}}
  & \multicolumn{3}{c}{\textbf{ACT~\cite{zhao2023act}}}
  & \multicolumn{3}{c}{\textbf{OpenVLA-OFT~\cite{kim2025openvlaoft}}}
  & \multicolumn{3}{c}{$\mathbf{\pi}_{\mathbf{0.5}}$~\cite{pi05}} \\
  \cmidrule(lr){2-4} \cmidrule(lr){5-7} \cmidrule(lr){8-10}
& \shortstack{Static \\[-0.5ex] \tiny S$\to$S} 
& \shortstack{ZS \\[-0.5ex] \tiny S$\to$D} 
& \shortstack{FT \\[-0.5ex] \tiny D$\to$D}  
& \shortstack{Static \\[-0.5ex] \tiny S$\to$S} 
& \shortstack{ZS \\[-0.5ex] \tiny S$\to$D} 
& \shortstack{FT \\[-0.5ex] \tiny D$\to$D}  
& \shortstack{Static \\[-0.5ex] \tiny S$\to$S} 
& \shortstack{ZS \\[-0.5ex] \tiny S$\to$D} 
& \shortstack{FT \\[-0.5ex] \tiny D$\to$D} \\
\midrule
\textit{Adjust Bottle} & 97\% & 37\% & 65\% & 47\% & 16\% & 58\% & 92\% & 12\% & 52\% \\
\textit{Click Alarmclock} & 29\% & \phantom{0}3\% & \phantom{0}8\% & 57\% & \phantom{0}9\% & \phantom{0}6\% & 48\% & \phantom{0}8\% & 14\% \\
\textit{Grab Roller} & 95\% & 24\% & 23\% & 47\% & 24\% & 28\% & 98\% & \phantom{0}6\% & 10\% \\
\textit{Handover Block} & 43\% & \phantom{0}0\% & 11\% & \phantom{0}3\% & \phantom{0}3\% & \phantom{0}9\% & \phantom{0}2\% & \phantom{0}0\% & \phantom{0}1\% \\
\textit{Handover Mic} & 82\% & 13\% & 23\% & 93\% & \phantom{0}8\% & 25\% & 63\% & \phantom{0}7\% & \phantom{0}5\% \\
\textit{Move Can Pot} & 23\% & \phantom{0}0\% & 26\% & 34\% & 29\% & 29\% & 14\% & \phantom{0}5\% & \phantom{0}7\% \\
\textit{Move Pillbottle Pad} & \phantom{0}1\% & \phantom{0}0\% & \phantom{0}1\% & \phantom{0}1\% & \phantom{0}0\% & \phantom{0}1\% & 20\% & \phantom{0}5\% & \phantom{0}6\% \\
\textit{Place Bread Skillet} & \phantom{0}9\% & \phantom{0}5\% & 12\% & \phantom{0}4\% & \phantom{0}1\% & \phantom{0}1\% & 39\% & \phantom{0}5\% & \phantom{0}9\% \\
\textit{Place Fan} & \phantom{0}0\% & \phantom{0}0\% & \phantom{0}1\% & \phantom{0}3\% & \phantom{0}0\% & \phantom{0}1\% & 30\% & \phantom{0}1\% & \phantom{0}2\% \\
\textit{Place Object Basket} & 18\% & 19\% & 21\% & 19\% & 16\% & 18\% & 57\% & \phantom{0}6\% & 11\% \\
\textit{......(35 tasks)} &&&&&&&&& \\
\textit{Press Stapler} & 31\% & \phantom{0}2\% & \phantom{0}4\% & 35\% & \phantom{0}2\% & \phantom{0}5\% & 35\% & \phantom{0}1\% & \phantom{0}6\% \\
\textit{Rotate QRcode} & \phantom{0}3\% & \phantom{0}0\% & \phantom{0}0\% & 10\% & \phantom{0}8\% & 11\% & 54\% & \phantom{0}2\% & \phantom{0}4\% \\
\textit{Scan Object} & \phantom{0}1\% & \phantom{0}2\% & \phantom{0}2\% & \phantom{0}0\% & \phantom{0}0\% & \phantom{0}4\% & \phantom{0}7\% & \phantom{0}0\% & \phantom{0}1\% \\
\textit{Shake Bottle} & 74\% & 18\% & 14\% & 63\% & 12\% & 17\% & 95\% & 19\% & 33\% \\
\textit{Shake Bottle Horiz.} & 63\% & 23\% & 18\% & 57\% & 19\% & 31\% & 97\% & 24\% & 42\% \\
\midrule
\textbf{\textit{Average (\%)}} & 27.7 & \phantom{0}6.5\textsubscript{\color{red}{-21.2}} & \phantom{0}9.4\textsubscript{\color{kaiming-green}{+2.9}} & 17.5 & \phantom{0}6.7\textsubscript{\color{red}{-10.8}} & \phantom{0}9.1\textsubscript{\color{kaiming-green}{+2.4}} & 44.8 & \phantom{0}7.5\textsubscript{\color{red}{-37.3}} & \phantom{0}9.6\textsubscript{\color{kaiming-green}{+2.1}} \\
\bottomrule
\end{tabular}
\vspace{-1.7em}
\end{table*}

\begin{wrapfigure}[13]{r}{0.4\textwidth}
    \vspace*{-2.5em}
    \centering
    \includegraphics[width=.4\textwidth]{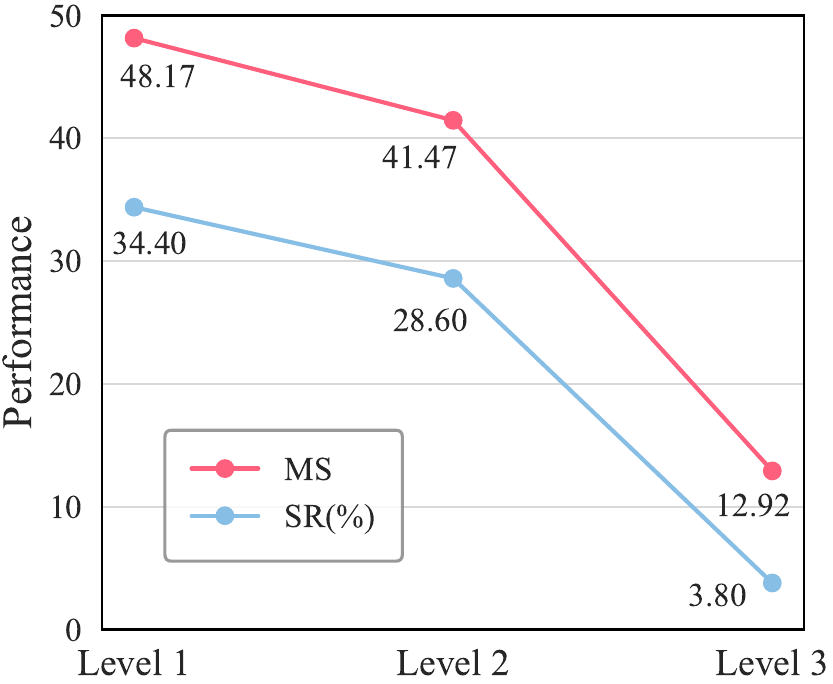}
    \vspace{-15pt}
    \caption{
    Performance degradation of the ACT model across three dynamic complexity (Level 1-3).
    }
    \vspace*{-1.5em}
    \label{fig:level_perf}
\end{wrapfigure}

To investigate the challenges of dynamic environments, we evaluate representative VLA architectures in both static and dynamic settings. As shown in~\tab{tab:vla_in_sd}, these models exhibit satisfactory performance in static scenarios (S$\to$S), but their performance degrades significantly in dynamic environments (S$\to$D). Under identical task settings with moving targets, the average success rate drops drastically. For instance, the performance of $\pi_{0.5}$~\cite{pi05} falls from 44.8\% to 7.5\%.

Furthermore, we explore whether fine-tuning these baselines with dynamic data (D$\to$D) can bridge this gap. We observe only marginal improvements, with average success rates increasing by less than 3\%. 
We attribute this bottleneck to fundamental architectural limitations rather than mere data distribution shifts. 
Standard VLA architectures rely on single-frame observations, inherently limiting their dynamic awareness.
We argue that effective dynamic manipulation requires \textit{capturing historical object context and anticipating future motion}. The current VLA paradigm lacks these essential dynamic modeling capabilities, resulting in sub-optimal dynamic manipulation performance.

While our primary experiments focus on Level 1 dynamics, we also investigate the impact of higher dynamic complexities. We train ACT models per-task on five tasks and evaluate them across all three dynamic levels. As illustrated in~\fig{fig:level_perf}, performance deteriorates rapidly as complexity increases. Level 2 and Level 3 environments are significantly more challenging than Level 1, highlighting the need for stronger dynamic modeling architectures in future research.

\begin{conclusionbox}
    \textit{\textbf{Finding 1}: Dynamic manipulation presents a challenging new frontier.} 
    The transition from static to dynamic environments introduces complex spatiotemporal challenges that fundamentally degrade the reliability of existing paradigms. It represents a distinct and demanding domain that cannot be solved by merely scaling static manipulation approaches.
\end{conclusionbox}

\begin{table}[t]
\setlength{\tabcolsep}{2.5mm}
\centering
\scriptsize
\caption{Results of explicitly modeling future trajectories on Level~1 \dybenchwo@$0.1$.}
\label{tab:exp_oracle}
\vspace{-10pt}
\begin{tabular}{ llcC{11mm}C{11mm} }
\toprule
Method & Settings & w/ GT Traj. & \makecell{SR (\%)$\uparrow$} & \makecell{MS$\uparrow$} \\
\midrule
\multirow{3}{*}{OpenVLA-OFT~\cite{kim2025openvlaoft}} & Static & $\times$ & 17.51 & 17.51 \\
 & Dynamic & $\times$ & 9.06 & 24.06 \\
 & Dynamic & \checkmark & 10.33 & 32.00 \\
\bottomrule
\end{tabular}
\vspace{-1.5em}
\end{table}

To validate this hypothesis and isolate the performance bottleneck of current VLAs, we conduct an oracle experiment, as detailed in~\tab{tab:exp_oracle}. We introduce an auxiliary dynamic ground-truth encoder, implemented as an MLP. This encoder takes the target object's future pose window combined with a continuous trajectory parameter vector as input to generate a dynamic conditional representation. During the forward pass, this conditional vector is concatenated with the action queries and passed through a subsequent MLP layer. 
Experimental 
results reveal a nuanced behavior: while the explicit introduction of future trajectories yields only a marginal improvement in overall SR, it significantly boosts MS. Qualitative analysis of the evaluation rollouts indicates that the policy successfully acquires and tracks the correct trajectory but exhibits control jitter and temporal inconsistency during the actual manipulation phase. We attribute this to the lack of historical frame observations; without historical context, the model fails to comprehend the target's underlying physical dynamics. Instead, the policy overfits to naive trajectory following, which inadvertently interferes with closed-loop manipulation learning. However, in successful episodes, the introduction of ground-truth trajectories yields exceptionally high manipulation quality that approaches expert demonstrations, confirming the potential of future spatial cues if appropriately grounded in physical context.

\begin{conclusionbox}
    \textit{\textbf{Finding 2}: Naive future trajectory injection is insufficient for dynamic manipulation.}
    While providing future spatial cues improves motion tracking, the absence of historical observations prevents the model from understanding physical dynamics, leading to control instability during manipulation. Effective dynamic manipulation requires a holistic integration of both historical context and future anticipation.
\end{conclusionbox}

\vspace{-1em}
\subsection{Effectiveness of Dynamic-Aware VLA}
\label{subsec:exp_method}

\begin{table}[t]
\setlength{\tabcolsep}{3.5mm}
\centering
\scriptsize
\caption{Comparison with SOTA methods on \dybenchwo@$0.1$ (Level 1 dynamics, clean setting, Aloha-AgileX robot). All VLA models are fine-tuned on 35 dynamic tasks.}
\label{tab:main_result}
\vspace{-10pt}
\begin{tabular}{ l c l C{11mm} C{11mm} }
\toprule
Method & Venue & LLM & \makecell{SR (\%)$\uparrow$} & \makecell{MS$\uparrow$} \\
\midrule
OpenVLA~\cite{kim2024openvla} & CoRL 24 & Prismatic & 1.54 & 6.10 \\
RDT-1B~\cite{liu2024rdt} & ICLR 25 & DiT & 5.34 & 17.71 \\
$\mathbf{\pi}_{\mathbf{0}}$~\cite{pi0} & RSS 25 & PaliGemma & 8.17 & 23.96 \\
$\mathbf{\pi}_{\mathbf{0.5}}$~\cite{pi05} & CoRL 25 & PaliGemma & 9.63 & 26.17 \\
InternVLA-M1~\cite{internvlam1} & arXiv 25 & Qwen2.5-VL & 5.40 & 27.57 \\
Isaac-GR00T~\cite{bjorck2025gr00t} & arXiv 25 & Qwen3-VL* & 6.10 & 28.60 \\
VLA-Adapter~\cite{wang2025vlaadapter} & AAAI 26 & Qwen2.5-VL & 4.40 & 24.31 \\
\midrule
\multirow{2}{*}{$\mathbf{\pi}_{\mathbf{0}}$-FAST~\cite{pertsch2025fast}} & \multirow{2}{*}{RSS 25} & PaliGemma & 3.54 & 20.87 \\
 & & Qwen3-VL* & 5.74 & 20.66 \\
\multirow{2}{*}{OpenVLA-OFT~\cite{kim2025openvlaoft}} & \multirow{2}{*}{RSS 25} & Prismatic & 9.06 & 24.06 \\
 & & Qwen3-VL* & 10.86 & 30.49 \\
\midrule
\rowcolor{lightblue} \textbf{\dymethod(OURS)} & \textbf- & Qwen3-VL & \textbf{17.20} & \textbf{34.97} \\
\bottomrule
\end{tabular}
\leftline{\tiny{\qquad\qquad* Methods with Qwen3-VL backbones are our re-implementations for fair comparison.}}
\vspace{-2.5em}
\end{table}

Building upon the insight that effective dynamic manipulation requires both historical context and future anticipation, we evaluate \dymethod as a solution to these spatiotemporal challenges. As shown in~\tab{tab:main_result}, \dymethod demonstrates a performance advantage over SOTA VLA models on dynamic manipulation tasks.

\begin{wrapfigure}[14]{r}{0.5\textwidth}
    \vspace*{-2em}
    \centering
    \includegraphics[width=.5\textwidth]{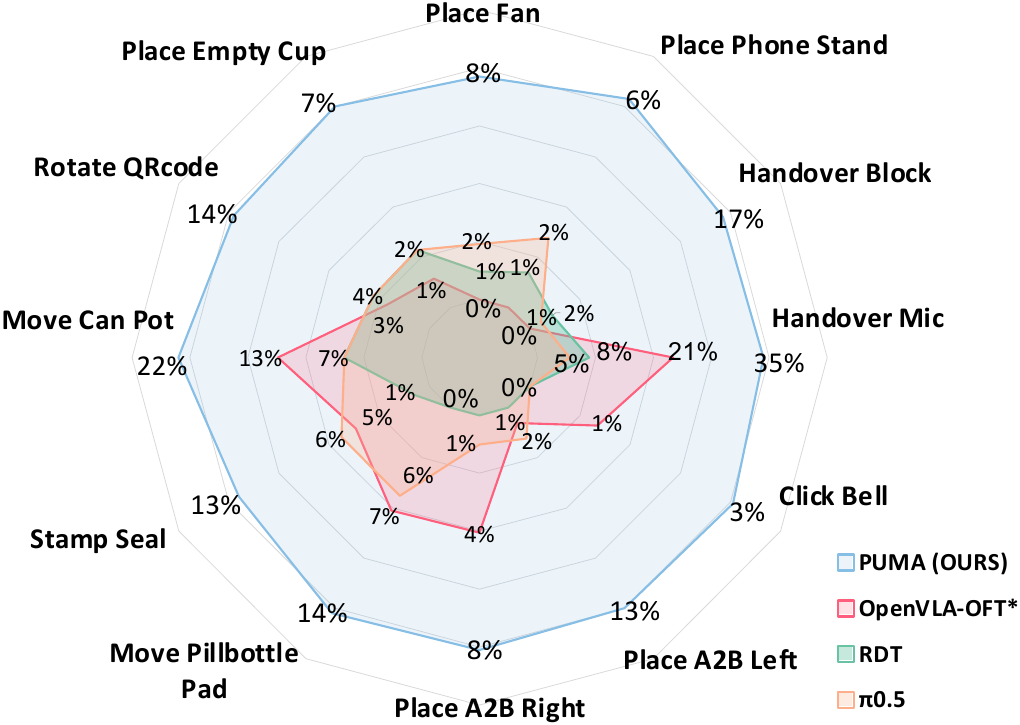}
    \vspace{-10pt}
    \caption{
    \dymethod performs significantly better than other baselines on difficult tasks.
    }
    \vspace*{-1.5em}
    \label{fig:radar}
\end{wrapfigure}

Specifically, \dymethod achieves the highest average success rate of 17.20\%, substantially outperforming recent strong baselines such as OpenVLA-OFT (Qwen3-VL-based)~\cite{starvla2025} (10.86\%) and $\pi_{0.5}$~\cite{pi05} (9.63\%). Furthermore, our method attains a peak Manipulation Score of 34.97, indicating a higher quality of interaction with moving targets. We attribute this performance leap to the explicit integration of historical observation frames and the introduction of the auxiliary future feature predictor $\psi_\omega$. By forcing shared representation to anticipate object dynamics during training, model learns to infer future states of moving objects during inference. This design effectively mitigates the dynamic awareness bottleneck identified in previous section.

To provide a detailed task-level comparison, we visualize the performance breakdown across individual manipulation tasks in~\fig{fig:radar}. Notably, \dymethod yields substantial improvements on highly challenging dynamic tasks where baseline methods struggle and achieve only marginal success rates. These pronounced performance gains underscore robustness of our approach in handling complex object dynamics. Overall, the quantitative results validate that our unified, dynamics-aware design represents an effective step toward generalizable manipulation in dynamic environments.

\begin{table}[t]
\centering
\scriptsize
\begin{minipage}[t]{0.48\textwidth}
\centering
\setlength{\tabcolsep}{1mm}
\caption{Task-type breakdown on Level~1 \dybenchwo@$0.1$. DI: Dynamic Interception, DT: Dynamic Tracking.}
\label{tab:task_type}
\vspace{-8pt}
\begin{tabular}{ l ccccc }
\toprule
\multirow{2}{*}{Method} & \multicolumn{2}{c}{DI} & \multicolumn{2}{c}{DT} \\
\cmidrule(lr){2-3} \cmidrule(lr){4-5}
& SR (\%) & MS & SR (\%) & MS \\
\midrule
Open.-OFT~\cite{kim2025openvlaoft} & 14.3 & 34.0 & 2.9 & 12.3 \\
$\mathbf{\pi}_{\mathbf{0.5}}$~\cite{pi05} & 11.8 & 33.6 & 7.0 & 17.3 \\
\rowcolor{lightblue} \textbf{\dymethod} & \textbf{24.2} & \textbf{47.9} & \textbf{8.9} & \textbf{19.6} \\
\bottomrule
\end{tabular}
\end{minipage}\hfill
\begin{minipage}[t]{0.48\textwidth}
\centering
\setlength{\tabcolsep}{1mm}
\caption{Cross-level evaluation on the 10-task subset. \dymethod adapts a Level~1 checkpoint with LoRA.}
\label{tab:cross_level_main}
\vspace{-8pt}
\begin{tabular}{lcccc}
\toprule
\multirow{2}{*}{Method} & \multicolumn{2}{c}{Level 2} & \multicolumn{2}{c}{Level 3} \\
\cmidrule(lr){2-3}\cmidrule(lr){4-5}
& SR (\%) & MS & SR (\%) & MS \\
\midrule
$\pi_{0.5}$~\cite{pi05} & 7.9 & 20.76 & 1.6 & 13.39 \\
\rowcolor{lightblue} \textbf{\dymethod} & \textbf{10.5} & \textbf{30.28} & \textbf{4.6} & \textbf{20.16} \\
\bottomrule
\end{tabular}
\end{minipage}
\vspace{-0.5em}
\end{table}

\tab{tab:task_type} shows that the two baselines have opposing strengths: OpenVLA-OFT favors DI (14.3\% vs.\ 2.9\% SR on DT) while $\pi_{0.5}$ is stronger on DT (7.0\% vs.\ 11.8\% on DI); \dymethod improves both, with DI gains from the auxiliary predictor estimating interception points and DT gains from optical flow enabling closed-loop tracking. This advantage holds under higher complexity (\tab{tab:cross_level_main}): \dymethod outperforms $\pi_{0.5}$ at both Level~2 (10.5\% vs.\ 7.9\%) and Level~3 (4.6\% vs.\ 1.6\% SR), confirming that temporal reasoning from simpler dynamics transfers to more complex motion patterns. Detailed analysis is provided in Appendix.

\vspace{-1em}
\subsection{The Role of Dynamic Data in Generalization}
\label{subsec:exp_data}
\vspace{-0.5em}

\begin{table}[t]
\setlength{\tabcolsep}{4mm}
\centering
\scriptsize
\caption{Comparison between co-training and dynamic data training on Level~1 \dybenchwo@$0.1$. * indicates our reproduction of OpenVLA-OFT~\cite{starvla2025} with the same Qwen3-VL backbone for fairness. To isolate gains from data mixing, \dymethod is evaluated with prediction horizon $N=2$.}
\label{tab:exp_mix}
\vspace{-10pt}
\makebox[\textwidth][c]{
\begin{tabular}{ lccC{11mm}C{11mm} }
\toprule
Method & Static & Dynamic & \makecell{SR (\%)$\uparrow$} & \makecell{MS$\uparrow$} \\
\midrule
\multirow{2}{*}{OpenVLA-OFT*~\cite{starvla2025}} & $\times$ & \checkmark & 10.86\phantom{\textsubscript{+0.00}} & 30.49\phantom{\textsubscript{+0.00}} \\
 & \checkmark & \checkmark & 12.89\textsubscript{\color{kaiming-green}{+2.03}} & 30.80\textsubscript{\color{kaiming-green}{+0.31}} \\
\midrule
\multirow{2}{*}{\dymethod} & $\times$ & \checkmark & 14.80\phantom{\textsubscript{+0.00}} & 32.74\phantom{\textsubscript{+0.00}} \\
 & \checkmark & \checkmark & 19.71\textsubscript{\color{kaiming-green}{+4.91}} & 37.76\textsubscript{\color{kaiming-green}{+5.02}} \\
\bottomrule
\end{tabular}%
}
\vspace{2em}
\includegraphics[width=1\textwidth]{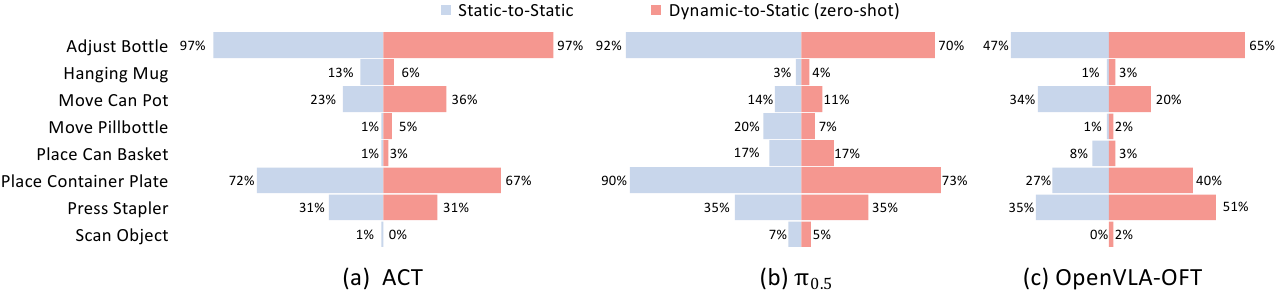}
\vspace{-2em}
\begin{tablenotes}[para, flushleft]
    \scriptsize
    The model trained on \dybench(\textcolor{coral}{coral}) demonstrates generalization ability when tested in zero-shot scenarios to static scenes, and achieves comparable results to static data(\textcolor{iceblue}{blue}) in some tasks.
\end{tablenotes}
\vspace{-1.5em}
\end{table}

To investigate the generalization capability of dynamic data, we evaluate the zero-shot performance of policies trained exclusively on dynamic datasets when deployed in static environments. As shown in the bar charts of~\tab{tab:exp_mix}, dynamic-trained models achieve comparable performance to their static-trained counterparts on certain tasks. For instance, using the OpenVLA-OFT baseline, the dynamic-trained policy outperforms the static-trained policy on Adjust Bottle (65\% vs. 47\%) and Place Container Plate (40\% vs. 27\%) tasks, while maintaining parity on Adjust Bottle task (97\%) under ACT baseline. These results suggest that dynamic training can supplement static training, yielding transfer benefits on tasks where the required manipulation primitives partially overlap.

Furthermore, we evaluate co-training policies on mixed static and dynamic datasets. As detailed in~\tab{tab:exp_mix}, this hybrid training strategy yields significant performance improvements in dynamic environments. Specifically, for \dymethod, incorporating static data increases the overall SR by 4.91\% (from 14.80\% to 19.71\%) and the MS by 5.02 compared to training on dynamic data alone. We hypothesize that this synergy arises because static data provides stable structural priors for foundational manipulation, while dynamic data introduces the spatiotemporal variations necessary for reactive dexterity.

\begin{conclusionbox}
    \textit{\textbf{Finding 3}: Dynamic data fosters generalizable representations.}
    Exposure to dynamic interactions mitigates overfitting to static positional biases, encouraging the policy to learn robust spatiotemporal representations. This facilitates effective zero-shot transfer to static environments and, when co-trained with static data, maximizes dynamic manipulation performance by combining stable foundational priors with reactive dexterity.
\end{conclusionbox}

\vspace{-1em}
\subsection{Real-World Robot Experiments}
\label{subsec:exp_real}
\vspace{-0.5em}

\begin{figure}[t]
\centering
\begin{minipage}[t]{0.68\textwidth}
  \centering
  \begin{minipage}[c]{0.43\linewidth}
    \centering
    \includegraphics[width=\linewidth]{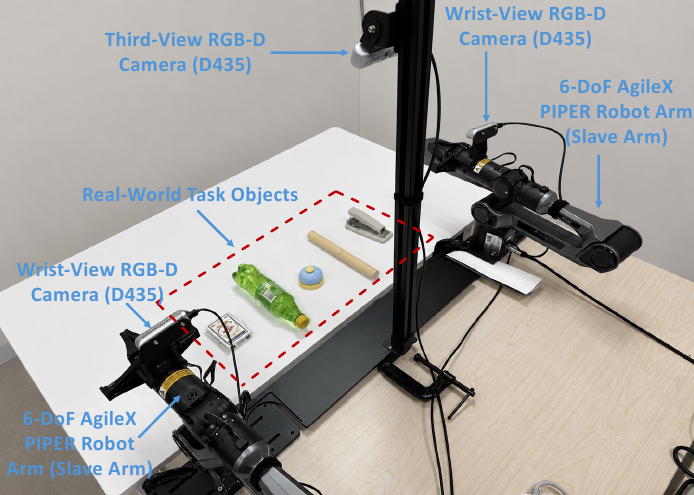}
  \end{minipage}\hfill
  \begin{minipage}[c]{0.55\linewidth}
    \centering
    \scriptsize
    \setlength{\tabcolsep}{0.5mm}
    \begin{tabular}{ l C{4.5mm}C{4.5mm}C{4.5mm}C{4.5mm}C{4.5mm}C{4.5mm} }
    \toprule
    Method & \rotatebox{55}{\scriptsize Bell} & \rotatebox{55}{\scriptsize Roller} & \rotatebox{55}{\scriptsize Bottle} & \rotatebox{55}{\scriptsize Card} & \rotatebox{55}{\scriptsize Stapler} & Avg \\
    \midrule
    ACT & 5 & 10 & 5 & 5 & 10 & 7 \\
    RDT-1B & 10 & 5 & 10 & 0 & 15 & 8 \\
    $\pi_{0.5}$ & 15 & 20 & 20 & 25 & 40 & 24 \\
    \rowcolor{lightblue} \textbf{\dymethod} & \textbf{50} & \textbf{45} & \textbf{35} & \textbf{25} & \textbf{55} & \textbf{42} \\
    \bottomrule
    \end{tabular}
  \end{minipage}
  \vspace{-0.5em}
  \caption{\textbf{Left:} Real-world setup with two 6-DoF AgileX Piper arms. \textbf{Right:} Real-world SR (\%) on five dynamic tasks. Each entry averages over 20 trials.}
  \label{fig:real_world}
\end{minipage}\hfill
\begin{minipage}[t]{0.28\textwidth}
  \centering
  \scriptsize
  \setlength{\tabcolsep}{1mm}
  \begin{tabular}{ l C{9mm} C{6mm} }
  \toprule
  Method & Latency (ms) & Freq. (Hz) \\
  \midrule
  OpenVLA & 166.7 & 6.0 \\
  Open.-OFT & 76.9 & 13.0 \\
  $\mathbf{\pi}_{\mathbf{0}}$ & 106.5 & 9.4 \\
  $\mathbf{\pi}_{\mathbf{0.5}}$ & 103.0 & 9.7 \\
  \rowcolor{lightblue} \textbf{\dymethod} & 103.7 & 9.6 \\
  \bottomrule
  \end{tabular}
  \vspace{-0.5em}
  \captionof{table}{Inference latency on a single RTX 4090 GPU.}
  \label{tab:latency}
\end{minipage}
\vspace{-1.5em}
\end{figure}

We validate sim-to-real transfer on the bimanual AgileX Piper platform illustrated in~\fig{fig:real_world} (left), whose hardware configuration matches our simulation. We select five dynamic tasks from \dybench (Click Bell, Grab Roller, Lift Bottle, Move Playingcard Away, Press Stapler) and collect 50 teleoperated demonstrations per task. All methods are trained on this shared dataset; \dymethod is LoRA fine-tuned from its simulation checkpoint.
As reported in~\fig{fig:real_world} (right), \dymethod achieves 42\% average SR, compared with 24\% for $\pi_{0.5}$ and below 10\% for ACT and RDT, which lack dynamic awareness. These results confirm that dynamic awareness acquired in simulation transfers to real hardware via lightweight LoRA adaptation.
Real-robot rollout visualizations are provided in Appendix.

\vspace{-1em}
\subsection{Inference Latency and Control Frequency}
\label{subsec:exp_efficiency}
\vspace{-0.5em}
We benchmark end-to-end inference latency on a single RTX 4090 (batch size 1, bf16, averaged over 500 queries after 50 warm-ups). As shown in~\tab{tab:latency}, \dymethod runs at 103.7\,ms per step (9.6\,Hz), matching $\pi_{0.5}$. The Qwen3-VL backbone accounts for 98.3\,ms (94.7\%); optical-flow computation adds 3.45\,ms (3.3\%) and preprocessing 2.0\,ms (2.0\%). The dynamic-awareness modules add negligible overhead, and action chunking amortizes each query over multiple control steps.

\vspace{-1em}
\begin{table}[t]
\centering
\scriptsize
\setlength{\tabcolsep}{2.5mm}
\caption{Ablation of core components and future prediction steps ($N$) on Level~1 \dybenchwo@$0.1$. We evaluate the impact of historical representations (Hist. Rep.), auxiliary future prediction (Aux. Pred.), and the prediction horizon.}
\label{tab:ablation_unified}
\vspace{-8pt}
\begin{tabular}{ l ccc C{11mm}C{11mm} }
\toprule
\multirow{2}{*}{Method} & \multicolumn{3}{c}{Components} & \multirow{2}{*}{SR (\%)$\uparrow$} & \multirow{2}{*}{MS$\uparrow$} \\
\cmidrule(lr){2-4}
& Hist. Rep. & Aux. Pred. & Steps ($N$) & & \\
\midrule
Baseline & $\times$ & $\times$ & - & 10.86 & 30.49 \\
+ Hist. Flow & \checkmark & $\times$ & - & 11.71 & 31.02 \\
+ Hist. Flow & \checkmark & \checkmark & 2 & 14.80 & 32.74 \\
+ Hist. Frames & \checkmark & \checkmark & 2 & 8.15 & 28.62 \\
\rowcolor{lightblue} + Hist. Flow & \checkmark & \checkmark & 4 & \textbf{17.20} & \textbf{34.97} \\
\bottomrule
\end{tabular}
\vspace{-2em}
\end{table}

\subsection{Ablation Study}
\label{subsec:exp_al}

We conduct an ablation study to evaluate our proposed modules, as detailed in~\tab{tab:ablation_unified}. First, providing explicit motion cues via optical flow (+ Hist. Flow) improves the baseline SR from 10.86\% to 11.71\%. Adding the auxiliary future prediction task at $N=2$ (+ Hist. Flow with Aux. Pred.) further boosts the SR to 14.80\%, confirming that anticipating future states effectively regularizes the action policy. Crucially, replacing optical flow with raw historical frames (+ Hist. Frames) degrades the SR to 8.15\%, demonstrating that implicitly deducing temporal dynamics from raw frames is suboptimal compared to utilizing explicit flow representations. Finally, extending the prediction horizon to $N=4$ achieves the best performance (17.20\% SR, 34.97 MS), suggesting that a longer anticipation horizon enables a more robust understanding of future trajectories.

\vspace{-1em}

\section{Related Work} 
\subsection{Vision-Language-Action Model} 
Vision-Language-Action models~\cite{shridhar2022cliport,reed2022generalist,brohan2022rt,zhang2024navid,zitkovich2023rt,belkhale2024rt} have transformed robotic manipulation by directly mapping multimodal inputs and instructions into control commands, and this paradigm has been extended to autonomous driving~\cite{fu2025orion,fu2025minddrive} and multi-task coordination~\cite{liang2026cook}. 
Recent architectures~\cite{liu2024rdt,kim2024openvla,pi0,pi05,internvlam1,kim2025openvlaoft,bjorck2025gr00t} leverage mechanisms like diffusion policies to achieve robust zero-shot adaptability. 
However, their application in dynamic real-world environments is bottlenecked by inadequate spatiotemporal modeling.
Regarding temporal modeling, the high computational cost of processing multi-frame observations forces most VLAs to operate as memoryless single-frame policies. 
This precludes the extraction of continuous dynamics. While recent efforts incorporate temporal contexts~\cite{jiang2023vima,driess2023palm,team2024octo,fan2025interleave,bu2025univla,patratskiy2025spatial,guan2026videostreamingthinking} or memory mechanisms for long-horizon tasks~\cite{shi2025memoryvla,jang2025contextvla,koo2025hamlet,liu2025ttf,lin2025hif,zheng2024tracevla}, they primarily track task progression. 
They fail to perform high-frequency motion estimation required for dynamic manipulation.
In terms of spatial representation, standard VLAs assume static environments and neglect the continuous motion of manipulated objects~\cite{qiu2026efficient}, despite progress in 3D spatial understanding~\cite{liang2024pointmamba,liang2024pointgst,wu2026vega}. Approaches like ReconVLA~\cite{song2025reconvla} allocate visual attention to target objects. DreamVLA~\cite{zhang2025dreamvla} integrated dream query to predict global scene transitions, and scene-level world models~\cite{liang2026UniFuture,zhou2025hermes,zhou2026hermespp} forecast future 3D states. However, these methods lack the fine-grained object-centric dynamic modeling essential for reactive planning. Furthermore, existing dynamic manipulation methods~\cite{zhang2025dynamic,wang2024train} are typically restricted to simplified motion tasks and fail to capture real-world dynamic complexity.
To bridge these gaps, we introduce spatiotemporal designs to effectively tackle dynamic manipulation tasks.

\vspace{-0.5em}
\subsection{Datasets and Benchmarks for Robotic Manipulation} 
\vspace{-0.5em}
Robot learning benchmarks primarily encompass real-world and simulation environments. While real-world platforms~\cite{yakefu2025robochallenge,khazatsky2024droid,wu2024robomind,bu2025agibot,walke2023bridgedata,o2024open} enable standardized training and evaluation on physical robots, they are often hindered by limited reproducibility, hardware variations, and safety constraints. Consequently, simulated closed-loop environments remain indispensable for scalable and reliable policy evaluation~\cite{zhu2020robosuite,jiang2025dexmimicgen,mu2021maniskill,ehsani2021manipulathor,makoviychuk2021isaac,puig2023habitat,yu2020meta}.
In simulation, foundational benchmarks like RLBench~\cite{james2019rlbench} and CALVIN~\cite{mees2022calvin} established multi-task, long-horizon evaluation. Subsequent platforms address diverse challenges: LIBERO~\cite{liu2023libero} for lifelong knowledge transfer, and RoboCasa~\cite{robocasa2024} for generative task scaling. Furthermore, The Colosseum~\cite{pumacay2024colosseum} and SIMPLER~\cite{li24simpler} evaluate out-of-distribution robustness, while RoboTwin2.0~\cite{chen2025robotwin} and VLABench~\cite{zhang2025vlabench} explore bimanual coordination and cognitive reasoning.
However, existing simulation benchmarks fundamentally rely on a static world assumption where state transitions are strictly robot-driven, failing to assess the manipulation of independently moving targets. To address this, we introduces a comprehensive benchmark designed to evaluate generalizable dual-arm manipulation capabilities in dynamic environments.

\vspace{-1em}

\section{Conclusion}
\label{sec:conclusion}
\vspace{-0.5em}
Generalizable robotic manipulation in dynamic environments remains a critical yet underexplored challenge in embodied AI.
We introduce \dybench, a scalable benchmark with hierarchical dynamic complexities, and show that existing VLAs suffer severe performance degradation when targets move, exposing a lack of spatiotemporal reasoning in current architectures.
To address this, we propose \dymethod, which integrates historical optical flow with object-centric future prediction to anticipate target motion and achieve state-of-the-art performance.
Our analysis further reveals that dynamic training yields representations that generalize to static tasks, confirming the complementarity of the two domains.

\section*{Acknowledgments}
This work was supported in part by NSFC (62225603, 623B2038) and in part by Hubei Science and Technology Major Project (2024BAA007).

%
%
\bibliographystyle{splncs04}
\bibliography{main}

\appendix
\clearpage

\begin{center}
    \textbf{\Large Towards Generalizable Robotic Manipulation \\in Dynamic Environments}
    \\ [0.8cm]
    {\Large Supplementary Material}
    \\ [0.8cm]
\end{center}
This is the supplementary material for the paper "Towards Generalizable Robotic Manipulation in Dynamic Environments". 
We organize the content as follows:\\

\vspace{-0.8em}

\textbf{\hyperref[appendix:contributions]{A}} -- Author Contributions

\textbf{\hyperref[appendix:setup]{B}} -- Datasets and Implementation Details

\textbf{\hyperref[appendix:qualitative]{C}} -- Qualitative Results and Visualizations

\textbf{\hyperref[appendix:methodology]{D}} -- Additional Methodology Description

\textbf{\hyperref[appendix:quantitative]{E}} -- Additional Quantitative Experiments

\textbf{\hyperref[appendix:discussion]{F}} -- Discussion

\vspace{-0.8em}

\begin{figure*}[htbp] 
    \vspace{-1em}
    \centering
    \def\colW{\dimexpr(\textwidth-0.75em)/4\relax}
    
    \begin{tabular}{
        @{}
        p{\colW} @{\hspace{0.25em}}
        p{\colW} @{\hspace{0.25em}}
        p{\colW} @{\hspace{0.25em}}
        p{\colW} @{}
    }
        \multicolumn{4}{@{}l@{}}{\scriptsize \textbf{Level 1:} Predictable Low-Order Dynamics} \\
        \noalign{\vspace{-0.1em}}
        
        \animategraphics[loop,autoplay,width=\linewidth]{10}{video/level1/1/frames/1.mp4_frame_}{1}{30} & 
        \animategraphics[loop,autoplay,width=\linewidth]{10}{video/level1/2/frames/2.mp4_frame_}{1}{60} & 
        \animategraphics[loop,autoplay,width=\linewidth]{10}{video/level1/3/frames/3.mp4_frame_}{1}{30} & 
        \animategraphics[loop,autoplay,width=\linewidth]{10}{video/level1/4/frames/4.mp4_frame_}{1}{60} \\
        \noalign{\vspace{-0.5em}}
        
        \centering\tiny\linespread{0.9}Drop the soft doughy bread into the medium oval breadbasket. & 
        \centering\tiny\linespread{0.9}Use the compact stamping seal to mark Tan. & 
        \centering\tiny\linespread{0.9}Lift the box with cards inside and shift it outward. & 
        \centering\tiny\linespread{0.9}Grab the medium yellow cylindrical bottle, placing it into the plastic dustbin.
    \end{tabular}

    \begin{tabular}{
        @{}
        p{\colW} @{\hspace{0.25em}}
        p{\colW} @{\hspace{0.25em}}
        p{\colW} @{\hspace{0.25em}}
        p{\colW} @{}
    }
        \multicolumn{4}{@{}l@{}}{\scriptsize \textbf{Level 2:} Predictable High-Order Dynamics} \\
        \noalign{\vspace{-0.1em}}
        
        \animategraphics[loop,autoplay,width=\linewidth]{10}{video/level2/1/frames/1.mp4_frame_}{1}{20} & 
        \animategraphics[loop,autoplay,width=\linewidth]{10}{video/level2/2/frames/2.mp4_frame_}{1}{40} & 
        \animategraphics[loop,autoplay,width=\linewidth]{10}{video/level2/3/frames/3.mp4_frame_}{1}{65} & 
        \animategraphics[loop,autoplay,width=\linewidth]{10}{video/level2/4/frames/4.mp4_frame_}{1}{50} \\
        \noalign{\vspace{-0.5em}}
        
        \centering\tiny\linespread{0.9}Press the center top of the compact bell with round base. & 
        \centering\tiny\linespread{0.9}Take the golden bread and place it inside the black pan with curved edges. & 
        \centering\tiny\linespread{0.9}Hold the teal microphone and pass it to the other hand. & 
        \centering\tiny\linespread{0.9}Lift the brown can, place it in the plastic basket, and raise the plastic basket.
    \end{tabular}

    \begin{tabular}{
        @{}
        p{\colW} @{\hspace{0.25em}}
        p{\colW} @{\hspace{0.25em}}
        p{\colW} @{\hspace{0.25em}}
        p{\colW} @{}
    }
        \multicolumn{4}{@{}l@{}}{\scriptsize \textbf{Level 3:} Stochastic and Abrupt Dynamics} \\
        \noalign{\vspace{-0.1em}}
        
        \animategraphics[loop,autoplay,width=\linewidth]{10}{video/level3/1/frames/1.mp4_frame_}{1}{30} & 
        \animategraphics[loop,autoplay,width=\linewidth]{10}{video/level3/2/frames/2.mp4_frame_}{1}{30} & 
        \animategraphics[loop,autoplay,width=\linewidth]{10}{video/level3/3/frames/3.mp4_frame_}{1}{20} & 
        \animategraphics[loop,autoplay,width=\linewidth]{10}{video/level3/4/frames/4.mp4_frame_}{1}{30} \\
        \noalign{\vspace{-0.5em}}
        
        \centering\tiny\linespread{0.9}Lift the paymentsign and rotate it until the QR code faces you. & 
        \centering\tiny\linespread{0.9}Place the off-white fan on the Coral mat and verify it is facing the robot. & 
        \centering\tiny\linespread{0.9}Use both arms to grip the medium-sized roller tightly. & 
        \centering\tiny\linespread{0.9}Pick the barcode scanner, grab the tea box, and scan it with the barcode scanner.
    \end{tabular}
    
    \caption{
        Qualitative demonstrations on the DOMINO dataset across hierarchical dynamic complexities. 
        The first two columns illustrate expert trajectories in the clean setting, while the last two columns present those under domain randomization. Brief task descriptions are provided below each sequence. 
        \textbf{Best viewed in Adobe Acrobat Reader. Animations play automatically or upon clicking.}
    }
    \label{fig:appendix_demo}
\end{figure*}

\clearpage

\section{Author Contributions}
\label{appendix:contributions}

\vspace{-2em}
\begin{table}[h]
\centering
\small
\setlength{\tabcolsep}{2mm}
\begin{tabular}{p{0.45\textwidth} p{0.45\textwidth}}
\textbf{\textit{Project Advisors}} & \textbf{\textit{Roadmap and Methodology}} \\
Dingkang Liang, Xiang Bai, Xuanyang Xi & Heng Fang, Dingkang Liang \\[1.5em]
\textbf{\textit{Paper Writing}} & \textbf{\textit{Benchmark Design}} \\
Heng Fang, Dingkang Liang, Shangru Li, Shuhan Wang & Shangru Li, Heng Fang \\[1.5em]
\textbf{\textit{Method Design}} & \textbf{\textit{Policies Training and Evaluation}} \\
Heng Fang & Shuhan Wang, Heng Fang, Shangru Li \\[1.5em]
\textbf{\textit{Data Pipeline and Generation}} & \textbf{\textit{Real-World Deployment}} \\
Shangru Li, Heng Fang & Heng Fang \\
\end{tabular}
\end{table}
\vspace{-3em}

\section{Datasets and Implementation Details}
\label{appendix:setup}

\subsection{Datasets}

\dybench comprises 117,000 expert trajectories covering 35 dynamic tasks across five robot embodiments. Each trajectory captures synchronized multi-view RGB observations from head and wrist cameras, alongside proprioceptive states including joint positions and end-effector poses. We apply domain randomization to enhance policy generalization. Full list of 35 dynamic tasks is in~\tab{tab:35_tasks}.

\begin{table}[h]
    \vspace{-1em}
    \centering
    \caption{The full list of 35 dynamic manipulation tasks in \dybench, categorized by dynamic task type.}
    \label{tab:35_tasks}
    \vspace{-10pt}
    \scriptsize
    \setlength{\tabcolsep}{4mm}
    \begin{tabular}{lll}
    \toprule
    \multicolumn{3}{l}{\textbf{Dynamic Interception}} \\
    \midrule
    Adjust Bottle & Dump Bin Bigbin & Grab Roller \\
    Handover Block & Handover Mic & Hanging Mug \\
    Move Can Pot & Move Playing Card Away & Place A2B Left \\
    Place A2B Right & Place Bread Skillet & Place Can Basket \\
    Place Object Basket & Put Bottles Dustbin & Put Object Cabinet \\
    Rotate QRcode & Scan Object & Shake Bottle \\
    Shake Bottle Horizontally & & \\
    \midrule
    \multicolumn{3}{l}{\textbf{Dynamic Tracking}} \\
    \midrule
    Beat Block Hammer & Click Alarmclock & Click Bell \\
    Move Pillbottle Pad & Move Stapler Pad & Place Bread Basket \\
    Place Container Plate & Place Empty Cup & Place Fan \\
    Place Mouse Pad & Place Object Scale & Place Object Stand \\
    Place Phone Stand & Place Shoe & Press Stapler \\
    Stamp Seal & & \\
    \bottomrule
    \end{tabular}
    \vspace{-1.5em}
\end{table}

\paragraph{\textbf{\emph{Data Storage Format.}}}
During data collection, per-frame observations are initially stored as individual pickle files before being merged into per-episode HDF5 files. Each HDF5 file contains JPEG-encoded multi-view RGB images, dual-arm and gripper joint actions, end-effector poses, and optional depth maps and point clouds. To facilitate integration with various policy learning frameworks, we provide scripts to convert the raw HDF5 data into widely used formats, such as the ALOHA HDF5 and LeRobot dataset formats. Users can also extend the pipeline to export data into custom formats.

\paragraph{\textbf{\emph{Static Counterpart.}}}
To establish a rigorous comparison between static and dynamic manipulation capabilities, we adopt the corresponding static tasks from RoboTwin 2.0~\cite{chen2025robotwin}. This paired design ensures identical object models and task configurations, effectively isolating the performance impact of dynamics.

\subsection{Implementation Details}

\paragraph{\textbf{\emph{Training Details.}}}
All baseline models are trained on the proposed \dybench dataset using their official default configurations. All experiments are conducted on NVIDIA A100 GPUs. 
For the Qwen3-VL~\cite{Qwen3-VL} based VLA models, the policy is initialized from Qwen3-VL-4B. The visual-language backbone hidden dimension is 2560. The action head is implemented as an MLP and predicts 14-dimensional absolute joint actions with a future action window of 15 steps. Training uses the primary observation stream without additional proprioceptive state input, and visual observations are resized to $224 \times 224$. The models are optimized using AdamW~\cite{loshchilov2019decoupled} with $\beta_1=0.9$, $\beta_2=0.95$, $\epsilon=10^{-8}$, and weight decay $10^{-8}$. We use a cosine learning rate scheduler with a linear warmup over the first 5,000 steps and a minimum learning rate of $5\times10^{-7}$. The learning rates for the visual-language backbone and the action head are set to $10^{-5}$ and $10^{-4}$, respectively. The training process spans 100 epochs with a maximum of 100,000 steps. Mixed precision training and gradient checkpointing are enabled. Full-parameter fine-tuning is conducted with the DeepSpeed ZeRO-2 strategy. Training uses 8 GPUs with a per-device batch size of 8.

For the proposed \dymethod, we introduce hyperparameters to effectively model the dynamic world representations and historical flow.
Specifically, we employ 4 world queries and assign a weight of 0.05 to the world model loss. The world branch adopts per-frame cosine supervision on future object-centric features extracted by a frozen DINOv2-B/14 encoder. We expand both the history and future windows to 4 frames with a stride of 4. Historical observations are represented by optical flow. The resolution for historical images and optical flow computation is set to $64 \times 64$.

\paragraph{\textbf{\emph{Evaluation Details.}}}
We conduct closed-loop evaluations to assess the policy performance. For each task, the policy is evaluated over 100 episodes. In dynamic environments, object motion is dynamically initialized at the beginning of each episode. To simulate realistic physical interactions, the environment continuously monitors contacts between the gripper and dynamic objects. Once a contact is detected, the object stops its autonomous motion to reflect the grasping state. We enforce strict boundary checks where an episode is immediately terminated and recorded as a failure if a dynamic object moves out of the camera view. For specific tasks like lifting, success requires the vertical position of the target object to exceed a predefined threshold. In addition to the overall success rate, we track metrics including manipulation score and route completion. We also penalize erratic behaviors to thoroughly assess the robustness of the policy in dynamic environments.
For the Level~1/2/3 dynamic complexity experiment in the main paper (\fig{fig:level_perf}), the five selected tasks are \textit{Adjust Bottle}, \textit{Handover Mic}, \textit{Hanging Mug}, \textit{Move Can Pot}, and \textit{Place Container Plate}.

\section{Qualitative Results and Visualizations}
\label{appendix:qualitative}

\subsection{Dataset Visualization}
\label{appendix:dataset_vis}

\fig{fig:appendix_demo} visualizes the \dybench dataset across all three dynamic complexity levels, spanning both dynamic interception and dynamic tracking categories.

\subsection{Policy Rollout Visualization}
\label{appendix:rollout_vis}

\dymethod policy rollouts are shown on a representative subset of tasks. Each row displays four temporally ordered keyframes from a single episode.

\paragraph{\textbf{\emph{Simulation Results.}}}
\fig{fig:sim_rollout_di} presents rollouts on 5 Dynamic Interception (DI) tasks and~\fig{fig:sim_rollout_dt} shows rollouts on 5 Dynamic Tracking (DT) tasks, which together illustrate the learned manipulation behaviors in the SAPIEN simulation environment.

\paragraph{\textbf{\emph{Real Robot Results.}}}
\fig{fig:real_rollout} shows \dymethod rollouts on a physical dual-arm PIPER robot platform. To reproduce dynamic object motion in the real world, target objects are attached to transparent strings and pulled at controlled speeds. The resulting linear trajectories match the Level~1 dynamics used in simulation. All other experimental settings, including camera placement, workspace layout, and task instructions, strictly follow the simulation configuration. The policy is initialized from the simulation-pretrained checkpoint and adapted with LoRA on 50 real-world demonstrations per task.
\fig{fig:real_third_view} further provides a third-person side-view visualization of the real-world rollouts, offering an alternative perspective on the manipulation process.

\addtocounter{figure}{3} 
\begin{figure}[h]
\centering
\def\fW{0.235\textwidth}
\begin{subfigure}{\textwidth}
\centering
\includegraphics[width=\fW]{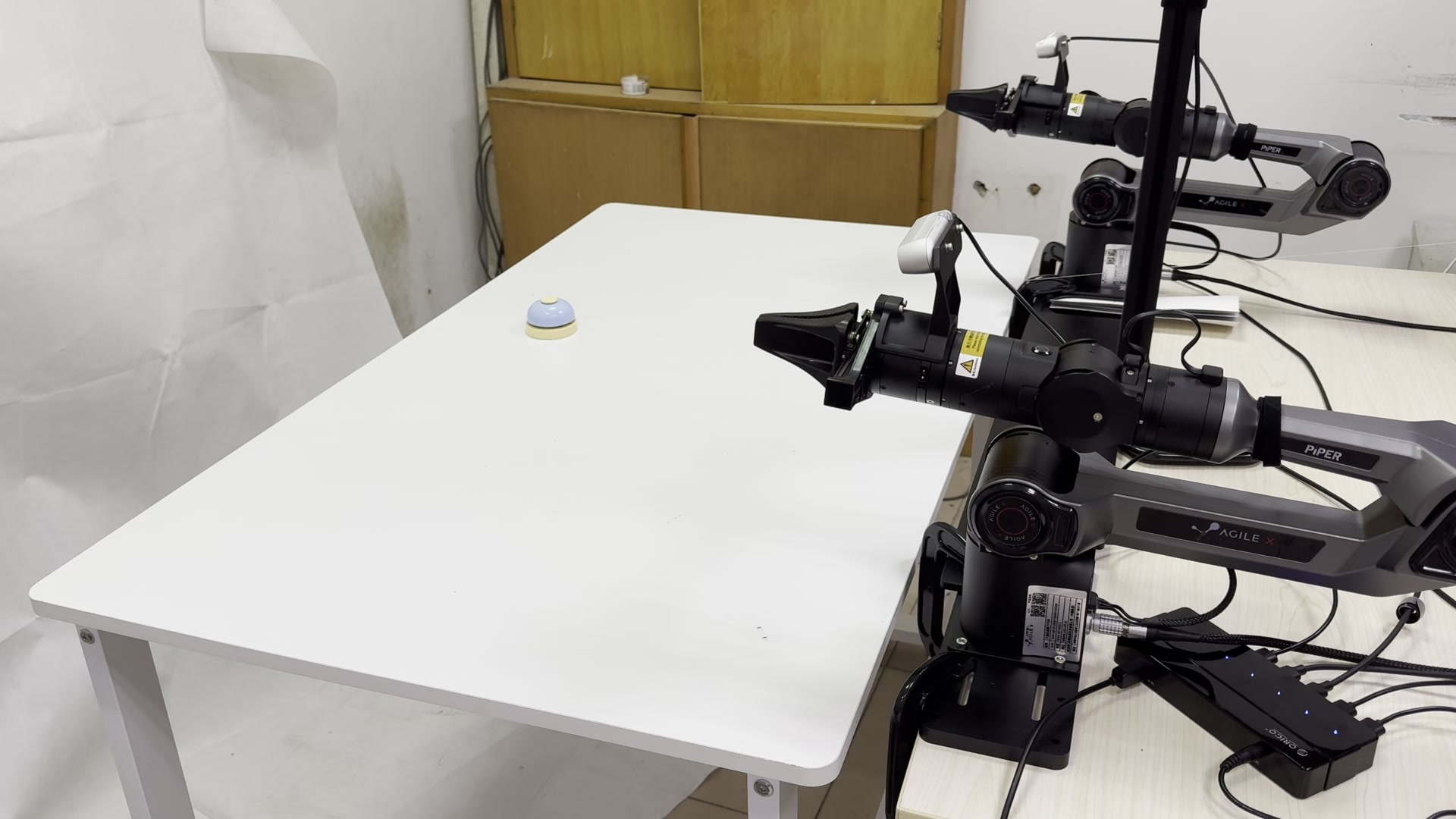}\hfill
\includegraphics[width=\fW]{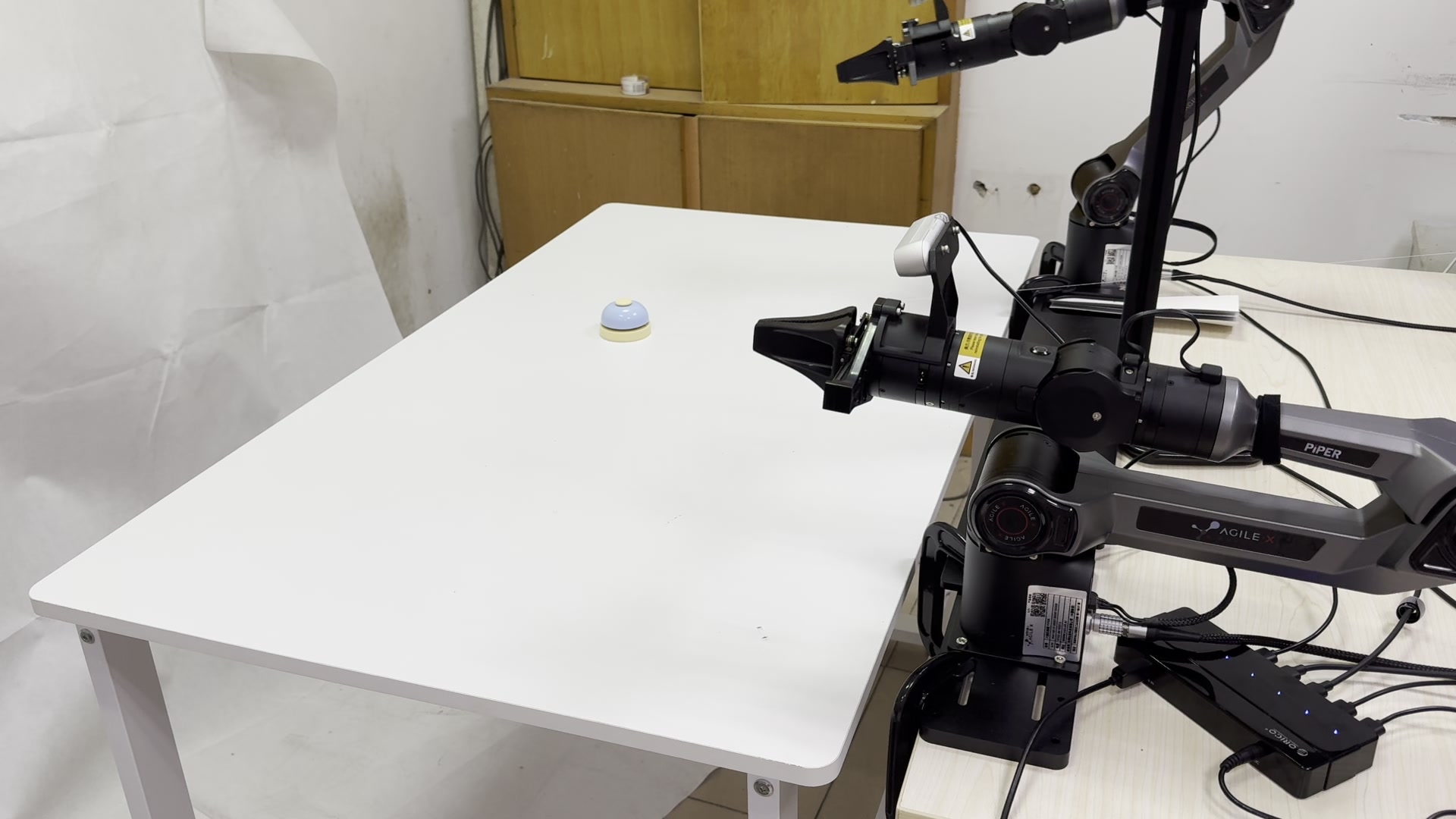}\hfill
\includegraphics[width=\fW]{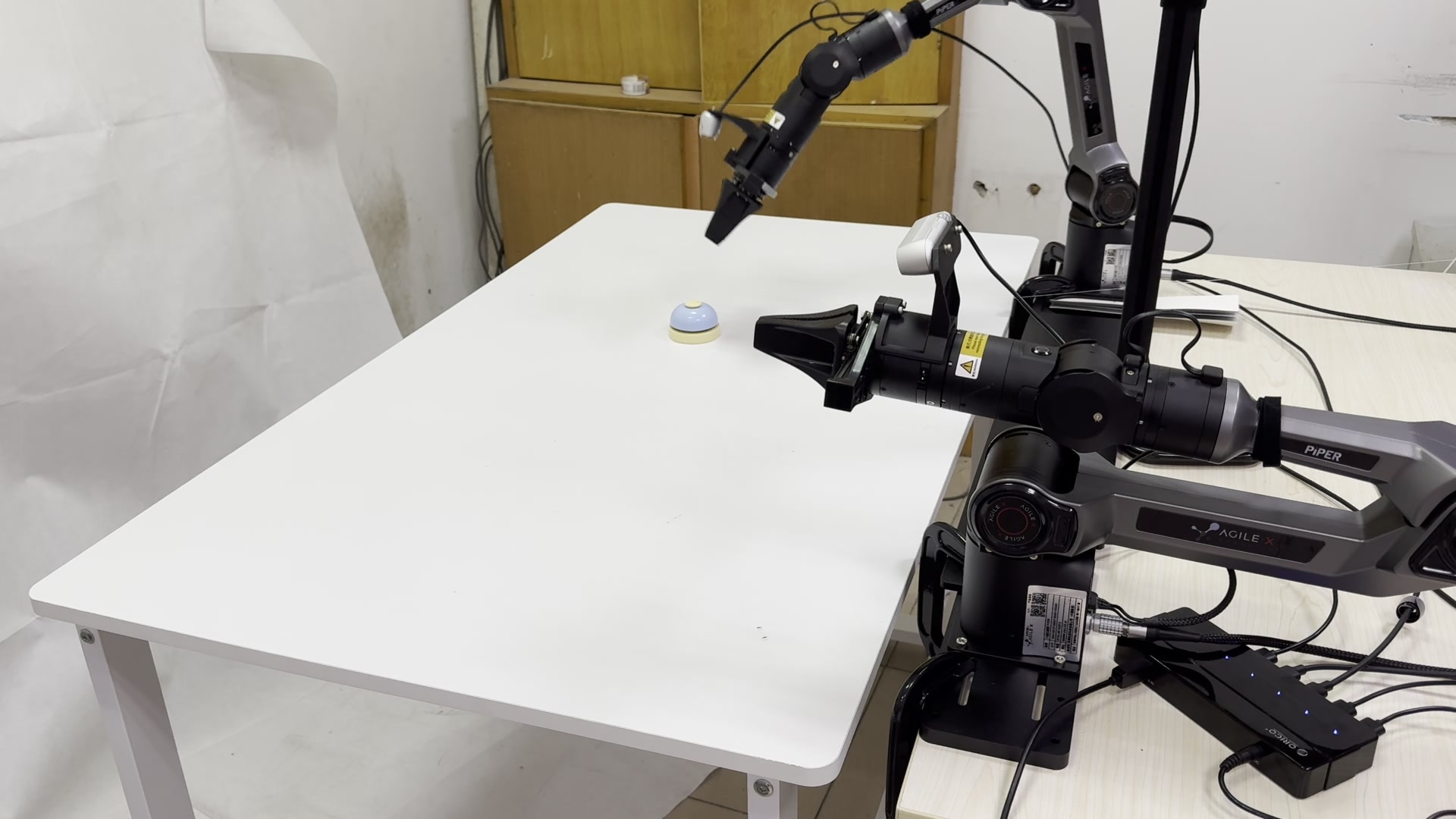}\hfill
\includegraphics[width=\fW]{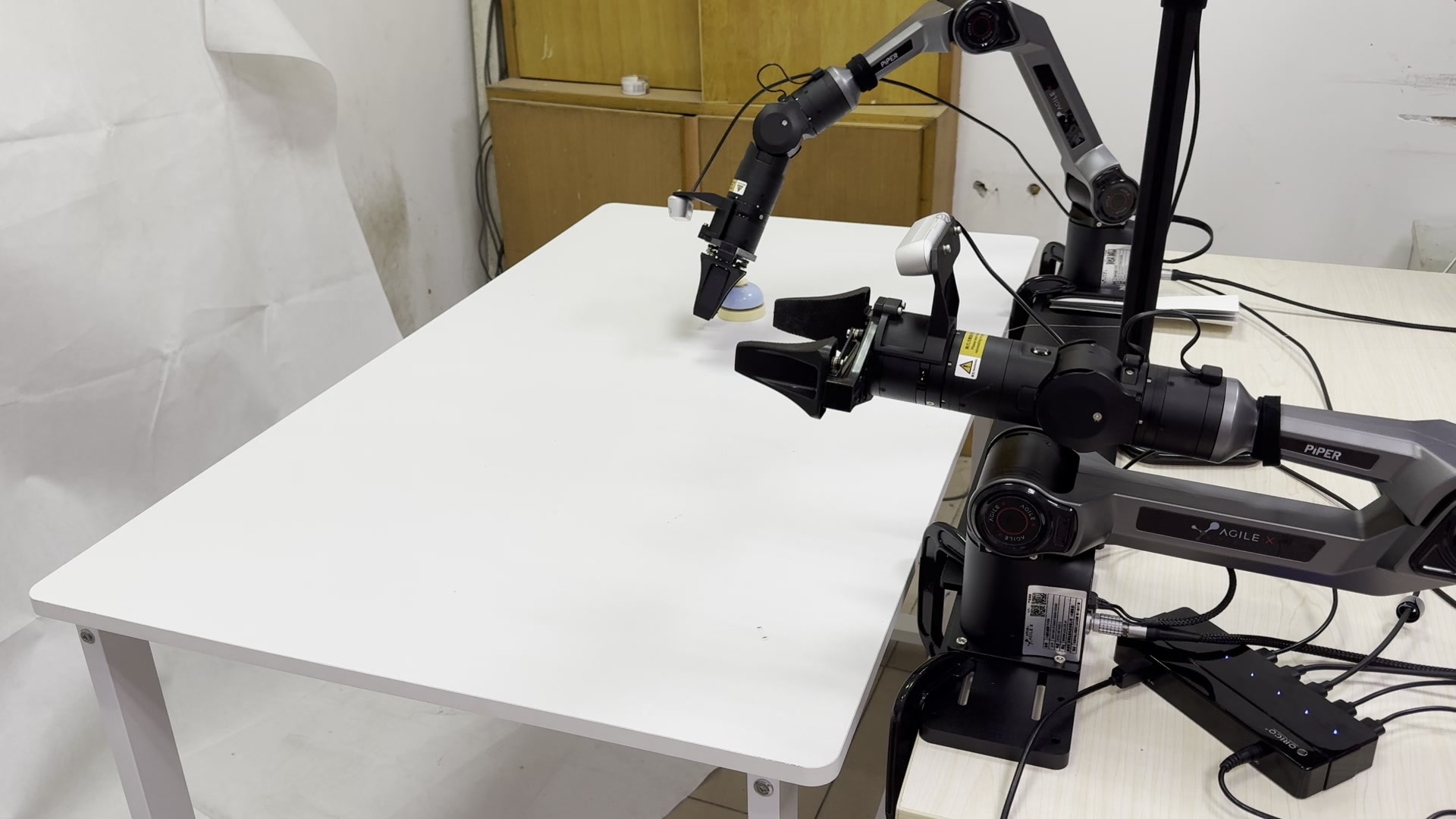}
\subcaption{Click Bell: ``Click the bell's top center on the table.''}
\end{subfigure}\\[0.4em]
\begin{subfigure}{\textwidth}
\centering
\includegraphics[width=\fW]{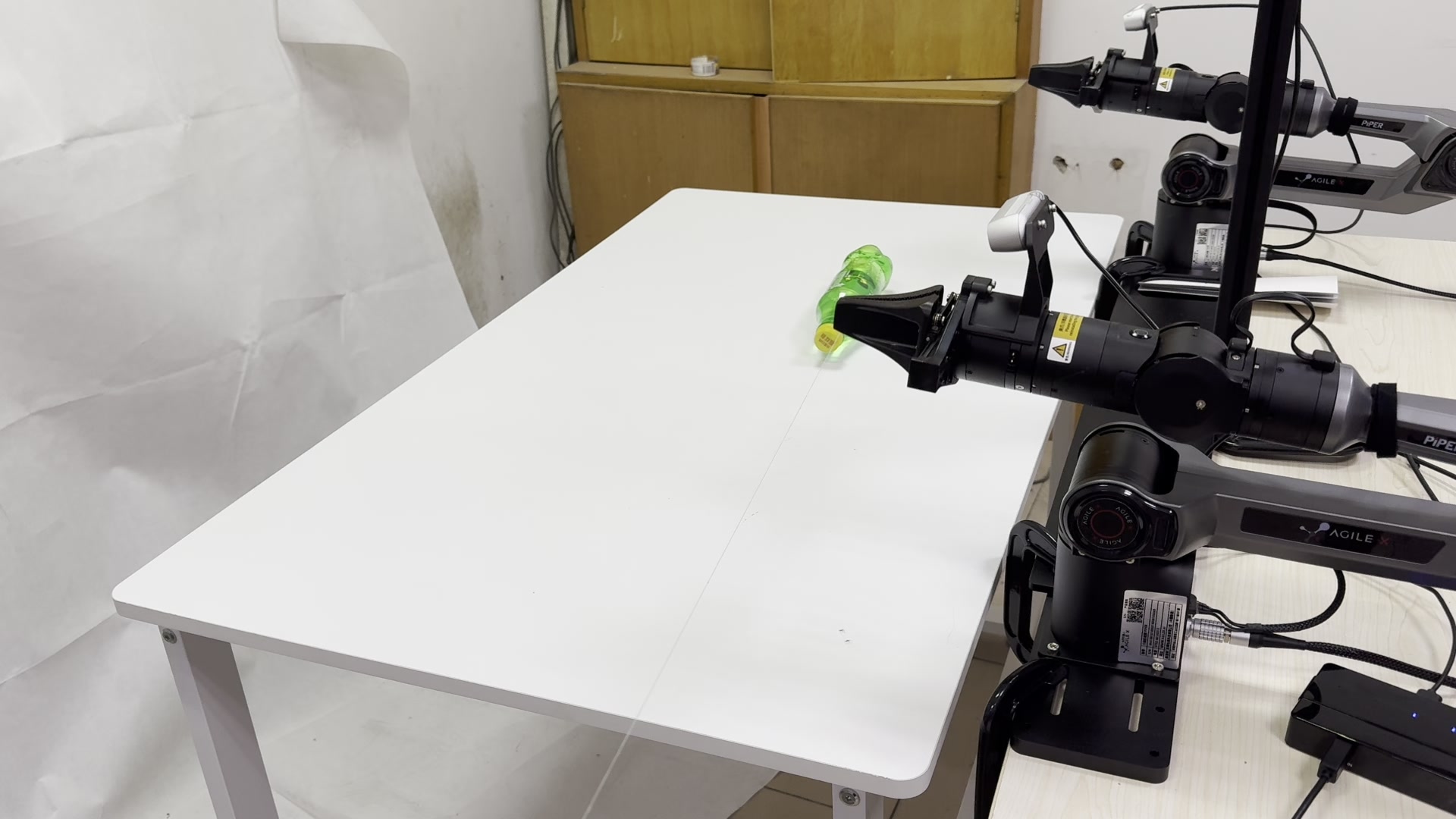}\hfill
\includegraphics[width=\fW]{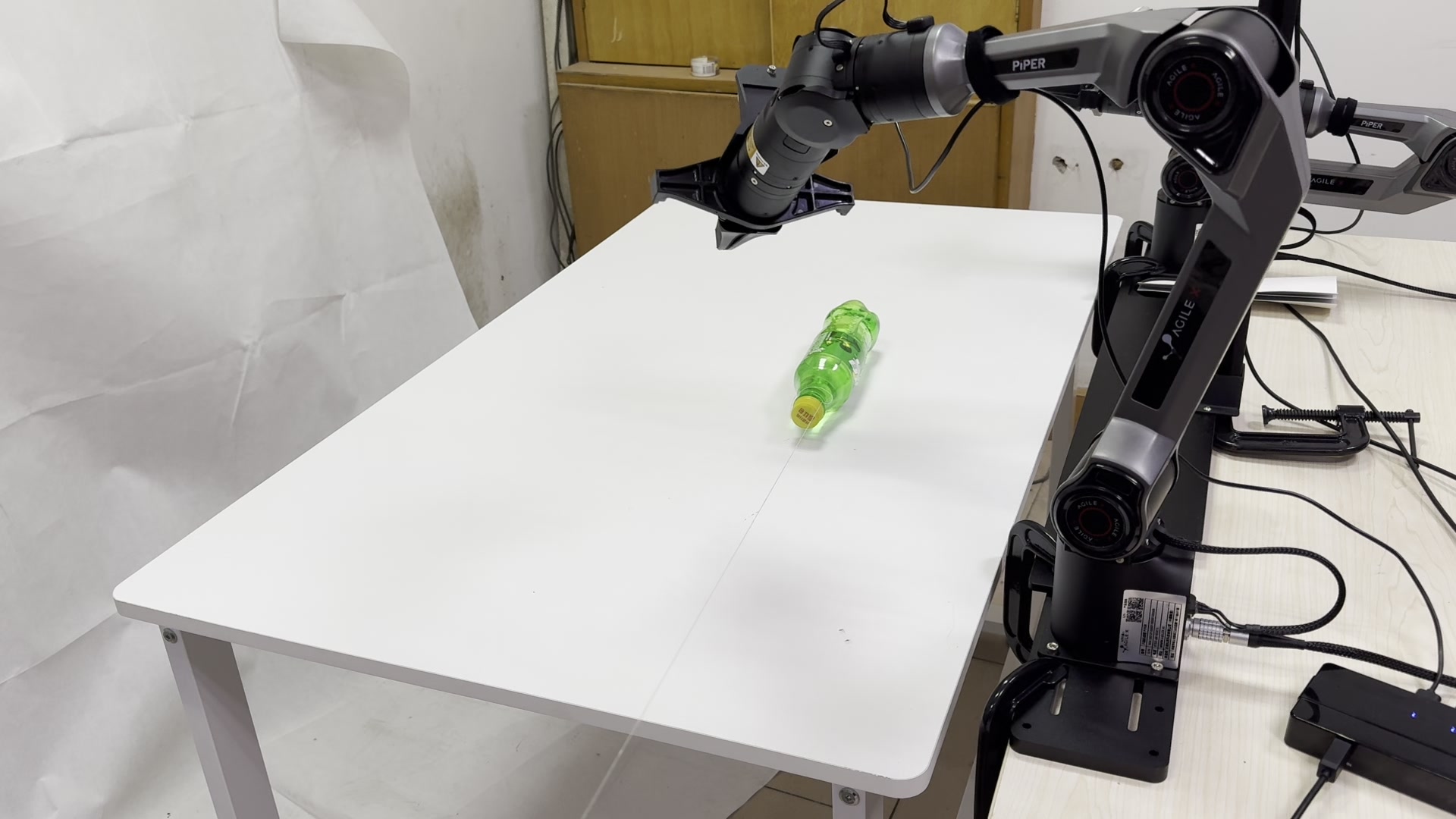}\hfill
\includegraphics[width=\fW]{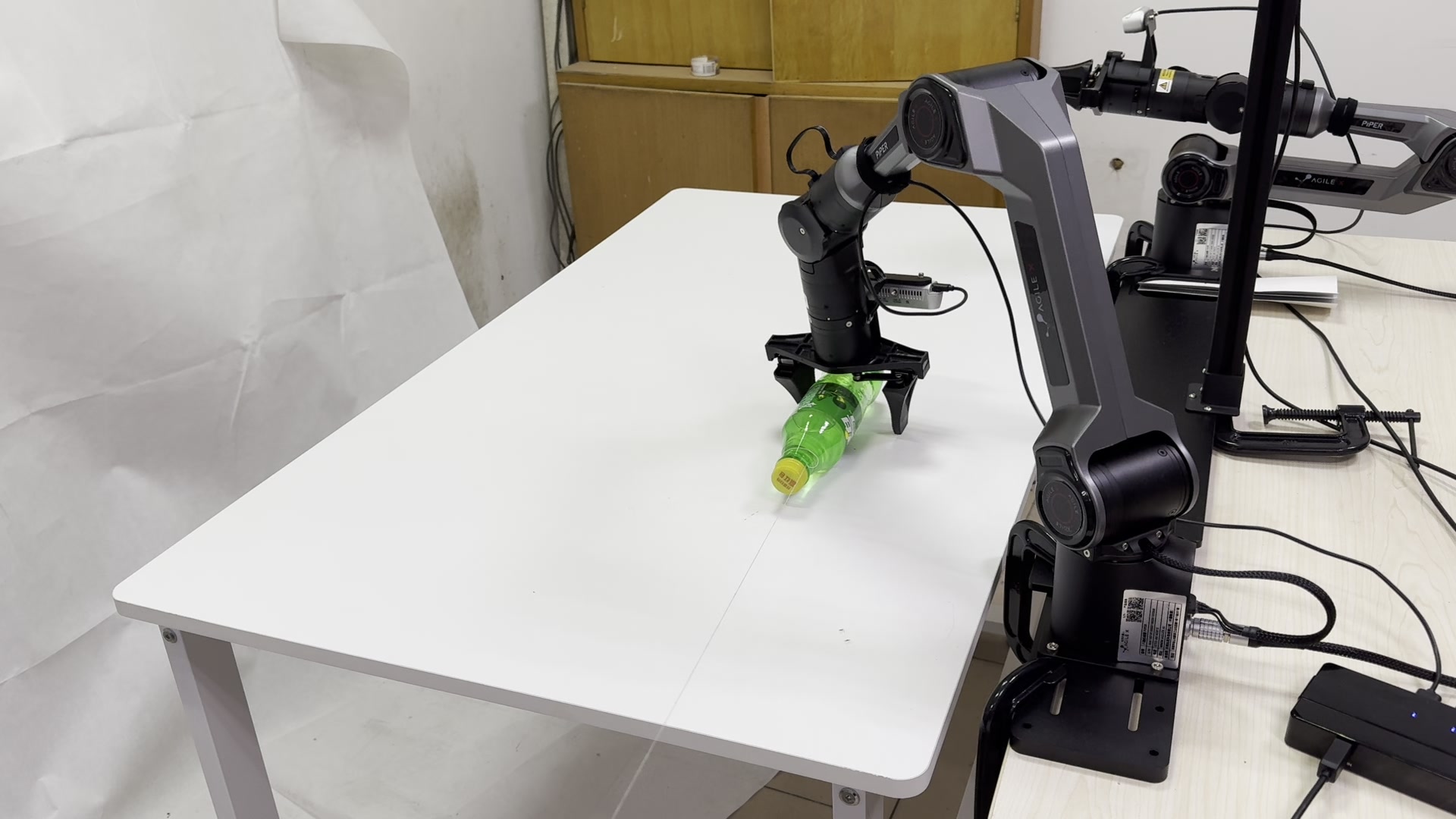}\hfill
\includegraphics[width=\fW]{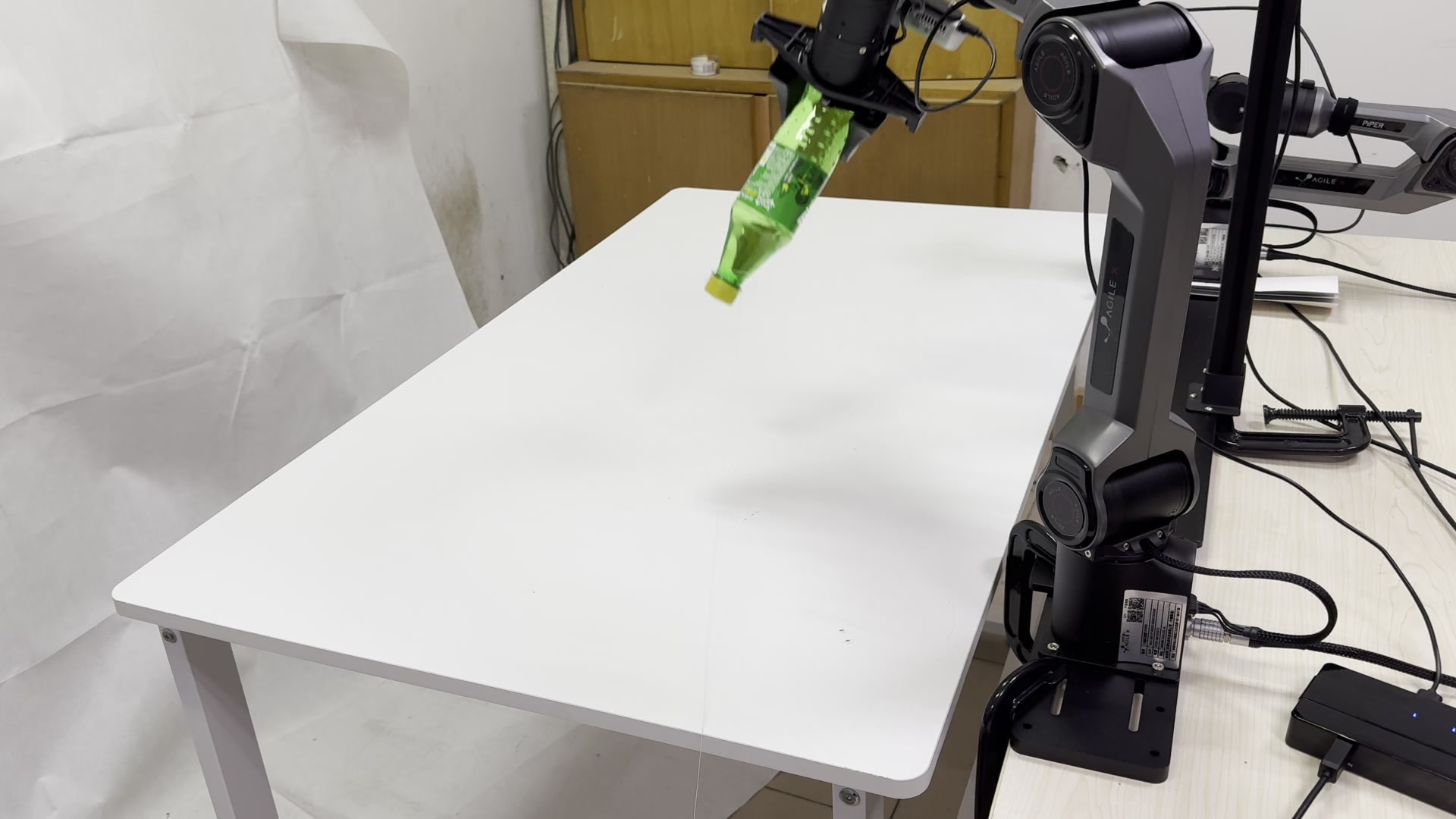}
\subcaption{Lift Bottle: ``Lift the Sprite bottle.''}
\end{subfigure}
\vspace{-1em}
\caption{Third-person side-view visualization of real-world rollouts. Each row shows four temporally ordered keyframes captured from a side-mounted camera.}
\label{fig:real_third_view}
\vspace{-1em}
\end{figure}

\section{Additional Methodology Description}
\label{appendix:methodology}

This section provides supplementary details on three components: the optical flow computation pipeline (\S\ref{appendix:flow_computation}), the Grounded-SAM~\cite{ren2024grounded} configuration for object-centric supervision (\S\ref{appendix:grounding_sam}), and the spatiotemporal synchronized data collection procedure (\S\ref{appendix:data_pipeline}).

\subsection{Optical Flow Computation}
\label{appendix:flow_computation}

We encode scene-centric historical dynamics using optical flow maps rather than raw frames. We detail the flow computation procedure and the caching strategy designed to minimize computational overhead during training and inference.

\paragraph{\textbf{\emph{Flow Computation.}}}
Given $h$ sampled historical frames and the current frame, we construct $h$ frame pairs. Both frames in each pair are first downsampled and converted to grayscale. We then compute dense optical flow using the Farneback algorithm~\cite{farneback2003two}. The resulting two-channel flow field (horizontal and vertical displacements) is mapped to an HSV color space, where hue encodes motion direction and value represents motion magnitude. This HSV image is subsequently converted to RGB format to serve as the flow map input for the visual encoder. 
To enhance robustness, we apply percentile-based normalization to the flow magnitude, preventing occasional large motions from dominating the representation. If the magnitude falls below a predefined threshold, the flow map is zeroed out to suppress noise.

\paragraph{\textbf{\emph{Caching Strategy.}}}
Computing optical flow on-the-fly during training introduces significant overhead. To mitigate this, we employ a disk-based caching mechanism. Each flow map is uniquely identified by a hash key encoding the dataset path, trajectory ID, step index, frame offsets, camera view, and target resolution. During the first training epoch, the computed RGB flow maps are saved as compressed NumPy arrays. In subsequent epochs, these cached maps are loaded directly, effectively amortizing the computational cost across the training process. 
During inference, optical flow is computed in real time. The system maintains a buffer of historical observations. At each step, the current and buffered frames are paired to compute the flow maps, which are then resized and fed into the model alongside the multi-view observations.

\subsection{Grounded-SAM Configuration}
\label{appendix:grounding_sam}

We utilize a frozen grounding module to extract object-centric supervisions for auxiliary future predictor. We detail its configuration and implementation.

\paragraph{\textbf{\emph{Module Setup.}}}
The grounding pipeline comprises two frozen models. GroundingDINO~\cite{liu2023grounding} uses a Swin-T backbone for open-vocabulary object detection. SAM2~\cite{ravi2024sam2segmentimages} uses a Hiera-Large backbone for mask prediction. GroundingDINO takes an image and a text prompt as input and outputs bounding boxes with confidence scores. We filter these bounding boxes using a box threshold of 0.35 and a text threshold of 0.25. The box with the highest score is then fed into SAM2 to generate a binary segmentation mask. Both models remain frozen during training and introduce no additional learnable parameters.
We employ a rule-based parser to extract the specific target object from the language instruction as the text prompt for GroundingDINO. If this extraction fails, the complete instruction serves as the fallback prompt. This strategy ensures the grounding module focuses precisely on the manipulated objects.

\paragraph{\textbf{\emph{Caching Strategy.}}}
Similar to the optical flow pipeline, we employ a disk-based caching mechanism for the grounding masks. Each mask is indexed by a hash comprising the model configuration, frame identity, and text prompt. Masks are computed and cached once during the initial training phase, ensuring that subsequent epochs only require disk read operations, thereby minimizing computational overhead.

\subsection{Spatiotemporal Synchronized Data Collection}
\label{appendix:data_pipeline}

\begin{figure}[h]
\centering
\includegraphics[width=0.6\linewidth]{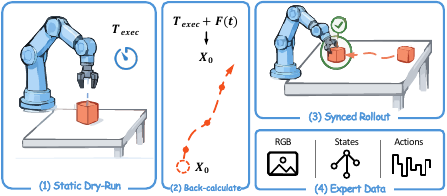}
\vspace{-0.5em}
\caption{Overview of the spatiotemporal synchronized data collection pipeline. The process comprises three stages: (1) a static dry-run to measure execution time, (2) kinematic back-calculation of the object's initial pose, and (3) a synchronized dynamic rollout.}
\label{fig:data_pipeline}
\vspace{-1em}
\end{figure}

\fig{fig:data_pipeline} illustrates three-stage data collection pipeline introduced in main paper.

\paragraph{\textbf{\emph{Static Dry-Run.}}}
The dry-run executes the robot's motion plan in a static scene while a step counter records the total simulation steps $T_{\text{total}}$. The physical execution time is $T_{\text{sec}} = T_{\text{total}} \times \Delta t$, where $\Delta t$ is the simulation timestep. The full environment state (joint positions, actor poses, and random number generator seed) is then checkpointed and restored so that the subsequent dynamic rollout starts from an identical configuration.

\paragraph{\textbf{\emph{Multi-Level Back-Calculation.}}}
Given the measured $T_{\text{sec}}$ and the manipulation end position $\mathbf{p}_{\text{end}}$, the trajectory generator back-calculates the object's starting position and motion plan. The procedure differs across the three complexity levels:

\textit{Level~1 (Predictable Low-Order Dynamics).}
A speed magnitude $v$ is drawn from a half-normal distribution $|\mathcal{N}(0.8\alpha,\,\alpha/3)|$ and clipped to $[0.01,\,\alpha]$~m/s. A direction angle $\theta$ is sampled uniformly from $[0, 2\pi)$, yielding the velocity vector $\mathbf{v} = v\,[\cos\theta,\,\sin\theta,\,0]^{\top}$. The start position follows directly:
\begin{equation}
  \mathbf{p}_{\text{start}} = \mathbf{p}_{\text{end}} - \mathbf{v} \cdot T_{\text{sec}}.
\end{equation}

\textit{Level~2 (Predictable High-Order Dynamics).}
The start position $\mathbf{p}_{\text{start}}$ is sampled uniformly within the workspace, and $n \in \{3,4,5\}$ control points are constructed: the first is $\mathbf{p}_{\text{start}}$, the last is $\mathbf{p}_{\text{end}}$, and the remaining $n{-}2$ intermediate points are obtained by linearly interpolating between start and end with additive per-axis noise $\mathcal{U}(-0.1,\,0.1)$, clipped to workspace bounds. These control points are placed at evenly-spaced normalized times $\{0, \frac{1}{n-1}, \ldots, 1\}$. Independent polynomials $p_x(\tau)$ and $p_y(\tau)$ of degree $\min(n{-}1,\,5)$ are fit via least-squares (\texttt{polyfit}) in normalized time $\tau \in [0,1]$. The instantaneous position at physical time $t$ is:
\begin{equation}
  \mathbf{x}(t) = \bigl[p_x(t/T_{\text{sec}}),\; p_y(t/T_{\text{sec}}),\; z_{\text{ws}}\bigr]^{\top}.
\end{equation}
A trajectory is accepted only if its average velocity (total arc length divided by $T_{\text{sec}}$) does not exceed $\alpha$.

\textit{Level~3 (Stochastic and Abrupt Dynamics).}
The total duration $T_{\text{sec}}$ is partitioned into $s \in \{2,3\}$ segments with durations drawn from a symmetric Dirichlet distribution $\text{Dir}(\mathbf{1}_s)$. The start position is sampled uniformly within the workspace, and each segment's endpoint is either a random workspace position (for intermediate segments) or $\mathbf{p}_{\text{end}}$ (for the final segment). Each segment independently adopts Level~1 dynamics with probability 0.2 or Level~2 dynamics with probability 0.8. For Level~1 sub-segments, the velocity is computed as $(\mathbf{p}_{\text{target}} - \mathbf{p}_{\text{current}})/\Delta T_i$ and clamped so that $\|\mathbf{v}\|_2 \leq \alpha$. For Level~2 sub-segments, 3--4 control points are fit with $\pm 0.08$ per-axis noise. Velocity and direction are generally discontinuous at segment boundaries, which produces abrupt directional changes.

All three levels enforce a quadrant-based visibility constraint: the end position determines a forbidden-exit region, and sampled trajectories that cross the corresponding workspace boundary are rejected. The generator retries up to 1000 attempts per episode.

\paragraph{\textbf{\emph{Contact Handling and Rollout.}}}
During the synchronized rollout, the environment monitors gripper-object collisions at each simulation step. Upon contact, the kinematic motion controller releases the target and converts it to a dynamic rigid body under physics simulation, so that grasping and manipulation forces behave realistically. An optional pre-motion buffer $T_{\text{pre}}$ extends the kinematic duration to $T_{\text{sec}} + T_{\text{pre}}$, so the object can travel further before the robot begins its action sequence.

\section{Additional Quantitative Experiments}
\label{appendix:quantitative}

This section provides additional experiments that supplement the results in the main paper, organized into three parts: extended ablation studies (\S\ref{appendix:ablation}), cross-difficulty generalization analysis (\S\ref{appendix:cross_difficulty}), and per-task detailed results (\S\ref{appendix:per_task}). Due to computational constraints, several experiments are conducted on a representative 10-task subset of \dybench, comprising 5 Dynamic Interception tasks (\textit{Grab Roller}, \textit{Handover Block}, \textit{Place A2B Left}, \textit{Place Can Basket}, \textit{Shake Bottle}) and 5 Dynamic Tracking tasks (\textit{Beat Block Hammer}, \textit{Move Pillbottle Pad}, \textit{Place Bread Basket}, \textit{Place Container Plate}, \textit{Press Stapler}). These tasks were selected to balance both task categories while spanning a range of manipulation complexities. For each experiment, we explicitly note whether evaluation is performed on this 10-task subset or the full 35-task set.

\subsection{Extended Ablation Studies}
\label{appendix:ablation}

\paragraph{\textbf{\emph{Comparison with History-Aware Baseline.}}}
To verify that the gain of \dymethod stems from how history is represented rather than from merely accessing past frames, we compare with ContextVLA~\cite{jang2025contextvla}, which concatenates raw historical frames as additional visual tokens (both fine-tuned on the 10-task subset under identical conditions). As shown in~\tab{tab:context_vla}, \dymethod outperforms ContextVLA by a wide margin (20.9\% vs.\ 7.8\% SR). ContextVLA shows no meaningful improvement over the no-history baseline, which indicates that the VLA backbone cannot extract useful temporal cues from raw frames. Optical-flow-based history encoding paired with future-state prediction captures target dynamics more effectively, and this result confirms that the \textit{representation} of historical information, not its mere presence, is the decisive factor for dynamic manipulation.

\begin{table}[h]
\centering
\scriptsize
\setlength{\tabcolsep}{3mm}
\caption{Comparison with history-aware baseline on the 10-task subset on Level~1 \dybenchwo@$0.1$.}
\label{tab:context_vla}
\vspace{-8pt}
\begin{tabular}{lcc}
\toprule
Method & SR (\%)$\uparrow$ & MS$\uparrow$ \\
\midrule
ContextVLA (raw-frame history) & 7.8 & 20.52 \\
\rowcolor{lightblue} \textbf{\dymethod} & \textbf{20.9} & \textbf{40.0} \\
\bottomrule
\end{tabular}
\vspace{-0.5em}
\end{table}

\paragraph{\textbf{\emph{History Length.}}}
The main paper adopts $h{=}4$ historical frames for optical flow computation. We investigate whether fewer frames suffice, particularly for Level~1 trajectories that follow linear paths, by evaluating a minimal setting of $h{=}1$ (optical flow computed from one historical frame and the current frame) on the full 35-task set.
Although Level~1 trajectories are in principle linearly extrapolable from two observation points, reducing from $h{=}4$ to $h{=}1$ causes a 4.2\% drop in SR (\tab{tab:history_length}). We attribute this gap to practical factors that violate the ideal linear assumption: partial occlusions may temporarily mask the target, viewpoint shifts introduce apparent motion inconsistencies, and control execution delays create a temporal gap between observation and action. A longer history window smooths these perturbations and yields a more stable estimate of the target's motion direction and speed.

\begin{table}[h]
\centering
\scriptsize
\setlength{\tabcolsep}{3mm}
\caption{Effect of history length on 35 tasks on Level~1 \dybenchwo@$0.1$. Both rows use the full \dymethod with auxiliary prediction at $N{=}4$.}
\label{tab:history_length}
\vspace{-8pt}
\begin{tabular}{lcc}
\toprule
History Length & SR (\%)$\uparrow$ & MS$\uparrow$ \\
\midrule
$h=1$ & 13.00 & 30.97 \\
\rowcolor{lightblue} $h=4$ & \textbf{17.20} & \textbf{34.97} \\
\bottomrule
\end{tabular}
\vspace{-0.5em}
\end{table}

\paragraph{\textbf{\emph{Extended Prediction Horizon.}}}
The main paper reports results for auxiliary prediction horizons $N{=}2$ and $N{=}4$, where $N$ is the number of future time steps over which the model predicts target object states. We extend this ablation to $N{=}6$ on the full 35-task set to test whether a longer horizon brings further improvement.
As shown in~\tab{tab:horizon_ext}, all prediction horizons improve over the no-prediction baseline, which confirms that future-state anticipation benefits dynamic manipulation. Performance peaks at $N{=}4$ and declines at $N{=}6$. We hypothesize that longer prediction windows force the model to forecast increasingly uncertain future states, especially for targets whose trajectories change direction or speed. The noisy supervision signal that results introduces conflicting gradients during training and degrades the learned representation. $N{=}4$ balances the benefit of forward-looking awareness against the cost of growing prediction uncertainty.

\begin{table}[h]
\centering
\scriptsize
\setlength{\tabcolsep}{3mm}
\caption{Prediction horizon ablation on 35 tasks on Level~1 \dybenchwo@$0.1$.}
\label{tab:horizon_ext}
\vspace{-8pt}
\begin{tabular}{lcc}
\toprule
Horizon & SR (\%)$\uparrow$ & MS$\uparrow$ \\
\midrule
\rowcolor{lightgray} Baseline (no prediction) & 10.86 & 30.49 \\
$N=2$ & 14.80 & 32.74 \\
\rowcolor{lightblue} $N=4$ & \textbf{17.20} & \textbf{34.97} \\
$N=6$ & 13.62 & 32.99 \\
\bottomrule
\end{tabular}
\vspace{-0.5em}
\end{table}

\subsection{Cross-Difficulty Generalization}
\label{appendix:cross_difficulty}

Most experiments in the main paper are conducted under Level~1 dynamics, where target objects follow linear trajectories. We extend the evaluation along two axes: (1) higher dynamic complexity levels (Level~2 and Level~3), where we examine both from-scratch training and transfer from Level~1 data, and (2) distributional shifts in the dynamic coefficient~$\alpha$, which controls target motion speed. All experiments use the 10-task subset.

\paragraph{\textbf{\emph{Cross-Level Evaluation.}}}
We compare two training strategies for \dymethod: (1) training from scratch directly on Level~2/3 data, and (2) fully fine-tuning on Level~1 first, then adapting to Level~2/3 with LoRA. The baseline $\pi_{0.5}$ is trained from scratch under the same conditions.
As shown in~\tab{tab:cross_level}, \dymethod trained from scratch outperforms $\pi_{0.5}$ at both complexity levels in SR and MS, which confirms that the dynamics-aware architecture retains its advantage on non-linear and reactive motion patterns. The LoRA-adapted variant, initialized from a Level~1 checkpoint, improves further (Level~3: 4.6\% vs.\ 2.5\% SR). This is notable because Level~1 data contains only constant-velocity linear trajectories, yet it provides a stronger initialization than random weights for higher-level fine-tuning. The temporal reasoning acquired from simple dynamics, such as estimating motion direction from optical flow and anticipating future object states, is not restricted to linear paths and can transfer to more complex patterns.
These results highlight that training data composition across difficulty levels substantially affects dynamic manipulation performance. Multiple strategies are viable, from single-level training to cross-level transfer and multi-level co-training, and each involves different trade-offs in data cost, training efficiency, and generalization. How to optimally organize and combine dynamic data across difficulty levels remains a promising open question.

\begin{table}[h]
\centering
\scriptsize
\setlength{\tabcolsep}{2mm}
\caption{Cross-level evaluation on the 10-task subset. \dymethod + LoRA adapts a Level~1 checkpoint with LoRA on the target level.}
\label{tab:cross_level}
\vspace{-8pt}
\begin{tabular}{lcccc}
\toprule
\multirow{2}{*}{Method} & \multicolumn{2}{c}{Level 2} & \multicolumn{2}{c}{Level 3} \\
\cmidrule(lr){2-3}\cmidrule(lr){4-5}
& SR (\%) & MS & SR (\%) & MS \\
\midrule
$\pi_{0.5}$ (from scratch) & 7.9 & 20.76 & 1.6 & 13.39 \\
\dymethod (from scratch) & 9.3 & 27.70 & 2.5 & 15.97 \\
\rowcolor{lightblue} \dymethod + LoRA & \textbf{10.5} & \textbf{30.28} & \textbf{4.6} & \textbf{20.16} \\
\bottomrule
\end{tabular}
\vspace{-0.5em}
\end{table}

\paragraph{\textbf{\emph{Dynamic Coefficient Generalization.}}}
The dynamic coefficient $\alpha$ controls the speed of target object motion within each complexity level. All methods in the main paper are trained with $\alpha{=}0.1$. We test generalization to two unseen settings: (1) a randomized regime where $\alpha$ is uniformly sampled from $[0.5, 1.5]$ per episode, which introduces substantial speed variability, and (2) a low-speed out-of-distribution setting with $\alpha{=}0.05$, where targets move slower than those seen during training.
As shown in~\tab{tab:alpha_gen}, \dymethod maintains a clear advantage over $\pi_{0.5}$ under both distributional shifts. We attribute this to the optical flow representation, which encodes frame-to-frame pixel displacement rather than absolute velocity. Changes in object speed therefore manifest as proportional scaling of the flow field, a variation that the model can accommodate without retraining.

\begin{table}[h]
\centering
\scriptsize
\setlength{\tabcolsep}{2mm}
\caption{Generalization to unseen dynamic coefficients on the 10-task subset on Level~1 \dybenchwo@$0.1$.}
\label{tab:alpha_gen}
\vspace{-8pt}
\begin{tabular}{lcccc}
\toprule
\multirow{2}{*}{Method} & \multicolumn{2}{c}{$\alpha$ random $[0.5, 1.5]$} & \multicolumn{2}{c}{$\alpha{=}0.05$ (OOD)} \\
\cmidrule(lr){2-3}\cmidrule(lr){4-5}
& SR (\%) & MS & SR (\%) & MS \\
\midrule
$\pi_{0.5}$ & 10.49 & 27.20 & 10.86 & 28.29 \\
\rowcolor{lightblue} \textbf{\dymethod} & \textbf{18.34} & \textbf{36.46} & \textbf{19.91} & \textbf{36.75} \\
\bottomrule
\end{tabular}
\vspace{-0.5em}
\end{table}

\paragraph{\textbf{\emph{Data Scaling.}}}
To examine how each method scales with more training data, we double the number of demonstrations (2$\times$ episodes per task) and retrain both $\pi_{0.5}$ and \dymethod on the 10-task subset under otherwise identical conditions.
With doubled data (\tab{tab:data_scale}), \dymethod reaches 27.4\% SR while $\pi_{0.5}$ reaches 15.6\%, a gap of 11.8 percentage points. This gap is wider than under the standard data budget, which indicates that \dymethod benefits more from additional demonstrations. The structured dynamic representations (optical flow and auxiliary future prediction) provide an inductive bias that helps the model extract more informative signal from each demonstration, particularly the temporal patterns of dynamic manipulation.

\begin{table}[h]
\centering
\scriptsize
\setlength{\tabcolsep}{3mm}
\caption{Performance with 2$\times$ training data on the 10-task subset on Level~1 \dybenchwo@$0.1$.}
\label{tab:data_scale}
\vspace{-8pt}
\begin{tabular}{lcc}
\toprule
Method & SR (\%)$\uparrow$ & MS$\uparrow$ \\
\midrule
$\pi_{0.5}$ & 15.6 & 30.8 \\
\rowcolor{lightblue} \textbf{\dymethod} & \textbf{27.4} & \textbf{43.9} \\
\bottomrule
\end{tabular}
\vspace{-0.5em}
\end{table}

\subsection{Per-Task Detailed Results}
\label{appendix:per_task}

The main paper reports average performance across all 35 tasks. Below we provide the complete per-task breakdown to facilitate detailed analysis.~\tab{tab:dy_full} presents the performance of \dymethod and all baseline methods on each of the 35 dynamic tasks. We additionally report results for four cross-domain transfer settings using the RoboTwin benchmark:~\tab{tab:robotwin-ss} (static$\to$static),~\tab{tab:robotwin-sd} (static$\to$dynamic),~\tab{tab:robotwin-ds} (dynamic$\to$static), and~\tab{tab:robotwin-dd} (dynamic$\to$dynamic).

\section{Discussion}
\label{appendix:discussion}

This section presents failure mode analysis (\S\ref{appendix:failure_modes}) and outlines directions for future work (\S\ref{appendix:future_work}).

\subsection{Failure Mode Analysis}
\label{appendix:failure_modes}

Overall success rates on \dybench remain low, indicating substantial room for improvement. We identify three recurring failure modes:
1)~\textit{Reactive chasing.} Single-frame policies react to the target's current position without anticipating its future location. The end-effector consistently lags behind the object, and this gap grows with target speed.
2)~\textit{Spatial distribution fitting.} Without temporal context, policies fit the spatial distribution of training-set target positions rather than learn motion dynamics, and fail when test-time trajectories deviate from training.
3)~\textit{Bimanual coordination failure.} In dual-arm tasks, an error in either arm can cause episode failure through collision, occlusion, or misaligned timing. Shared-representation architectures provide no mechanism to isolate single-arm errors.

\subsection{Limitations and Future Work}
\label{appendix:future_work}

Dynamic manipulation remains a largely open research direction with multiple avenues for policy improvement:
1)~\textit{Richer dynamics modeling.} \dybench focuses on rigid-body motion with stop-on-contact simplification. Extending to post-contact dynamics, deformable objects, and multi-target interactions would better approximate real-world conditions.
2)~\textit{Scalable data acquisition.} Dynamic demonstrations are costly due to synchronization between object motion and robot actions. Learning from human videos, simulation, or self-play could reduce this bottleneck.
3)~\textit{Inference-aware evaluation.} Incorporating realistic control delays into the benchmark would better reflect deployment conditions, where slower models fall further behind fast-moving targets.
4)~\textit{Sim-to-real transfer.} Small errors in physical parameters compound over time in dynamic settings, amplifying the sim-to-real gap beyond what standard domain adaptation addresses.

\vspace{0.3em}
\noindent We hope \dybench provides a useful foundation for the community, and that progress on dynamic manipulation will advance robot deployment in unstructured, real-world environments.

\addtocounter{figure}{-5}
\begin{figure*}[htbp]
\centering
\def\fW{0.235\textwidth}
\begin{subfigure}{\textwidth}
\centering
\includegraphics[width=\fW]{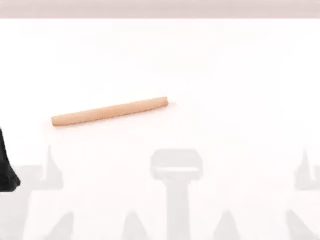}\hfill
\includegraphics[width=\fW]{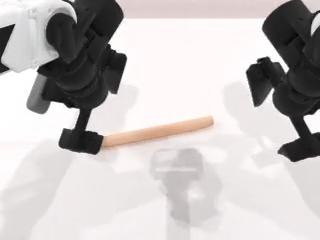}\hfill
\includegraphics[width=\fW]{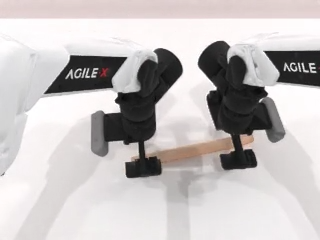}\hfill
\includegraphics[width=\fW]{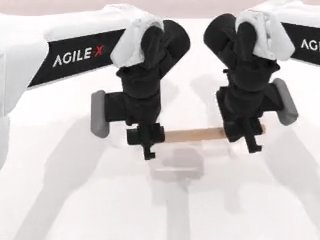}
\subcaption{Grab Roller: ``Use both arms to grab the roller on the table.''}
\end{subfigure}\\[0.4em]
\begin{subfigure}{\textwidth}
\centering
\includegraphics[width=\fW]{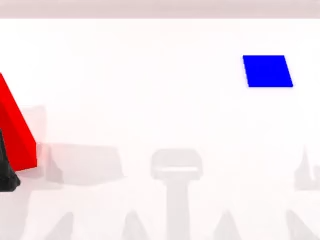}\hfill
\includegraphics[width=\fW]{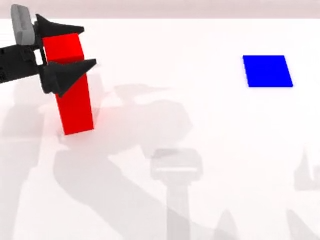}\hfill
\includegraphics[width=\fW]{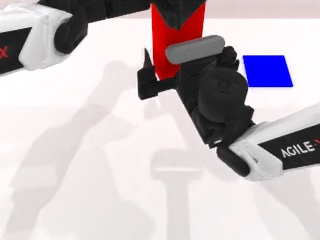}\hfill
\includegraphics[width=\fW]{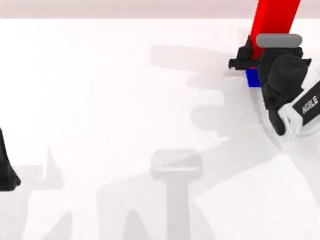}
\subcaption{Handover Block: ``Grasp the block with left arm and hand it over to the right arm.''}
\end{subfigure}\\[0.4em]
\begin{subfigure}{\textwidth}
\centering
\includegraphics[width=\fW]{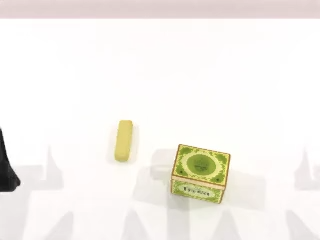}\hfill
\includegraphics[width=\fW]{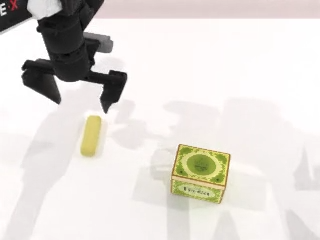}\hfill
\includegraphics[width=\fW]{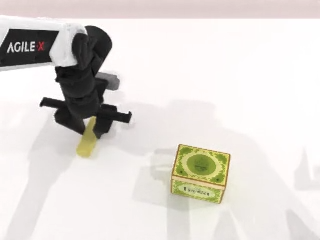}\hfill
\includegraphics[width=\fW]{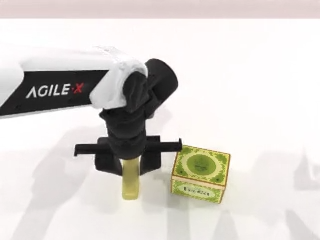}
\subcaption{Place A2B Left: ``Use appropriate arm to place object A on the left of object B.''}
\end{subfigure}\\[0.4em]
\begin{subfigure}{\textwidth}
\centering
\includegraphics[width=\fW]{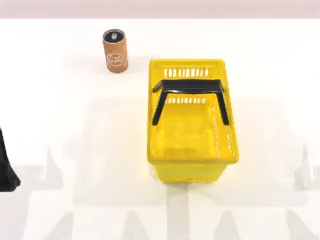}\hfill
\includegraphics[width=\fW]{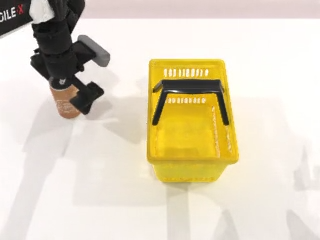}\hfill
\includegraphics[width=\fW]{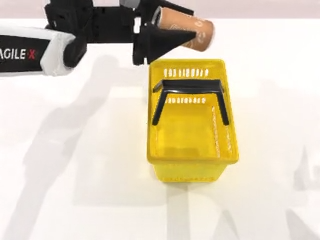}\hfill
\includegraphics[width=\fW]{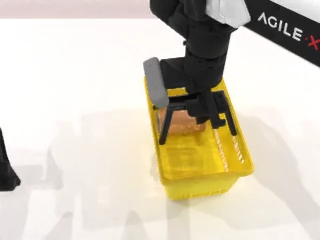}
\subcaption{Place Can Basket: ``Pick up the can, put it into the basket, and lift the basket.''}
\end{subfigure}\\[0.4em]
\begin{subfigure}{\textwidth}
\centering
\includegraphics[width=\fW]{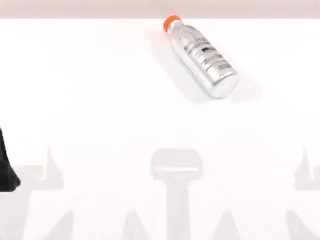}\hfill
\includegraphics[width=\fW]{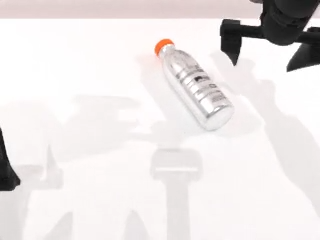}\hfill
\includegraphics[width=\fW]{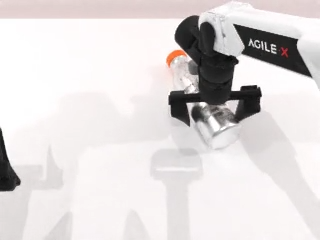}\hfill
\includegraphics[width=\fW]{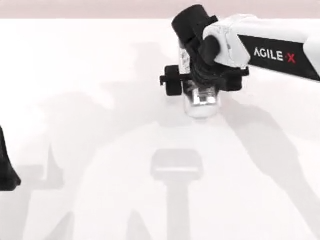}
\subcaption{Shake Bottle: ``Shake the bottle with proper arm.''}
\end{subfigure}
\caption{Simulation rollouts of \dymethod on 5 \textbf{Dynamic Interception} tasks. The robot must intercept and interact with a moving target at the right moment. Each row shows four temporally ordered keyframes from a single episode.}
\label{fig:sim_rollout_di}
\end{figure*}

\begin{figure*}[htbp]
\centering
\def\fW{0.235\textwidth}
\begin{subfigure}{\textwidth}
\centering
\includegraphics[width=\fW]{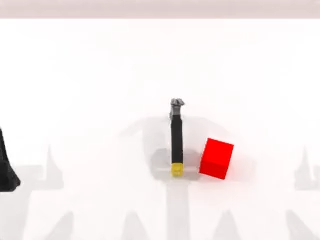}\hfill
\includegraphics[width=\fW]{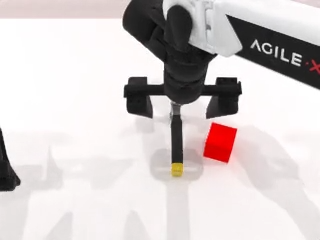}\hfill
\includegraphics[width=\fW]{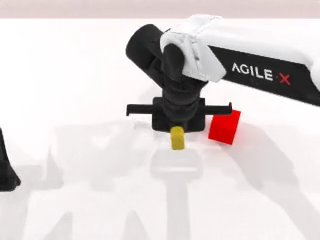}\hfill
\includegraphics[width=\fW]{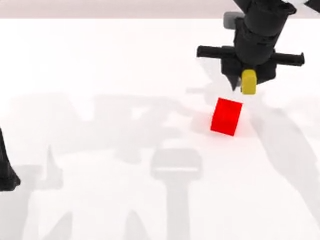}
\subcaption{Beat Block Hammer: ``Grab the hammer and beat the block.''}
\end{subfigure}\\[0.4em]
\begin{subfigure}{\textwidth}
\centering
\includegraphics[width=\fW]{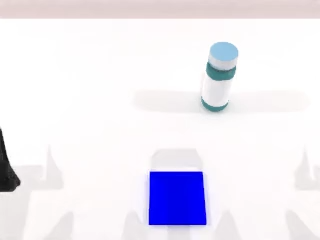}\hfill
\includegraphics[width=\fW]{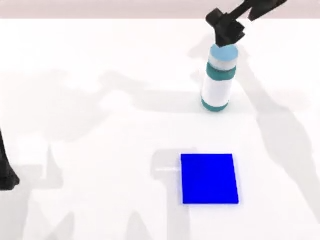}\hfill
\includegraphics[width=\fW]{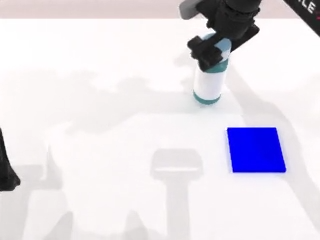}\hfill
\includegraphics[width=\fW]{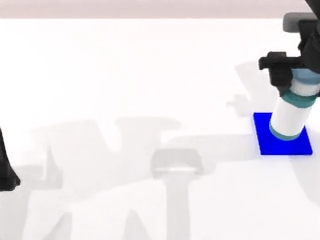}
\subcaption{Move Pillbottle Pad: ``Pick the pillbottle and place it onto the pad.''}
\end{subfigure}\\[0.4em]
\begin{subfigure}{\textwidth}
\centering
\includegraphics[width=\fW]{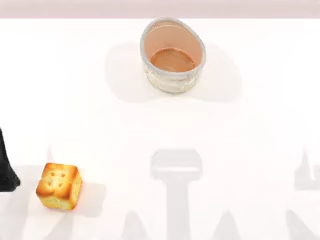}\hfill
\includegraphics[width=\fW]{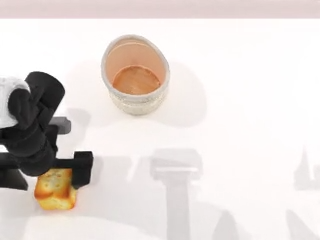}\hfill
\includegraphics[width=\fW]{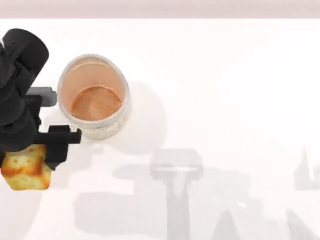}\hfill
\includegraphics[width=\fW]{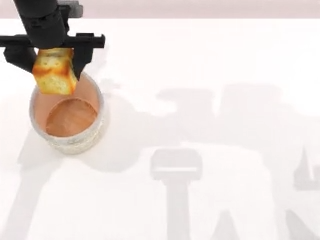}
\subcaption{Place Bread Basket: ``Grab the bread and put it in the basket.''}
\end{subfigure}\\[0.4em]
\begin{subfigure}{\textwidth}
\centering
\includegraphics[width=\fW]{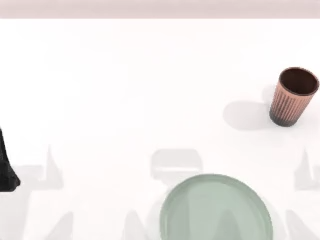}\hfill
\includegraphics[width=\fW]{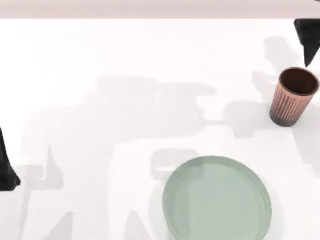}\hfill
\includegraphics[width=\fW]{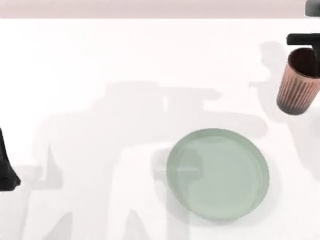}\hfill
\includegraphics[width=\fW]{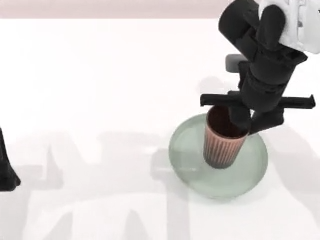}
\subcaption{Place Container Plate: ``Place the container onto the plate.''}
\end{subfigure}\\[0.4em]
\begin{subfigure}{\textwidth}
\centering
\includegraphics[width=\fW]{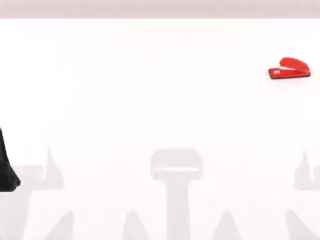}\hfill
\includegraphics[width=\fW]{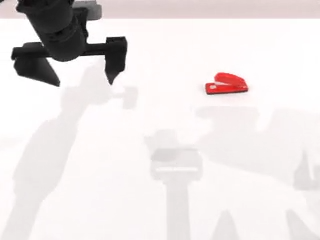}\hfill
\includegraphics[width=\fW]{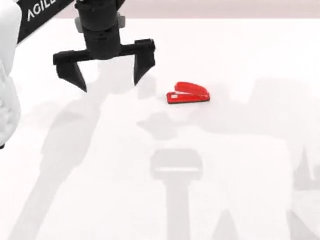}\hfill
\includegraphics[width=\fW]{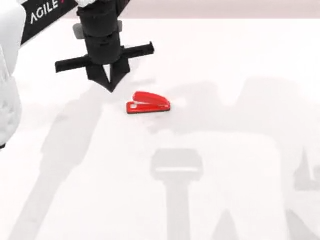}
\subcaption{Press Stapler: ``Use one arm to press the stapler.''}
\end{subfigure}
\caption{Simulation rollouts of \dymethod on 5 \textbf{Dynamic Tracking} tasks. The robot must continuously track a moving target while performing the manipulation. Each row shows four temporally ordered keyframes from a single episode.}
\label{fig:sim_rollout_dt}
\end{figure*}

\begin{figure*}[htbp]
\centering
\def\fW{0.235\textwidth}
\begin{subfigure}{\textwidth}
\centering
\includegraphics[width=\fW]{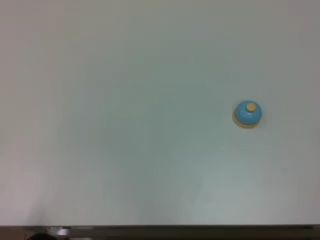}\hfill
\includegraphics[width=\fW]{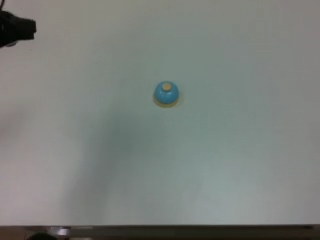}\hfill
\includegraphics[width=\fW]{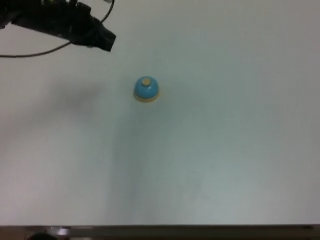}\hfill
\includegraphics[width=\fW]{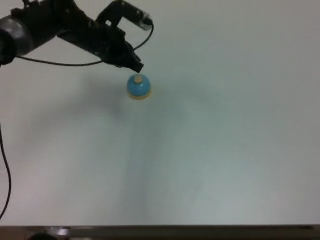}
\subcaption{Click Bell: ``Click the bell's top center on the table.''}
\end{subfigure}\\[0.4em]
\begin{subfigure}{\textwidth}
\centering
\includegraphics[width=\fW]{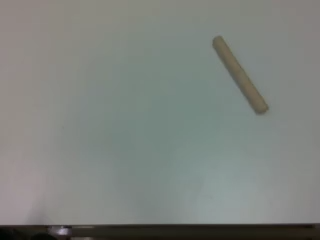}\hfill
\includegraphics[width=\fW]{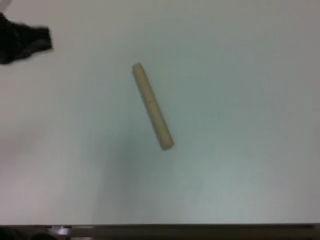}\hfill
\includegraphics[width=\fW]{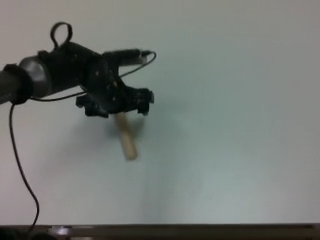}\hfill
\includegraphics[width=\fW]{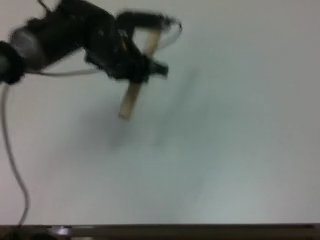}
\subcaption{Grab Roller: ``Use arm to grab the roller on the table.''}
\end{subfigure}\\[0.4em]
\begin{subfigure}{\textwidth}
\centering
\includegraphics[width=\fW]{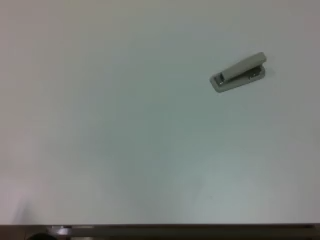}\hfill
\includegraphics[width=\fW]{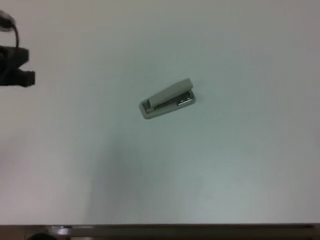}\hfill
\includegraphics[width=\fW]{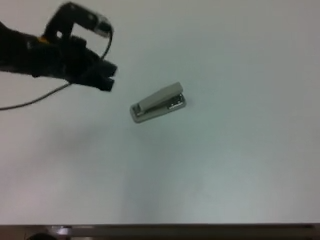}\hfill
\includegraphics[width=\fW]{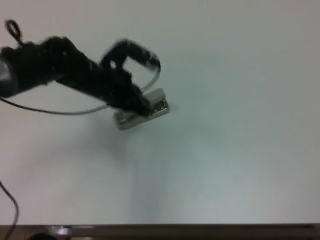}
\subcaption{Press Stapler: ``Use one arm to press the stapler.''}
\end{subfigure}\\[0.4em]
\begin{subfigure}{\textwidth}
\centering
\includegraphics[width=\fW]{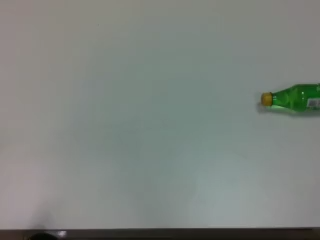}\hfill
\includegraphics[width=\fW]{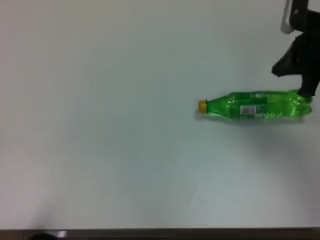}\hfill
\includegraphics[width=\fW]{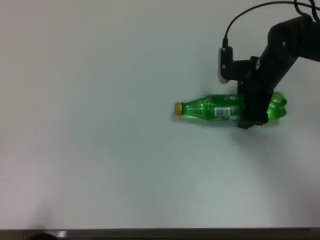}\hfill
\includegraphics[width=\fW]{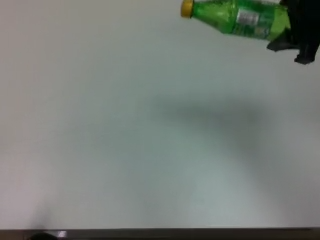}
\subcaption{Lift Bottle: ``Lift the Sprite bottle.''}
\end{subfigure}\\[0.4em]
\begin{subfigure}{\textwidth}
\centering
\includegraphics[width=\fW]{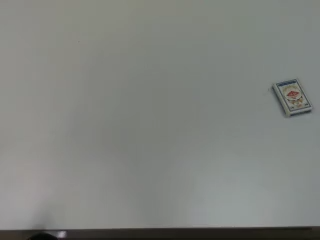}\hfill
\includegraphics[width=\fW]{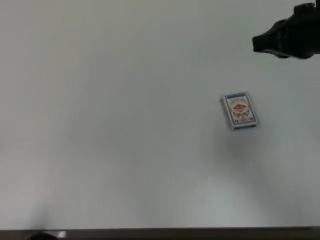}\hfill
\includegraphics[width=\fW]{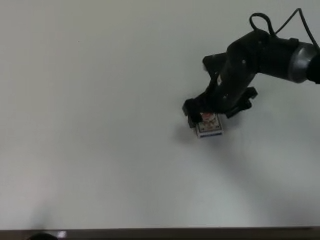}\hfill
\includegraphics[width=\fW]{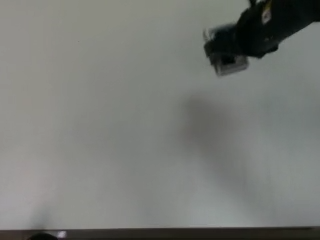}
\subcaption{Move Playingcard: ``Pick up the playing card and move it away.''}
\end{subfigure}
\vspace{-0.5em}
\caption{Real-world rollouts of \dymethod on a dual-arm PIPER robot. Target objects are driven by transparent strings to produce dynamic motion. Each row shows four keyframes from a single successful episode. Tasks (a)--(c) and (e) directly correspond to simulation tasks; task (d) is a variant of Shake Bottle adapted to the real-world setup.}
\label{fig:real_rollout}
\end{figure*}
\addtocounter{figure}{2}

\begingroup
\FloatBarrier
\scriptsize
\setlength{\LTpre}{1pt}
\setlength{\LTpost}{0pt}

\newcommand{\TThreeFirstColW}{28mm} 
\newcommand{\TThreeMetricColW}{13mm} 
\newcommand{\TThreeModelColW}{26mm}  
\begin{longtable}{@{}p{\TThreeFirstColW} *{6}{>{\centering\arraybackslash}p{\TThreeMetricColW}}@{}}
\caption{Detailed per-task performance comparison of each SOTA model on the DOMINO dataset.}
\label{tab:dy_full}\\
\noalign{\vspace{3mm}}
\toprule
\multirow{2}{*}{\textbf{Task}} 
 & \multicolumn{2}{>{\centering\arraybackslash}p{\TThreeModelColW}}{\textbf{RDT-1~\cite{liu2024rdt}}}
 & \multicolumn{2}{>{\centering\arraybackslash}p{\TThreeModelColW}}{\textbf{OpenVLA~\cite{kim2024openvla}}}
 & \multicolumn{2}{>{\centering\arraybackslash}p{\TThreeModelColW}}{\textbf{OpenVLA-OFT~\cite{kim2025openvlaoft}}} \\
\cmidrule(lr){2-3} \cmidrule(lr){4-5} \cmidrule(lr){6-7}
& SR$\uparrow$ & MS$\uparrow$ 
& SR$\uparrow$ & MS$\uparrow$ 
& SR$\uparrow$ & MS$\uparrow$ \\
\midrule
\endfirsthead

\multicolumn{7}{c}{\tablename\ \thetable\ -- \textit{Continued from previous page}}\\
\toprule
\multirow{2}{*}{\textbf{Task}} 
 & \multicolumn{2}{>{\centering\arraybackslash}p{\TThreeModelColW}}{\textbf{RDT-1~\cite{liu2024rdt}}}
 & \multicolumn{2}{>{\centering\arraybackslash}p{\TThreeModelColW}}{\textbf{OpenVLA~\cite{kim2024openvla}}}
 & \multicolumn{2}{>{\centering\arraybackslash}p{\TThreeModelColW}}{\textbf{OpenVLA-OFT~\cite{kim2025openvlaoft}}} \\
\cmidrule(lr){2-3} \cmidrule(lr){4-5} \cmidrule(lr){6-7}
& SR$\uparrow$ & MS$\uparrow$ 
& SR$\uparrow$ & MS$\uparrow$ 
& SR$\uparrow$ & MS$\uparrow$ \\
\midrule
\endhead

\midrule
\multicolumn{7}{r}{\textit{Continued on next page}}\\
\endfoot

\bottomrule
\endlastfoot

\textit{Adjust Bottle} & 18.00 & 23.32 & 0.00 & 1.73 & 58.00 & 63.57\\
\textit{Beat Block Hammer} & 15.00 & 25.08 & 0.00 & 4.23 & 11.00 & 23.20\\
\textit{Click Alarmclock} & 10.00 & 13.55 & 4.00 & 6.02 & 6.00 & 10.53\\
\textit{Click Bell} & 0.00 & 6.07 & 1.00 & 3.29 & 3.00 & 11.24\\
\textit{Dump Bin Bigbin} & 0.00 & 10.96 & 0.00 & 2.43 & 0.00 & 17.84\\
\addlinespace
\textit{Grab Roller} & 12.00 & 27.27 & 0.00 & 3.21 & 28.00 & 43.87\\
\textit{Handover Block} & 2.00 & 26.99 & 0.00 & 7.02 & 9.00 & 61.17\\
\textit{Handover Mic} & 8.00 & 15.16 & 0.00 & 2.34 & 25.00 & 37.58\\
\textit{Hanging Mug} & 7.00 & 18.32 & 0.00 & 3.58 & 12.00 & 35.49\\
\textit{Move Can Pot} & 7.00 & 40.88 & 0.00 & 6.80 & 29.00 & 57.26\\
\addlinespace
\textit{Move Pillbottle Pad} & 0.00 & 9.69 & 0.00 & 1.86 & 1.00 & 9.93\\
\textit{Move P.Card Away} & 1.00 & 9.82 & 0.00 & 3.26 & 1.00 & 7.94\\
\textit{Move Stapler Pad} & 0.00 & 11.33 & 0.00 & 3.54 & 0.00 & 10.87\\
\textit{Place A2B Left} & 0.00 & 22.92 & 0.00 & 3.10 & 1.00 & 22.37\\
\textit{Place A2B Right} & 0.00 & 21.27 & 1.00 & 6.14 & 1.00 & 25.00\\
\addlinespace
\textit{Place Bread Basket} & 2.00 & 8.77 & 0.00 & 4.45 & 1.00 & 7.98\\
\textit{Place Bread Skillet} & 4.00 & 15.20 & 0.00 & 7.31 & 1.00 & 20.32\\
\textit{Place Can Basket} & 10.00 & 30.84 & 0.00 & 12.21 & 5.00 & 37.78\\
\textit{Place Container Plate} & 1.00 & 12.90 & 0.00 & 4.70 & 10.00 & 16.87\\
\textit{Place Empty Cup} & 2.00 & 7.96 & 0.00 & 2.98 & 0.00 & 7.74\\
\addlinespace
\textit{Place Fan} & 1.00 & 8.62 & 0.00 & 6.04 & 1.00 & 6.25\\
\textit{Place Mouse Pad} & 0.00 & 10.20 & 0.00 & 3.08 & 1.00 & 13.61\\
\textit{Place Object Basket} & 11.00 & 25.20 & 0.00 & 6.88 & 18.00 & 37.36\\
\textit{Place Object Scale} & 1.00 & 13.07 & 0.00 & 3.92 & 0.00 & 10.92\\
\textit{Place Object Stand} & 0.00 & 7.91 & 0.00 & 7.55 & 0.00 & 8.53\\
\addlinespace
\textit{Place Phone Stand} & 1.00 & 13.11 & 0.00 & 3.16 & 0.00 & 12.66\\
\textit{Place Shoe} & 5.00 & 22.66 & 0.00 & 4.45 & 4.00 & 17.39\\
\textit{Press Stapler} & 11.00 & 17.30 & 25.00 & 27.12 & 5.00 & 13.53\\
\textit{Put Bottles Dustbin} & 12.00 & 21.39 & 1.00 & 7.25 & 19.00 & 29.81\\
\textit{Put Object Cabinet} & 12.00 & 36.39 & 0.00 & 6.54 & 1.00 & 42.07\\
\addlinespace
\textit{Rotate QRcode} & 4.00 & 12.05 & 0.00 & 3.67 & 11.00 & 23.31\\
\textit{Scan Object} & 0.00 & 14.08 & 0.00 & 9.69 & 4.00 & 19.93\\
\textit{Shake Bottle} & 15.00 & 21.38 & 14.00 & 18.49 & 17.00 & 23.83\\
\textit{Shake Bottle Horiz.} & 14.00 & 24.18 & 8.00 & 12.33 & 31.00 & 38.75\\
\textit{Stamp Seal} & 1.00 & 13.93 & 0.00 & 3.13 & 3.00 & 15.47\\
\midrule
\textbf{\textit{Average}} & 5.34 & 17.71 & 1.54 & 6.10 & 9.06 & 24.06\\
\end{longtable}%

\addtocounter{table}{-1}%
\begin{longtable}{@{}p{\TThreeFirstColW} *{6}{>{\centering\arraybackslash}p{\TThreeMetricColW}}@{}}
\toprule
\multirow{2}{*}{\textbf{Task}} 
 & \multicolumn{2}{>{\centering\arraybackslash}p{\TThreeModelColW}}{\textbf{$\mathbf{\pi}_{\mathbf{0}}$-FAST~\cite{pertsch2025fast}}}
 & \multicolumn{2}{>{\centering\arraybackslash}p{\TThreeModelColW}}{\textbf{VLA-Adapter~\cite{wang2025vlaadapter}}}
 & \multicolumn{2}{>{\centering\arraybackslash}p{\TThreeModelColW}}{\textbf{InternVLA-M1~\cite{internvlam1}}} \\
\cmidrule(lr){2-3} \cmidrule(lr){4-5} \cmidrule(lr){6-7}
& SR$\uparrow$ & MS$\uparrow$ 
& SR$\uparrow$ & MS$\uparrow$ 
& SR$\uparrow$ & MS$\uparrow$ \\
\midrule
\endfirsthead

\multicolumn{7}{c}{\tablename\ \thetable\ -- \textit{Continued from previous page}}\\
\toprule
\multirow{2}{*}{\textbf{Task}} 
 & \multicolumn{2}{>{\centering\arraybackslash}p{\TThreeModelColW}}{\textbf{$\mathbf{\pi}_{\mathbf{0}}$-FAST~\cite{pertsch2025fast}}}
 & \multicolumn{2}{>{\centering\arraybackslash}p{\TThreeModelColW}}{\textbf{VLA-Adapter~\cite{wang2025vlaadapter}}}
 & \multicolumn{2}{>{\centering\arraybackslash}p{\TThreeModelColW}}{\textbf{InternVLA-M1~\cite{internvlam1}}} \\
\cmidrule(lr){2-3} \cmidrule(lr){4-5} \cmidrule(lr){6-7}
& SR$\uparrow$ & MS$\uparrow$ 
& SR$\uparrow$ & MS$\uparrow$ 
& SR$\uparrow$ & MS$\uparrow$ \\
\midrule
\endhead

\midrule
\multicolumn{7}{r}{\textit{Continued on next page}}\\
\endfoot

\bottomrule
\endlastfoot

\textit{Adjust Bottle} & 11.00 & 37.09 & 9.00 & 38.07 & 20.00 & 62.02\\
\textit{Beat Block Hammer} & 1.00 & 18.13 & 1.00 & 15.71 & 0.00 & 17.18\\
\textit{Click Alarmclock} & 4.00 & 10.63 & 10.00 & 17.00 & 9.00 & 17.12\\
\textit{Click Bell} & 0.00 & 9.32 & 1.00 & 11.56 & 2.00 & 15.30\\
\textit{Dump Bin Bigbin} & 0.00 & 13.37 & 0.00 & 23.93 & 0.00 & 30.52\\
\addlinespace
\textit{Grab Roller} & 0.00 & 15.59 & 30.00 & 59.19 & 33.00 & 60.47\\
\textit{Handover Block} & 0.00 & 25.35 & 1.00 & 54.15 & 0.00 & 38.71\\
\textit{Handover Mic} & 0.00 & 21.61 & 0.00 & 19.06 & 2.00 & 33.10\\
\textit{Hanging Mug} & 0.00 & 13.39 & 0.00 & 30.31 & 0.00 & 28.51\\
\textit{Move Can Pot} & 9.00 & 49.88 & 0.00 & 24.25 & 0.00 & 26.55\\
\addlinespace
\textit{Move Pillbottle Pad} & 6.00 & 19.19 & 0.00 & 13.24 & 0.00 & 15.41\\
\textit{Move P.Card Away} & 0.00 & 20.00 & 0.00 & 22.26 & 0.00 & 24.19\\
\textit{Move Stapler Pad} & 0.00 & 17.12 & 0.00 & 13.46 & 0.00 & 14.33\\
\textit{Place A2B Left} & 1.00 & 28.69 & 0.00 & 23.02 & 0.00 & 25.63\\
\textit{Place A2B Right} & 0.00 & 27.23 & 0.00 & 22.11 & 0.00 & 27.24\\
\addlinespace
\textit{Place Bread Basket} & 2.00 & 17.95 & 1.00 & 15.04 & 0.00 & 13.30\\
\textit{Place Bread Skillet} & 0.00 & 13.39 & 13.00 & 35.87 & 12.00 & 35.34\\
\textit{Place Can Basket} & 0.00 & 24.49 & 0.00 & 21.33 & 0.00 & 21.14\\
\textit{Place Container Plate} & 26.00 & 35.47 & 0.00 & 7.50 & 1.00 & 16.56\\
\textit{Place Empty Cup} & 2.00 & 12.11 & 0.00 & 8.41 & 0.00 & 14.37\\
\addlinespace
\textit{Place Fan} & 0.00 & 10.93 & 0.00 & 15.14 & 0.00 & 12.14\\
\textit{Place Mouse Pad} & 0.00 & 18.23 & 0.00 & 13.74 & 0.00 & 15.15\\
\textit{Place Object Basket} & 0.00 & 25.51 & 0.00 & 23.10 & 1.00 & 26.16\\
\textit{Place Object Scale} & 0.00 & 15.18 & 0.00 & 14.31 & 0.00 & 14.61\\
\textit{Place Object Stand} & 2.00 & 11.97 & 0.00 & 12.40 & 0.00 & 17.41\\
\addlinespace
\textit{Place Phone Stand} & 0.00 & 19.25 & 1.00 & 14.46 & 0.00 & 16.26\\
\textit{Place Shoe} & 3.00 & 16.13 & 0.00 & 12.53 & 0.00 & 17.87\\
\textit{Press Stapler} & 7.00 & 16.30 & 20.00 & 29.49 & 17.00 & 28.98\\
\textit{Put Bottles Dustbin} & 3.00 & 18.76 & 16.00 & 34.84 & 18.00 & 39.65\\
\textit{Put Object Cabinet} & 1.00 & 24.13 & 2.00 & 38.14 & 7.00 & 47.28\\
\addlinespace
\textit{Rotate QRcode} & 0.00 & 16.75 & 0.00 & 19.19 & 0.00 & 25.05\\
\textit{Scan Object} & 0.00 & 13.77 & 0.00 & 27.82 & 0.00 & 33.85\\
\textit{Shake Bottle} & 18.00 & 30.88 & 27.00 & 51.16 & 39.00 & 58.41\\
\textit{Shake Bottle Horiz.} & 26.00 & 41.11 & 23.00 & 51.98 & 28.00 & 56.48\\
\textit{Stamp Seal} & 2.00 & 21.38 & 0.00 & 17.10 & 0.00 & 18.82\\
\midrule
\textbf{\textit{Average}} & 3.54 & 20.87 & 4.40 & 24.31 & 5.40 & 27.57\\
\end{longtable}%

\addtocounter{table}{-1}%
\begin{longtable}{@{}p{\TThreeFirstColW} *{6}{>{\centering\arraybackslash}p{\TThreeMetricColW}}@{}}
\toprule
\multirow{2}{*}{\textbf{Task}} 
 & \multicolumn{2}{>{\centering\arraybackslash}p{\TThreeModelColW}}{\textbf{$\mathbf{\pi}_{\mathbf{0}}$~\cite{pi0}}}
 & \multicolumn{2}{>{\centering\arraybackslash}p{\TThreeModelColW}}{\textbf{$\mathbf{\pi}_{\mathbf{0.5}}$~\cite{pi05}}}
 & \multicolumn{2}{>{\centering\arraybackslash}p{\TThreeModelColW}}{\textbf{OpenVLA-OFT*}} \\
\cmidrule(lr){2-3} \cmidrule(lr){4-5} \cmidrule(lr){6-7}
& SR$\uparrow$ & MS$\uparrow$ 
& SR$\uparrow$ & MS$\uparrow$ 
& SR$\uparrow$ & MS$\uparrow$ \\
\midrule
\endfirsthead

\multicolumn{7}{c}{\tablename\ \thetable\ -- \textit{Continued from previous page}}\\
\toprule
\multirow{2}{*}{\textbf{Task}} 
 & \multicolumn{2}{>{\centering\arraybackslash}p{\TThreeModelColW}}{\textbf{$\mathbf{\pi}_{\mathbf{0}}$~\cite{pi0}}}
 & \multicolumn{2}{>{\centering\arraybackslash}p{\TThreeModelColW}}{\textbf{$\mathbf{\pi}_{\mathbf{0.5}}$~\cite{pi05}}}
 & \multicolumn{2}{>{\centering\arraybackslash}p{\TThreeModelColW}}{\textbf{OpenVLA-OFT*}} \\
\cmidrule(lr){2-3} \cmidrule(lr){4-5} \cmidrule(lr){6-7}
& SR$\uparrow$ & MS$\uparrow$ 
& SR$\uparrow$ & MS$\uparrow$ 
& SR$\uparrow$ & MS$\uparrow$ \\
\midrule
\endhead

\midrule
\multicolumn{7}{r}{\textit{Continued on next page}}\\
\endfoot

\bottomrule
\endlastfoot

\textit{Adjust Bottle} & 19.00 & 46.89 & 52.00 & 71.45 & 44.00 & 67.64\\
\textit{Beat Block Hammer} & 10.00 & 23.30 & 10.00 & 23.22 & 9.00 & 22.47\\
\textit{Click Alarmclock} & 15.00 & 18.95 & 14.00 & 18.47 & 5.00 & 10.07\\
\textit{Click Bell} & 0.00 & 8.04 & 0.00 & 6.35 & 1.00 & 11.56\\
\textit{Dump Bin Bigbin} & 0.00 & 20.88 & 0.00 & 16.64 & 0.00 & 29.44\\
\addlinespace
\textit{Grab Roller} & 11.00 & 31.46 & 10.00 & 34.32 & 31.00 & 53.45\\
\textit{Handover Block} & 0.00 & 39.21 & 1.00 & 33.95 & 0.00 & 44.75\\
\textit{Handover Mic} & 4.00 & 26.18 & 5.00 & 29.10 & 21.00 & 50.93\\
\textit{Hanging Mug} & 9.00 & 26.03 & 5.00 & 25.59 & 0.00 & 28.51\\
\textit{Move Can Pot} & 16.00 & 41.72 & 7.00 & 40.34 & 13.00 & 37.07\\
\addlinespace
\textit{Move Pillbottle Pad} & 7.00 & 16.53 & 6.00 & 15.93 & 7.00 & 15.81\\
\textit{Move P.Card Away} & 7.00 & 21.53 & 8.00 & 24.84 & 0.00 & 27.98\\
\textit{Move Stapler Pad} & 1.00 & 11.96 & 0.00 & 13.38 & 0.00 & 14.18\\
\textit{Place A2B Left} & 2.00 & 21.81 & 2.00 & 26.04 & 1.00 & 34.14\\
\textit{Place A2B Right} & 1.00 & 20.08 & 1.00 & 23.63 & 4.00 & 33.38\\
\addlinespace
\textit{Place Bread Basket} & 6.00 & 11.67 & 8.00 & 15.89 & 7.00 & 22.30\\
\textit{Place Bread Skillet} & 2.00 & 16.11 & 9.00 & 22.30 & 19.00 & 38.30\\
\textit{Place Can Basket} & 8.00 & 37.49 & 6.00 & 29.48 & 0.00 & 29.17\\
\textit{Place Container Plate} & 15.00 & 21.79 & 22.00 & 28.46 & 8.00 & 17.02\\
\textit{Place Empty Cup} & 0.00 & 8.05 & 2.00 & 12.71 & 1.00 & 11.59\\
\addlinespace
\textit{Place Fan} & 2.00 & 9.91 & 2.00 & 11.72 & 0.00 & 6.54\\
\textit{Place Mouse Pad} & 4.00 & 14.93 & 5.00 & 19.03 & 1.00 & 17.29\\
\textit{Place Object Basket} & 16.00 & 33.55 & 11.00 & 30.19 & 6.00 & 35.01\\
\textit{Place Object Scale} & 0.00 & 9.71 & 3.00 & 15.37 & 0.00 & 15.06\\
\textit{Place Object Stand} & 1.00 & 8.00 & 1.00 & 8.97 & 1.00 & 13.68\\
\addlinespace
\textit{Place Phone Stand} & 0.00 & 13.11 & 2.00 & 17.98 & 0.00 & 17.86\\
\textit{Place Shoe} & 20.00 & 32.11 & 25.00 & 32.84 & 6.00 & 19.36\\
\textit{Press Stapler} & 9.00 & 18.77 & 6.00 & 13.27 & 9.00 & 17.64\\
\textit{Put Bottles Dustbin} & 20.00 & 37.01 & 17.00 & 33.30 & 43.00 & 56.41\\
\textit{Put Object Cabinet} & 11.00 & 41.38 & 11.00 & 46.24 & 26.00 & 59.78\\
\addlinespace
\textit{Rotate QRcode} & 6.00 & 15.90 & 4.00 & 17.90 & 3.00 & 25.08\\
\textit{Scan Object} & 0.00 & 19.64 & 1.00 & 20.48 & 4.00 & 31.66\\
\textit{Shake Bottle} & 28.00 & 44.02 & 33.00 & 49.40 & 53.00 & 67.42\\
\textit{Shake Bottle Horiz.} & 33.00 & 53.03 & 42.00 & 62.72 & 52.00 & 65.78\\
\textit{Stamp Seal} & 3.00 & 17.75 & 6.00 & 22.58 & 5.00 & 18.69\\
\midrule
\textbf{\textit{Average}} & 8.17 & 23.96 & 9.63 & 26.17 & 10.86 & 30.49\\
\end{longtable}%

\addtocounter{table}{-1}%
\begin{longtable}{@{}p{\TThreeFirstColW} *{6}{>{\centering\arraybackslash}p{\TThreeMetricColW}}@{}}
\toprule
\multirow{2}{*}{\textbf{Task}} 
 & \multicolumn{2}{>{\centering\arraybackslash}p{\TThreeModelColW}}{\textbf{Isaac-GR00T*}}
 & \multicolumn{2}{>{\centering\arraybackslash}p{\TThreeModelColW}}{\textbf{$\mathbf{\pi}_{\mathbf{0}}$-FAST*}}
 & \multicolumn{2}{>{\centering\arraybackslash}p{\TThreeModelColW}}{\textbf{PUMA(OURS)}} \\
\cmidrule(lr){2-3} \cmidrule(lr){4-5} \cmidrule(lr){6-7}
& SR$\uparrow$ & MS$\uparrow$ 
& SR$\uparrow$ & MS$\uparrow$ 
& SR$\uparrow$ & MS$\uparrow$ \\
\midrule
\endfirsthead

\multicolumn{7}{c}{\tablename\ \thetable\ -- \textit{Continued from previous page}}\\
\toprule
\multirow{2}{*}{\textbf{Task}} 
 & \multicolumn{2}{>{\centering\arraybackslash}p{\TThreeModelColW}}{\textbf{Isaac-GR00T*}}
 & \multicolumn{2}{>{\centering\arraybackslash}p{\TThreeModelColW}}{\textbf{$\mathbf{\pi}_{\mathbf{0}}$-FAST*}}
 & \multicolumn{2}{>{\centering\arraybackslash}p{\TThreeModelColW}}{\textbf{PUMA(OURS)}} \\
\cmidrule(lr){2-3} \cmidrule(lr){4-5} \cmidrule(lr){6-7}
& SR$\uparrow$ & MS$\uparrow$ 
& SR$\uparrow$ & MS$\uparrow$ 
& SR$\uparrow$ & MS$\uparrow$ \\
\midrule
\endhead

\midrule
\multicolumn{7}{r}{\textit{Continued on next page}}\\
\endfoot

\bottomrule
\endlastfoot

\textit{Adjust Bottle} & 19.00 & 50.82 & 8.00 & 33.15 & 65.00 & 74.04\\
\textit{Beat Block Hammer} & 3.00 & 21.53 & 4.00 & 17.87 & 15.00 & 29.09\\
\textit{Click Alarmclock} & 11.00 & 18.50 & 18.00 & 20.75 & 4.00 & 8.38\\
\textit{Click Bell} & 2.00 & 14.82 & 1.00 & 7.62 & 3.00 & 13.64\\
\textit{Dump Bin Bigbin} & 0.00 & 26.91 & 0.00 & 12.91 & 0.00 & 39.05\\
\addlinespace
\textit{Grab Roller} & 28.00 & 53.56 & 20.00 & 46.32 & 33.00 & 56.48\\
\textit{Handover Block} & 0.00 & 44.88 & 2.00 & 31.34 & 17.00 & 64.79\\
\textit{Handover Mic} & 3.00 & 35.52 & 5.00 & 23.84 & 35.00 & 54.40\\
\textit{Hanging Mug} & 0.00 & 32.21 & 1.00 & 20.06 & 9.00 & 41.74\\
\textit{Move Can Pot} & 1.00 & 35.09 & 9.00 & 35.02 & 22.00 & 47.80\\
\addlinespace
\textit{Move Pillbottle Pad} & 0.00 & 12.94 & 1.00 & 9.67 & 14.00 & 21.82\\
\textit{Move P.Card Away} & 0.00 & 25.95 & 6.00 & 29.73 & 6.00 & 27.90\\
\textit{Move Stapler Pad} & 0.00 & 15.97 & 0.00 & 12.71 & 1.00 & 16.24\\
\textit{Place A2B Left} & 0.00 & 27.02 & 3.00 & 25.02 & 13.00 & 41.65\\
\textit{Place A2B Right} & 0.00 & 26.63 & 1.00 & 23.83 & 8.00 & 34.39\\
\addlinespace
\textit{Place Bread Basket} & 1.00 & 18.65 & 0.00 & 12.53 & 12.00 & 22.44\\
\textit{Place Bread Skillet} & 10.00 & 30.34 & 2.00 & 26.15 & 19.00 & 35.95\\
\textit{Place Can Basket} & 0.00 & 30.30 & 0.00 & 22.91 & 14.00 & 45.79\\
\textit{Place Container Plate} & 4.00 & 19.55 & 7.00 & 16.85 & 26.00 & 34.45\\
\textit{Place Empty Cup} & 0.00 & 15.68 & 0.00 & 8.99 & 7.00 & 17.85\\
\addlinespace
\textit{Place Fan} & 2.00 & 11.88 & 1.00 & 6.87 & 8.00 & 14.36\\
\textit{Place Mouse Pad} & 0.00 & 17.91 & 0.00 & 14.04 & 2.00 & 17.86\\
\textit{Place Object Basket} & 1.00 & 31.11 & 7.00 & 21.74 & 13.00 & 41.67\\
\textit{Place Object Scale} & 0.00 & 16.87 & 2.00 & 13.50 & 4.00 & 16.05\\
\textit{Place Object Stand} & 1.00 & 17.65 & 0.00 & 10.25 & 1.00 & 10.66\\
\addlinespace
\textit{Place Phone Stand} & 0.00 & 18.84 & 1.00 & 15.28 & 6.00 & 19.02\\
\textit{Place Shoe} & 0.00 & 17.21 & 9.00 & 19.82 & 16.00 & 27.23\\
\textit{Press Stapler} & 9.00 & 22.64 & 25.00 & 32.76 & 10.00 & 18.40\\
\textit{Put Bottles Dustbin} & 13.00 & 33.42 & 3.00 & 20.39 & 23.00 & 36.82\\
\textit{Put Object Cabinet} & 14.00 & 50.22 & 2.00 & 13.11 & 34.00 & 61.64\\
\addlinespace
\textit{Rotate QRcode} & 0.00 & 28.08 & 0.00 & 11.72 & 14.00 & 29.27\\
\textit{Scan Object} & 3.00 & 30.69 & 0.00 & 20.21 & 5.00 & 30.58\\
\textit{Shake Bottle} & 43.00 & 60.65 & 28.00 & 33.48 & 55.00 & 65.37\\
\textit{Shake Bottle Horiz.} & 46.00 & 67.33 & 34.00 & 39.33 & 75.00 & 80.57\\
\textit{Stamp Seal} & 0.00 & 19.65 & 1.00 & 13.23 & 13.00 & 26.50\\
\midrule
\textbf{\textit{Average}} & 6.10 & 28.60 & 5.74 & 20.66 & 17.20 & 34.97\\
\end{longtable}
\leftline{\tiny{\qquad* Indicates that the method is implemented using Qwen3-VL~\cite{starvla2025}.}}
\endgroup
\begin{table}[ht]
  \centering
  \scriptsize
  \caption{\textbf{Detailed quantitative evaluation of VLA models in the static setting ($S \to S$).} This table reports the per-task performance across all 35 tasks for models trained and evaluated in the static environment. These results expand upon the \textit{Static} columns in Tab. 1 of the main text.}
  \label{tab:robotwin-ss}
  \vspace{-5pt}
  \setlength{\tabcolsep}{3pt}
  \begin{tabular}{p{2.8cm} *{6}{>{\centering\arraybackslash}p{1.1cm}}}
  \toprule
  \multirow{3}{*}{\textbf{Simulation Task}}
    & \multicolumn{2}{c}{{\textbf{ACT}~\cite{zhao2023act}}}
    & \multicolumn{2}{c}{{\textbf{OpenVLA-OFT}~\cite{kim2025openvlaoft}}}
    & \multicolumn{2}{c}{{$\mathbf{\pi}_{\mathbf{0.5}}$~\cite{pi05}}} \\
  \cmidrule(lr){2-3} \cmidrule(lr){4-5} \cmidrule(lr){6-7}
  & SR & MS & SR & MS & SR & MS\\
  \midrule
  \textit{Adjust Bottle}
    & 97\% & 97.00
    & 47\% & 47.00
    & 92\% & 92.00 \\
  \textit{Beat Block Hammer}
    & 53\% & 53.00
    & 12\% & 12.00
    & 55\% & 55.00 \\
  \textit{Click Alarmclock}
    & 29\% & 29.00
    & 57\% & 57.00
    & 48\% & 48.00 \\
  \textit{Click Bell}
    & 59\% & 59.00
    & 24\% & 24.00
    & 58\% & 58.00 \\
  \textit{Dump Bin Bigbin}
    & 66\% & 66.00
    & 18\% & 18.00
    & 73\% & 73.00 \\
  \addlinespace
  \textit{Grab Roller}
    & 95\% & 95.00
    & 47\% & 47.00
    & 98\% & 98.00 \\
  \textit{Handover Block}
    & 43\% & 43.00
    & 3\% & 3.00
    & 2\% & 2.00 \\
  \textit{Handover Mic}
    & 82.00 & 82.00
    & 93\% & 93.00
    & 63\% & 63.00 \\
  \textit{Hanging Mug}
    & 13\% & 13.00
    & 1\% & 1.00
    & 3\% & 3.00 \\
  \textit{Move Can Pot}
    & 23\% & 23.00
    & 34\% & 34.00
    & 14\% & 14.00 \\
  \addlinespace
  \textit{Move Pillbottle Pad}
    & 1\% & 1.00
    & 1\% & 1.00
    & 20\% & 20.00 \\
  \textit{Move P.Card Away}
    & 34\% & 34.00
    & 6\% & 6.00
    & 74\% & 74.00 \\
  \textit{Move Stapler Pad}
    & 0\% & 0.00
    & 0\% & 0.00
    & 8\% & 8.00 \\
  \textit{Place A2B Left}
    & 0\% & 0.00
    & 2\% & 2.00
    & 39\% & 39.00 \\
  \textit{Place A2B Right}
    & 0\% & 0.00
    & 0\% & 0.00
    & 38\% & 38.00 \\
  \textit{Place Bread Basket}
    & 4\% & 4.00
    & 6\% & 6.00
    & 42\% & 42.00 \\
   \addlinespace
  \textit{Place Bread Skillet}
    & 9\% & 9.00
    & 4\% & 4.00
    & 39\% & 39.00 \\
  \textit{Place Can Basket}
    & 1\% & 1.00
    & 8\% & 8.00
    & 17\% & 17.00 \\
  \textit{Place Container Plate}
    & 72\% & 72.00
    & 27\% & 27.00
    & 90\% & 90.00 \\
  \textit{Place Empty Cup}
    & 58\% & 58.00
    & 1\% & 1.00
    & 68\% & 68.00 \\
  \textit{Place Fan}
    & 0\% & 0.00
    & 3\% & 3.00
    & 30\% & 30.00 \\
   \addlinespace
  \textit{Place Mouse Pad}
    & 0\% & 0.00
    & 0\% & 0.00
    & 22\% & 22.00 \\
  \textit{Place Object Basket}
    & 18\% & 18.00
    & 19\% & 19.00
    & 57\% & 57.00 \\
  \textit{Place Object Scale}
    & 0\% & 0.00
    & 0\% & 0.00
    & 41\% & 41.00 \\
  \textit{Place Object Stand}
    & 0\% & 0.00
    & 17\% & 17.00
    & 57\% & 57.00 \\
  \textit{Place Phone Stand}
    & 1\% & 1.00
    & 3\% & 3.00
    & 37\% & 37.00 \\
   \addlinespace
  \textit{Place Shoe}
    & 3\% & 3.00
    & 3\% & 3.00
    & 39\% & 39.00 \\
  \textit{Press Stapler}
    & 31\% & 31.00
    & 35\% & 35.00
    & 35\% & 35.00 \\
  \textit{Put Bottles Dustbin}
    & 28\% & 28.00
    & 1\% & 1.00
    & 10\% & 10.00 \\
  \textit{Put Object Cabinet}
    & 9\% & 9.00
    & 11\% & 11.00
    & 36\% & 36.00 \\
  \textit{Rotate QRcode}
    & 3\% & 3.00
    & 10\% & 10.0
    & 54\% & 54.0 \\
   \addlinespace
  \textit{Scan Object}
    & 1\% & 1.00
    & 0\% & 0.00
    & 7\% & 7.00 \\
  \textit{Shake Bottle}
    & 74\% & 74.00
    & 63\% & 63.00
    & 95\% & 95.00 \\
  \textit{Shake Bottle Horiz.}
    & 63\% & 63.00
    & 57\% & 57.00
    & 97\% & 97.00 \\
  \textit{Stamp Seal}
    & 1\% & 1.00
    & 0\% & 0.00
    & 11\% & 11.00 \\
  \midrule
  \textbf{\textit{Average (\%)}}
    & 27.74 & 27.74
    & 17.51 & 17.51
    & 44.83 & 44.83 \\
  \bottomrule
  \end{tabular}
  \vspace{-0.3cm}
\end{table}
\begin{table}[ht]
  \centering
  \scriptsize
  \caption{\textbf{Detailed quantitative evaluation of VLA models in the zero-shot dynamic setting ($S \to D$).} This table reports the per-task performance across all 35 tasks for models trained in a static environment and directly evaluated in our proposed DOMINO without fine-tuning. These results expand upon the \textit{ZS} columns in Tab. 1 of the main text.}
  \label{tab:robotwin-sd}
  \vspace{-5pt}
  \setlength{\tabcolsep}{3pt}
  \begin{tabular}{p{2.8cm} *{6}{>{\centering\arraybackslash}p{1.1cm}}}
  \toprule
  \multirow{3}{*}{\textbf{Simulation Task}}
    & \multicolumn{2}{c}{{\textbf{ACT}~\cite{zhao2023act}}}
    & \multicolumn{2}{c}{{\textbf{OpenVLA-OFT}~\cite{kim2025openvlaoft}}}
    & \multicolumn{2}{c}{{$\mathbf{\pi}_{\mathbf{0.5}}$~\cite{pi05}}} \\
  \cmidrule(lr){2-3} \cmidrule(lr){4-5} \cmidrule(lr){6-7}
  & SR & MS & SR & MS & SR & MS\\
  \midrule
  \textit{Adjust Bottle}
   & 37\% & 46.18
   & 16\% & 26.19
   & 12\% & 24.65 \\
  \textit{Beat Block Hammer}
   & 17\% & 30.85
   & 7\% & 19.99
   & 23\% & 36.32 \\
  \textit{Click Alarmclock}
   & 3\% & 5.80
   & 9\% & 13.63
   & 8\% & 13.92 \\
  \textit{Click Bell}
   & 0\% & 7.89
   & 1\% & 9.41
   & 1\% & 10.77 \\
  \textit{Dump Bin Bigbin}
   & 0\% & 15.42
   & 0\% & 14.31
   & 0\% & 15.70 \\
  \addlinespace
  \textit{Grab Roller}
   & 24\% & 52.27
   & 24\% & 39.02
   & 6\% & 20.18 \\
  \textit{Handover Block}
   & 0\% & 49.51
   & 3\% & 29.02
   & 0\% & 29.13 \\
  \textit{Handover Mic}
   & 13\% & 36.52
   & 8\% & 18.30
   & 7\% & 7.00 \\
  \textit{Hanging Mug}
   & 21\% & 36.38
   & 11\% & 23.98
   & 0\% & 13.82 \\
  \textit{Move Can Pot}
   & 0\% & 42.52
   & 29\% & 49.75
   & 5\% & 37.66 \\
  \addlinespace
  \textit{Move Pillbottle Pad}
   & 0\% & 14.99
   & 0\% & 11.24
   & 5\% & 16.76 \\
  \textit{Move P.Card Away}
   & 9\% & 32.83
   & 0\% & 13.92
   & 1\% & 15.87 \\
  \textit{Move Stapler Pad}
   & 0\% & 19.83
   & 0\% & 10.46
   & 2\% & 13.90 \\
  \textit{Place A2B Left}
   & 0\% & 27.15
   & 3\% & 23.52
   & 3\% & 24.65 \\
  \textit{Place A2B Right}
   & 0\% & 28.01
   & 0\% & 17.74
   & 1\% & 22.73 \\
  \textit{Place Bread Basket}
   & 0\% & 16.71
   & 0\% & 15.94
   & 17\% & 22.80 \\
   \addlinespace
  \textit{Place Bread Skillet}
   & 5\% & 33.74
   & 1\% & 17.92
   & 5\% & 15.43 \\
  \textit{Place Can Basket}
   & 1\% & 32.08
   & 8\% & 37.73
   & 7\% & 23.75 \\
  \textit{Place Container Plate}
   & 24\% & 30.77
   & 14\% & 20.61
   & 33\% & 38.11 \\
  \textit{Place Empty Cup}
   & 2\% & 11.59
   & 0\% & 7.46
   & 7\% & 13.74 \\
  \textit{Place Fan}
   & 0\% & 10.97
   & 0\% & 4.57
   & 1\% & 10.85 \\
   \addlinespace
  \textit{Place Mouse Pad}
   & 0\% & 24.06
   & 0\% & 13.63
   & 6\% & 18.59 \\
  \textit{Place Object Basket}
   & 19\% & 30.93
   & 16\% & 35.51
   & 6\% & 23.96 \\
  \textit{Place Object Scale}
   & 1\% & 17.69
   & 0\% & 9.10
   & 5\% & 15.93 \\
  \textit{Place Object Stand}
   & 0\% & 13.27
   & 2\% & 17.23
   & 3\% & 11.39 \\
  \textit{Place Phone Stand}
   & 0\% & 22.93
   & 0\% & 14.73
   & 1\% & 16.35 \\
   \addlinespace
  \textit{Place Shoe}
   & 1\% & 31.70
   & 6\% & 21.76
   & 28\% & 36.09 \\
  \textit{Press Stapler}
   & 2\% & 11.00
   & 2\% & 12.31
   & 1\% & 10.59 \\
  \textit{Put Bottles Dustbin}
   & 0\% & 17.39
   & 29\% & 47.36
   & 13\% & 28.30 \\
  \textit{Put Object Cabinet}
   & 7\% & 27.01
   & 7\% & 24.61
   & 2\% & 17.23 \\
  \textit{Rotate QRcode}
   & 0\% & 34.38
   & 8\% & 17.51
   & 2\% & 13.31 \\
   \addlinespace
  \textit{Scan Object}
   & 2\% & 25.78
   & 0\% & 7.15
   & 0\% & 14.22 \\
  \textit{Shake Bottle}
   & 18\% & 36.52
   & 12\% & 16.43
   & 19\% & 24.15 \\
  \textit{Shake Bottle Horiz.}
   & 23\% & 45.63
   & 19\% & 25.61
   & 24\% & 32.82 \\
  \textit{Stamp Seal}
   & 0\% & 15.73
   & 0\% & 12.93
   & 7\% & 24.76 \\
  \midrule
  \textbf{\textit{Average (\%)}}
   & 6.54 & 26.74
   & 6.71 & 20.02
   & 7.46 & 20.44 \\
  \bottomrule
  \end{tabular}
  \vspace{-0.3cm}
\end{table}
\begin{table}[ht]
  \centering
  \scriptsize
  \caption{\textbf{Detailed quantitative evaluation of VLA models transferring from dynamic to static settings ($D \to S$).} This table reports the per-task performance across all 35 tasks for models trained in the dynamic environment (DOMINO) and evaluated back in the standard static environment.}
  \label{tab:robotwin-ds}
  \vspace{-5pt}
  \setlength{\tabcolsep}{3pt}
  \begin{tabular}{p{2.8cm} *{6}{>{\centering\arraybackslash}p{1.1cm}}}
  \toprule
  \multirow{3}{*}{\textbf{Simulation Task}}
    & \multicolumn{2}{c}{{\textbf{ACT}~\cite{zhao2023act}}}
    & \multicolumn{2}{c}{{\textbf{OpenVLA-OFT}~\cite{kim2025openvlaoft}}}
    & \multicolumn{2}{c}{{$\mathbf{\pi}_{\mathbf{0.5}}$~\cite{pi05}}} \\
  \cmidrule(lr){2-3} \cmidrule(lr){4-5} \cmidrule(lr){6-7}
  & SR & MS & SR & MS & SR & MS\\
  \midrule
  \textit{Adjust Bottle}
   & 97\% & 97.00
   & 65\% & 65.00
   & 70\% & 70.00 \\
  \textit{Beat Block Hammer}
   & 0\% & 0.00
   & 18\% & 38.42
   & 27\% & 27.00 \\
  \textit{Click Alarmclock}
   & 7\% & 7.00
   & 24\% & 24.00
   & 24\% & 24.00 \\
  \textit{Click Bell}
   & 6\% & 6.00
   & 10\% & 10.00
   & 13\% & 13.00 \\
  \textit{Dump Bin Bigbin}
   & 20\% & 20.00
   & 23\% & 23.00
   & 26\% & 26.00 \\
  \addlinespace
  \textit{Grab Roller}
   & 59\% & 59.00
   & 42\% & 42.00
   & 59\% & 59.00 \\
  \textit{Handover Block}
   & 14\% & 14.00
   & 5\% & 5.00
   & 3\% & 3.00 \\
  \textit{Handover Mic}
   & 1\% & 1.00
   & 0\% & 0.00
   & 7\% & 7.00 \\
  \textit{Hanging Mug}
   & 6\% & 6.00
   & 3\% & 3.00
   & 4\% & 4.00 \\
  \textit{Move Can Pot}
   & 36\% & 36.00
   & 20\% & 20.00
   & 11\% & 11.00 \\
  \addlinespace
  \textit{Move Pillbottle Pad}
   & 5\% & 5.00
   & 2\% & 2.00
   & 7\% & 7.00 \\
  \textit{Move P.Card Away}
   & 2\% & 2.00
   & 5\% & 5.00
   & 26\% & 26.00 \\
  \textit{Move Stapler Pad}
   & 0\% & 0.00
   & 0\% & 0.00
   & 0\% & 0.00 \\
  \textit{Place A2B Left}
   & 0\% & 0.00
   & 0 & 0.00
   & 19\% & 19.00 \\
  \textit{Place A2B Right}
   & 0\% & 0.00
   & 0\% & 0.00
   & 21\% & 21.00 \\
  \textit{Place Bread Basket}
   & 0\% & 0.00
   & 0\% & 0.00
   & 3\% & 3.00 \\
   \addlinespace
  \textit{Place Bread Skillet}
   & 1\% & 1.00
   & 0\% & 0.00
   & 2\% & 2.00 \\
  \textit{Place Can Basket}
   & 3\% & 3.00
   & 3\% & 3.00
   & 17\% & 17.00 \\
  \textit{Place Container Plate}
   & 67\% & 67.00
   & 40\% & 40.00
   & 73\% & 73.00 \\
  \textit{Place Empty Cup}
   & 37\% & 37.00
   & 1\% & 1.00
   & 43\% & 43.00 \\
  \textit{Place Fan}
   & 0\% & 0.00
   & 0\% & 0.00
   & 6\% & 6.00 \\
   \addlinespace
  \textit{Place Mouse Pad}
   & 0\% & 0.00
   & 0\% & 0.00
   & 3\% & 3.00 \\
  \textit{Place Object Basket}
   & 19\% & 19.00
   & 23\% & 23.00
   & 16\% & 16.00 \\
  \textit{Place Object Scale}
   & 0\% & 0.00
   & 0\% & 0.00
   & 8\% & 8.00 \\
  \textit{Place Object Stand}
   & 0\% & 0.00
   & 22\% & 22.00
   & 27\% & 27.00 \\
  \textit{Place Phone Stand}
   & 0\% & 0.00
   & 1\% & 1.00
   & 20\% & 20.00 \\
   \addlinespace
  \textit{Place Shoe}
   & 0\% & 0.00
   & 5\% & 5.00
   & 15\% & 15.00 \\
  \textit{Press Stapler}
   & 31\% & 31.00
   & 51\% & 51.00
   & 35\% & 35.00 \\
  \textit{Put Bottles Dustbin}
   & 0\% & 0.00
   & 0\% & 0.00
   & 1\% & 1.00 \\
  \textit{Put Object Cabinet}
   & 1\% & 1.00
   & 7\% & 7.00
   & 20\% & 20.00 \\
  \textit{Rotate QRcode}
   & 0\% & 0.00
   & 12\% & 12.00
   & 10\% & 10.00 \\
   \addlinespace
  \textit{Scan Object}
   & 0\% & 0.00
   & 2\% & 2.00
   & 5\% & 5.00 \\
  \textit{Shake Bottle}
   & 43\% & 43.00
   & 54\% & 54.00
   & 83\% & 83.00 \\
  \textit{Shake Bottle Horiz.}
   & 34\% & 34.00
   & 55\% & 55.00
   & 84\% & 84.00 \\
  \textit{Stamp Seal}
   & 0\% & 0.00
   & 0\% & 0.00
   & 1\% & 1.00 \\
  \midrule
  \textbf{\textit{Average (\%)}}
   & 13.97 & 13.97
   & 14.09 & 14.67
   & 22.54 & 22.54 \\
  \bottomrule
  \end{tabular}
  \vspace{-0.3cm}
\end{table}
\begin{table}[ht]
  \centering
  \scriptsize
  \caption{\textbf{Detailed quantitative evaluation of VLA models in the fine-tuning dynamic setting ($D \to D$).} This table reports the per-task success rates across all 35 tasks for models fine-tuned and evaluated in the dynamic environment. These results expand upon the \textit{FT} columns in Tab. 1 of the main text.}
  \label{tab:robotwin-dd}
  \vspace{-5pt}
  \setlength{\tabcolsep}{3pt}
  \begin{tabular}{p{2.8cm} *{6}{>{\centering\arraybackslash}p{1.1cm}}}
  \toprule
  \multirow{3}{*}{\textbf{Simulation Task}}
    & \multicolumn{2}{c}{{\textbf{ACT}~\cite{zhao2023act}}}
    & \multicolumn{2}{c}{{\textbf{OpenVLA-OFT}~\cite{kim2025openvlaoft}}}
    & \multicolumn{2}{c}{{$\mathbf{\pi}_{\mathbf{0.5}}$~\cite{pi05}}} \\
  \cmidrule(lr){2-3} \cmidrule(lr){4-5} \cmidrule(lr){6-7}
  & SR & MS & SR & MS & SR & MS\\
  \midrule
  \textit{Adjust Bottle}
   & 65\% & 67.57
   & 58\% & 63.57
   & 52\% & 71.45 \\
  \textit{Beat Block Hammer}
   & 0\% & 5.14
   & 11\% & 23.20
   & 10\% & 23.22 \\
  \textit{Click Alarmclock}
   & 8\% & 11.05
   & 6\% & 10.53
   & 14\% & 18.47 \\
  \textit{Click Bell}
   & 0\% & 8.24
   & 3\% & 11.24
   & 0\% & 6.35 \\
  \textit{Dump Bin Bigbin}
   & 0\% & 11.14
   & 0\% & 17.84
   & 0\% & 16.64 \\
  \addlinespace
  \textit{Grab Roller}
   & 23\% & 47.80
   & 28\% & 43.87
   & 10\% & 34.32 \\
  \textit{Handover Block}
   & 11\% & 51.07
   & 9\% & 61.17
   & 1\% & 33.95 \\
  \textit{Handover Mic}
   & 23\% & 40.35
   & 25\% & 37.58
   & 5\% & 29.10 \\
  \textit{Hanging Mug}
   & 28\% & 43.62
   & 12\% & 35.49
   & 5\% & 25.59 \\
  \textit{Move Can Pot}
   & 26\% & 53.55
   & 29\% & 57.26
   & 7\% & 40.34 \\
  \addlinespace
  \textit{Move Pillbottle Pad}
   & 1\% & 8.15
   & 1\% & 9.93
   & 6\% & 15.93 \\
  \textit{Move P.Card Away}
   & 1\% & 26.54
   & 1\% & 7.94
   & 8\% & 24.84 \\
  \textit{Move Stapler Pad}
   & 0\% & 13.12
   & 0\% & 10.87
   & 0\% & 13.38 \\
  \textit{Place A2B Left}
   & 0\% & 21.39
   & 1\% & 22.37
   & 2\% & 26.04 \\
  \textit{Place A2B Right}
   & 0\% & 28.33
   & 1\% & 25.00
   & 1\% & 23.63 \\
  \textit{Place Bread Basket}
   & 1\% & 15.05
   & 1\% & 7.98
   & 8\% & 15.89 \\
   \addlinespace
  \textit{Place Bread Skillet}
   & 12\% & 28.54
   & 1\% & 20.32
   & 9\% & 22.30 \\
  \textit{Place Can Basket}
   & 4\% & 29.47
   & 5\% & 37.78
   & 6\% & 29.48 \\
  \textit{Place Container Plate}
   & 30\% & 35.75
   & 10\% & 16.87
   & 22\% & 28.46 \\
  \textit{Place Empty Cup}
   & 6\% & 12.49
   & 0\% & 7.74
   & 2\% & 12.71 \\
  \textit{Place Fan}
   & 1\% & 6.64
   & 1\% & 6.25
   & 2\% & 11.72 \\
   \addlinespace
  \textit{Place Mouse Pad}
   & 0\% & 14.31
   & 1\% & 13.61
   & 5\% & 19.03 \\
  \textit{Place Object Basket}
   & 21\% & 35.96
   & 18\% & 37.36
   & 11\% & 30.19 \\
  \textit{Place Object Scale}
   & 0\% & 12.89
   & 0\% & 10.92
   & 3\% & 15.37 \\
  \textit{Place Object Stand}
   & 0\% & 7.42
   & 0\% & 8.53
   & 1\% & 8.97 \\
  \textit{Place Phone Stand}
   & 0\% & 16.45
   & 0\% & 12.66
   & 2\% & 17.98 \\
   \addlinespace
  \textit{Place Shoe}
   & 0\% & 7.92
   & 4\% & 17.39
   & 25\% & 32.84 \\
  \textit{Press Stapler}
   & 4\% & 15.31
   & 5\% & 13.53
   & 6\% & 13.27 \\
  \textit{Put Bottles Dustbin}
   & 21\% & 38.57
   & 19\% & 29.81
   & 17\% & 33.30 \\
  \textit{Put Object Cabinet}
   & 2\% & 39.61
   & 1\% & 42.07
   & 11\% & 46.24 \\
  \textit{Rotate QRcode}
   & 0\% & 12.63
   & 11\% & 23.31
   & 4\% & 17.90 \\
   \addlinespace
  \textit{Scan Object}
   & 2\% & 19.82
   & 4\% & 19.93
   & 1\% & 20.48 \\
  \textit{Shake Bottle}
   & 14\% & 27.23
   & 17\% & 23.83
   & 33\% & 49.40 \\
  \textit{Shake Bottle Horiz.}
   & 18\% & 41.70
   & 31\% & 38.75
   & 42\% & 62.72 \\
  \textit{Stamp Seal}
   & 8\% & 19.10
   & 3\% & 15.47
   & 6\% & 22.58 \\
  \midrule
  \textbf{\textit{Average (\%)}}
   & 9.43 & 24.97
   & 9.06 & 24.06
   & 9.63 & 26.12 \\
  \bottomrule
  \end{tabular}
  \vspace{-0.3cm}
\end{table}

\end{document}